\documentclass[journal]{IEEEtran}

\pdfoutput=1

\usepackage{bobbystyle_IEEEjournal}

\usepackage{tikz}
\usetikzlibrary{spy}

\usepackage[resetlabels]{multibib}
\newcites{Supp}{References}
\bstctlcite[@auxoutSupp]{BSTcontrol2}

\newcommand{\dquotes}[1]{``#1''}

\usepackage{chngcntr}
\counterwithout{equation}{section}

\newcommand{\specialcell}[2][c]{%
  \begin{tabular}[#1]{@{}c@{}}#2\end{tabular}}

\newcolumntype{C}[1]{>{\centering\let\newline\\\arraybackslash\hspace{0pt}}m{#1}}  

\renewcommand\thetheorem{\arabic{section}.\arabic{theorem}}
\renewcommand{\theequation}{\arabic{equation}}

\newenvironment{myquote}[1]%
  {\list{}{\leftmargin=#1\rightmargin=#1}\item[]}%
  {\endlist}

\begin{document}

\title{Convolutional Dictionary Learning: Acceleration and Convergence}

\author{Il Yong Chun, \textit{Member}, \textit{IEEE}, and Jeffrey A. Fessler, \textit{Fellow}, \textit{IEEE}

\thanks{This work is supported in part by the Keck Foundation and by a UM-SJTU seed grant.}

\thanks{This paper has supplementary material. The prefix ``S'' indicates the numbers in section, equation, figure, and table in the supplementary material.}

\thanks{Il Yong Chun and Jeffrey A. Fessler are with the Department of Electrical Engineering and Computer Science, The University of Michigan, Ann Arbor, MI 48019 USA (email: iychun@umich.edu; fessler@umich.edu).}
}

\maketitle

\begin{abstract}
Convolutional dictionary learning (CDL or sparsifying CDL) has many applications in image processing and computer vision.
There has been growing interest in developing efficient algorithms for CDL, mostly relying on the augmented Lagrangian (AL) method or the variant alternating direction method of multipliers (ADMM).
When their parameters are properly tuned, AL methods have shown fast convergence in CDL.
However, the parameter tuning process is not trivial due to its data dependence and, in practice, the convergence of AL methods depends on the AL parameters for nonconvex CDL problems.
To moderate these problems, this paper proposes a new practically feasible and convergent \textit{Block Proximal Gradient method using a Majorizer} (BPG-M) for CDL. 
The BPG-M-based CDL is investigated with different block updating schemes and majorization matrix designs, and further accelerated by incorporating some momentum coefficient formulas and restarting techniques.
All of the methods investigated incorporate a boundary artifacts removal (or, more generally, sampling) operator in the learning model.
Numerical experiments show that, without needing any parameter tuning process, the proposed BPG-M approach converges more stably to desirable solutions of lower objective values than the existing state-of-the-art ADMM algorithm and its memory-efficient variant do. Compared to the ADMM approaches, the BPG-M method using a multi-block updating scheme is particularly useful in single-threaded CDL algorithm handling large datasets, due to its lower memory requirement and no polynomial computational complexity.
Image denoising experiments show that, for relatively strong additive white Gaussian noise, the filters learned by BPG-M-based CDL outperform those trained by the ADMM approach.
\end{abstract}


\section{Introduction} \label{sec:intro}

\IEEEPARstart{A}{daptive} sparse representations can model intricate redundancies of complex structured images in a wide range of applications. \dquotes{Learning} sparse representations from large datasets, such as (sparsifying) dictionary learning, is a growing trend.
Patch-based dictionary learning is a well-known technique for obtaining sparse representations of training signals \cite{Bruckstein&Dohono&Elad:09SIAM, Coates&NG:bookCh, Mairal&Bach&Ponce:14FTCGV, Aharon&Elad&Bruckstein:06TSP, Xu&Yin:16IPI}. The learned dictionaries from patch-based techniques have been applied to various image processing and computer vision problems, i.e., image inpainting, denoising, deblurring, compression, classification, etc. (see \cite{Bruckstein&Dohono&Elad:09SIAM, Coates&NG:bookCh, Mairal&Bach&Ponce:14FTCGV, Aharon&Elad&Bruckstein:06TSP, Xu&Yin:16IPI} and references therein). However, patch-based dictionary learning has three fundamental limitations. Firstly, learned basis elements often are shifted versions of each other (i.e., \textit{translation-variant} dictionaries) and underlying structure of the signal may be lost, because each patch---rather than an entire image---is synthesized (or reconstructed) individually.  Secondly, sparse representation for a whole image is highly redundant because neighboring and even overlapping patches are sparsified independently.
Thirdly, using many overlapping patches across the training and test signals hinders using \dquotes{big data}---i.e., training data with the large number of signals or high-dimensional signals; for example, see \cite[\S3.2]{Heide&eta:15CVPR} or Section~\ref{sec:result:BPGMvsADMM}---and discourages the learned dictionary from being applied to large-scale inverse problems, respectively.

\textit{Convolutional dictionary learning} (CDL or sparsifying CDL), motivated by the perspective of modeling receptive fields in human vision \cite{Olshausen&Field:96Nature, Olshausen&Field:97VR} and convolutional neural networks \cite{LeCun&Bengio&Hinton:15Nature, LeCun&etal:98ProcIEEE, Krizhevsky&etal:12NIPS}, can overcome the problems of patch-based dictionary learning \cite{Zeiler&etal:10CVPR, Kavukcuoglu&etal:10NIPS, Bristow&etal:13CVPR, Wohlberg:14ICASSP, Kong&Fowkes:14techRep, Bristow&Lucey:14arXiv, Heide&eta:15CVPR, Wohlberg:16TIP, Wohlberg:16ICIP, Sorel&Michal:16DSP, Papyan&Sulam&Elad:16arXiv-part2, Chun&Fessler:17SAMPTA}.
In particular, signals displaying translation invariance in any dimension (e.g., natural images and sounds) are better represented using a CDL approach \cite{Kavukcuoglu&etal:10NIPS}. 
In addition, the sparse coding step (e.g., see Section~\ref{sec:BPGM:spCd}) in CDL is closely related to convolutional neural networks \cite{Papyan&Romano&Elad:16arXiv-CNN}.
Learned convolutional dictionaries have been applied to various image processing and computer vision problems, e.g., image inpainting, denoising, classification, recognition, detection, etc. (see \cite{Zeiler&etal:10CVPR, Kavukcuoglu&etal:10NIPS, Bristow&etal:13CVPR, Heide&eta:15CVPR, Wohlberg:16TIP, Chun&Fessler:17SAMPTA}).

CDL in 2D (and beyond) has two major challenges.
The first concern lies in its optimization techniques: \textit{1)} computational complexity, \textit{2) and memory-inefficient algorithm (particularly augmented Lagrangian (AL) method),} and \textit{3)} convergence guarantees.
In terms of computational complexity, the most recent advances include algorithmic development with AL method (e.g., alternating direction method of multipliers, ADMM \cite{Boyd&Parikh&Chu&Peleato&Eckstein:11FTML, Parikh&Boyd:14FTO}) \cite{Bristow&etal:13CVPR, Wohlberg:14ICASSP, Kong&Fowkes:14techRep, Bristow&Lucey:14arXiv, Heide&eta:15CVPR, Wohlberg:16TIP, Wohlberg:16ICIP, Sorel&Michal:16DSP, Papyan&Sulam&Elad:16arXiv-part2} and fast proximal gradient (FPG) method \cite{Chalasani&etal:13IJCNN} (e.g., fast iterative shrinkage-thresholding algorithm, FISTA \cite{Beck&Teboulle:09SIAM}).
Although AL methods have shown fast convergence in \cite{Bristow&etal:13CVPR, Heide&eta:15CVPR, Wohlberg:16TIP} (and faster than the continuation-type approach in \cite{Zeiler&etal:10CVPR, Bristow&etal:13CVPR}), they require tricky parameter tuning processes for acceleration and (stable) convergence, due to their dependence on training data (specifically, preprocessing of training data, types of training data, and the number and size of training data and filters).
In particular, in the AL frameworks, the number of AL parameters to be tuned increases as CDL models become more sophisticated, e.g., \textit{a)} for the CDL model including a boundary truncation (or, more generally, sampling) operator \cite{Heide&eta:15CVPR}, one needs to tune four additional AL parameters; \textit{b)} for the CDL model using adaptive contrast enhancement (CDL-ACE), six additional AL parameters should be tuned \cite{Chun&Fessler:17SAMPTA}!
The FPG method introduced in \cite{Chalasani&etal:13IJCNN} is still not free from the parameter selection problem. The method uses a backtracking scheme because it is impractical to compute the Lipschitz constant of the tremendous-sized system matrix of 2D CDL.
Another limitation of the AL approaches is that they require larger amount of memory as CDL models become more sophisticated (see examples above), because one often needs to introduce more auxiliary variables.
This drawback can be particularly problematic for some applications, e.g., image classification \cite{Chen&etal:16ICIP}, because their performance improves as the number of training images increases for learning filters \cite[Fig.~3]{Chen&etal:16ICIP}.
In terms of theoretical aspects, there exists no known convergence analysis (even for local convergence) noting that CDL is a nonconvex problem.
Without a convergence theorem, it is possible that the iterates could diverge.

\begin{figure}[!t]
\small\addtolength{\tabcolsep}{-2pt}
\centering

\begin{tabular}{cc}
\includegraphics[scale=0.7]{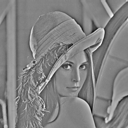} & \hspace{-0.55em}
\includegraphics[scale=0.7]{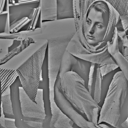} \\
\multicolumn{2}{c}{\small (a) Target (preprocessed) images to synthesize} \\ 
\multicolumn{2}{c}{\hspace{3.2em}\includegraphics[scale=0.5, trim=4em 13.5em 1em 12em, clip]{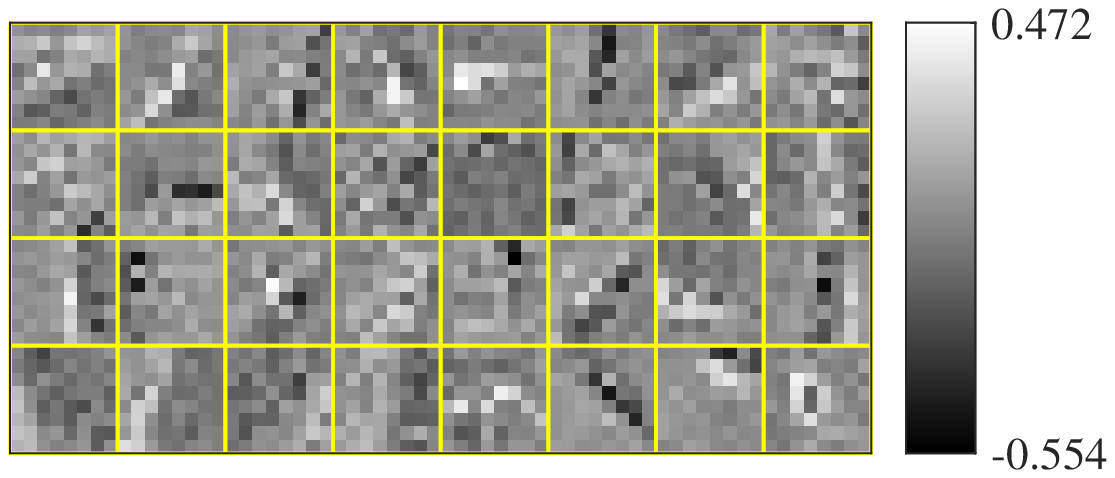}} \\
\includegraphics[scale=0.7]{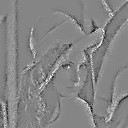} &
\includegraphics[scale=0.7]{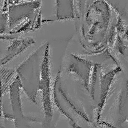} \vspace{-1.5em} \\
{\scriptsize \YL{$\textmd{PSNR} = 25.02$~dB}} & {\scriptsize \YL{$\textmd{PSNR} = 23.87$~dB}} \vspace{0.5em} \\
\multicolumn{2}{c}{\small (b) Results from the CDL model in \protect\cite{Bristow&etal:13CVPR} \vspace{-0.25em}} \\
\multicolumn{2}{c}{\small (without considering sampling operator)} \\
\multicolumn{2}{c}{\hspace{3.2em}\includegraphics[scale=0.5, trim=4em 13.5ems 1em 12em, clip]{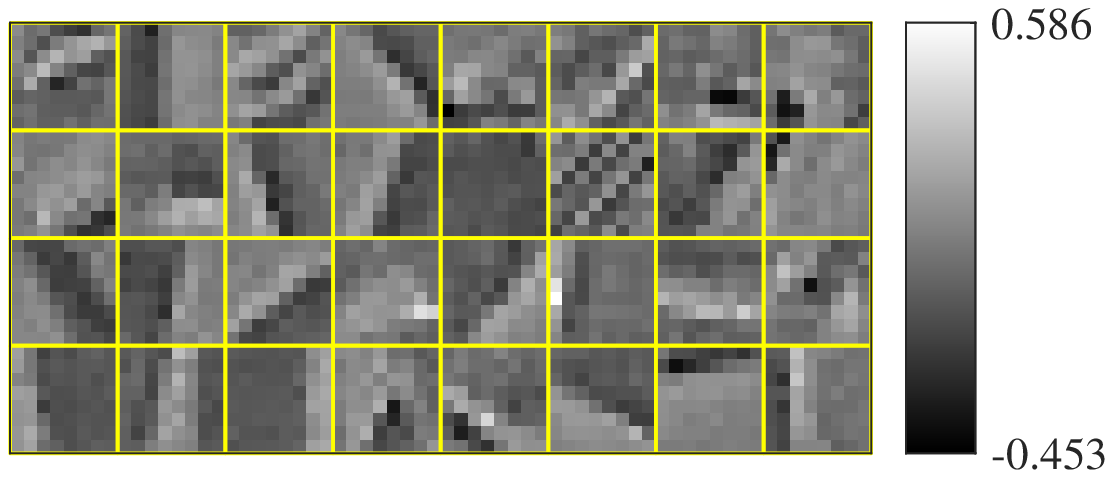}} \\
\includegraphics[scale=0.7]{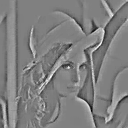} &
\includegraphics[scale=0.7]{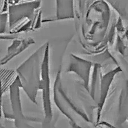} \vspace{-1.5em} \\
{\scriptsize \YL{$\textmd{PSNR} = 29.31$~dB}} & {\scriptsize \YL{$\textmd{PSNR} = 28.98$~dB}} \vspace{0.5em} \\
\multicolumn{2}{c}{\small (c) Results from the CDL model in \protect\cite{Heide&eta:15CVPR} \vspace{-0.25em}} \\ 
\multicolumn{2}{c}{\small (including sampling operator)} 
\end{tabular}

\caption{Examples of learned filters and synthesized images from sparse datasets with different CDL models ($32$ $8 \!\times\! 8$-sized filters were learned from two \textit{sparse} $128 \!\times\! 128$-sized training images; sparse images were generated by $\approx \! 60$\% random sampling---see, for example, \protect\cite[Fig.~5]{Heide&eta:15CVPR}; the experiments are based on the dataset, initialization, parameters, and preprocessing method used in \protect\cite[ver.~0.0.7,~\dquotes{demo\_cbpdndlmd.m}]{Wohlberg:sporco}). In sparse data settings, including a sampling operator in CDL (e.g., $P_B$ in \R{eq:sys:synth}) allows to learn filters leading better image synthesis performance (note that the results in (c) correspond to those in \protect\cite[\S4.3]{Heide&eta:15CVPR}). Note that the image synthesis accuracy in training affects the performance in testing models, e.g., image denoising---see Fig.~\ref{fig:denoise}.}
\label{fig:CDL:PBvsNoPB}
\end{figure}

The second problem is boundary effects associated with the convolution operators.
The authors of \cite{Bristow&etal:13CVPR} experimentally showed that neglecting boundary effects might be acceptable in learning small-sized filters under periodic boundary condition.  
However, this is not necessarily true as illustrated in \cite[\S4.2]{Heide&eta:15CVPR} with examples using $11$-by-$11$ filters: high-frequency components still exist on image boundaries of synthesized images (i.e., $\sum_{k} d_{k} \circledast z_{l,k}$ in \R{eq:sys:synth}).
Neglecting them can be unreasonable for learning larger filters or for general boundary conditions.
As pointed out in \cite{Kavukcuoglu&etal:10NIPS}, if one does not properly handle the boundary effects, the CDL model can be easily influenced by the boundary effects even if one uses non-periodic boundary conditions for convolutions (e.g., the reflective boundary condition \cite[\S2]{Bristow&etal:13CVPR}). 
Specifically, the synthesis errors (i.e., the $\ell^2$ data fitting term in \R{eq:sys:synth} without the truncation operator $P_B$) close to the boundaries might grow much larger compared to those in the interior, because sparse code pixels near the boundaries are convolved less than those in the interior.
To remove the boundary artifacts, the formulation in \cite{Heide&eta:15CVPR} used a boundary truncation operator that was also used in image deblurring problem in \cite{Matakos&etal:13TIP, Almeida&Figueiredo:13TIP}.
The truncation operator is inherently considered in the local patch-based CDL framework \cite{Papyan&Sulam&Elad:16arXiv-part2}.
In the big data setup, it is important to learn decent filters with less training data (but not necessarily decreasing the number of training signals---see above).
The boundary truncation operator in \cite{Heide&eta:15CVPR, Matakos&etal:13TIP, Almeida&Figueiredo:13TIP} can be generalized to a sampling operator that reduces the amount of each training signal \cite[\S4.3]{Heide&eta:15CVPR}. 
Considering the sampling operator, the CDL model in \cite{Heide&eta:15CVPR} learns filters leading better image synthesis accuracy than that without it, e.g., \cite{Bristow&etal:13CVPR}; see Fig.~\ref{fig:CDL:PBvsNoPB}.

In this paper, we consider the CDL model in \cite{Heide&eta:15CVPR} that avoids boundary artifacts or, more generally, incorporates sampling operators.\footnote{We do not consider the boundary handling CDL model in \cite[\S3.1]{Wohlberg:16TIP}, because of its inconsistency in boundary conditions. 
Its constraint trick in \cite[(13)]{Wohlberg:16TIP} casts zero-boundary on sparse codes \cite[(14)]{Wohlberg:16TIP}; however, its CDL algorithm solves the model with the Parseval tricks \cite{Bristow&etal:13CVPR, Heide&eta:15CVPR, Wohlberg:16TIP} using periodic boundary condition. See more potential issues in \cite[\S3.1]{Wohlberg:16TIP}.}
We propose a new practically feasible and convergent block proximal gradient (BPG \cite{Xu&Yin:13SIAM}) algorithmic framework, called \textit{Block Proximal Gradient method using Majorizer} (BPG-M), and accelerate it with two momentum coefficient formulas and two restarting techniques.
For the CDL model \cite{Heide&eta:15CVPR}, we introduce two block updating schemes within the BPG-M framework: \textit{a)} \textit{two-block} scheme and \textit{b)} \textit{multi-block} scheme.
In particular, the proposed multi-block BPG-M approach has several benefits over the ADMM approach \cite{Heide&eta:15CVPR} and its memory-efficient variant (see below): \textit{1)} guaranteed local convergence (or global convergence if some conditions are satisfied) without difficult parameter tuning processes, \textit{2)} lower memory usage, \textit{3) no polynomial computational complexity (particularly, no quadratic complexity with the number of training images)}, and \textit{4)} (empirically) reaching lower objective values.
Specifically, for small datasets, the BPG-M approach converges more stably to a \dquotes{desirable} solution of lower objective values with comparable computational time;\footnote{Throughout the paper, \dquotes{desirable} solutions mean that \textit{1)} the learned filters capture structures of training images; \textit{2)} the corresponding sparse codes are sufficiently sparse; \textit{3)} the filters and sparse codes can properly synthesize training images through convolutional operators.} for larger datasets (i.e., datasets with the larger number of training images or larger sized training images), it stably converges to the desirable solution with better memory flexibility and/or lower computational complexity.
Section~\ref{sec:BPG-CDL} introduces the BPG-M, analyzes its convergence, and develops acceleration methods. 
Section~\ref{sec:BPGM-twoBlk} applies the proposed BPG-M methods to two-block CDL.
Section~\ref{sec:BPGM-multiBlk} proposes multi-block CDL using BPG-M that is particularly useful for single-thread computation.
Sections~\ref{sec:BPGM-twoBlk}--\ref{sec:BPGM-multiBlk} include computationally efficient majorization matrices and efficient proximal mapping methods.
Section~\ref{sec:CDL-ACE} summarizes CDL-ACE \cite{Chun&Fessler:17SAMPTA} and the corresponding image denoising model using learned filters.
Section~\ref{sec:result} reports numerical experiments that show the benefits---convergence stability, memory flexibility, no polynomial (specifically, quadratic and cubic) complexity, and reaching lower objective values---of the BPG-M framework in CDL, and illustrate the effectiveness of a tight majorizer design and the accelerating schemes on BPG-M convergence rate in CDL. 
Furthermore, Section~\ref{sec:result} reports image denoising experiments that show, for relatively strong additive white Gaussian (AWGN) noise,  i.e., $\textmd{SNR} \!=\! 10$~dB, the learned filters by BPG-M-based CDL improve image denoising compared to the filters trained by ADMM \cite{Heide&eta:15CVPR}.

Throughout the paper, we compare the proposed BPG-M methods mainly to Heide et al.'s ADMM framework in \cite{Heide&eta:15CVPR} using their suggested ADMM parameters, and its memory-efficient variant applying the linear solver in \cite[\S\Romnum{3}-B]{Wohlberg:16TIP} to solve \cite[(10)]{Heide&eta:15CVPR}. 
Thees ADMM frameworks can be viewed as a \textit{theoretically stable} block coordinate descent (BCD, e.g., \cite{Xu&Yin:13SIAM}) method using two blocks if sufficient inner (i.e., ADMM) iterations are used to ensure descent for each block update, whereas methods that use a single inner iteration for each block update may not be guaranteed to descend---see, for example, \cite[Fig.~2,~AVA-MD]{Wohlberg:16ICIP}.

\section{CDL Model and Existing AL-Based Optimization Methods} \label{sec:CDLmodel}

The CDL problem corresponds to the following joint optimization problem \cite{Heide&eta:15CVPR} (mathematical notations are provided in Appendix):
\begingroup
\setlength{\thinmuskip}{1.5mu}
\setlength{\medmuskip}{2mu plus 1mu minus 2mu}
\setlength{\thickmuskip}{2.5mu plus 2.5mu}
\ea{
\begin{split}
\label{eq:sys:synth}
\hspace{-1.05em}\min_{\{ d_k \}, \{ z_{l,k} \}} &~ \sum_{l=1}^L \frac{1}{2} \left\| y_l - P_B \sum_{k=1}^K d_{k} \circledast z_{l,k} \right\|_2^2 + \alpha \sum_{k=1}^K \| z_{l,k} \|_1 
\\
\hspace{-1.05em}\mbox{s.t.} ~~~~&~ \| d_{k} \|_2^2 \leq 1, \qquad k = 1,\ldots, K,
\end{split}
}
\endgroup
where $\{ d_{k} \in \bbC^{D}: k = 1,\ldots, K \}$ is a set of synthesis convolutional filters to be learned, $\{ y_l \in \bbC^{N} : l = 1,\ldots,L \}$ is a set of training data, $\circledast$ denotes a circular convolution operator, $\{ z_{l,k} \in \bbC^{N} : l = 1,\ldots, L, k = 1,\ldots,K \}$ is a set of sparse codes, $P_{B} \in \bbC^{N \times \tilde{N}}$ is a projection matrix with $| B | = N$ and $N < \tilde{N}$, and $B$ is a list of distinct indices from the set $\{1, ..., \tilde{N}\}$ that correspond to truncating the boundaries of the padded convolution $\sum_{k=1}^K d_{k} \circledast z_{l,k}$. 
Here, $D$ is the filter size, $K$ is the number of convolution operators, $N$ is the dimension of training data, $\tilde{N}$ is the dimension after convolution with padding,\footnote{The convolved signal has size of $\tilde{N} = (N_{\mathrm{h}} + K_{\mathrm{h}} - 1) \times (N_{\mathrm{w}} + K_{\mathrm{h}} - 1)$, where the original signal has size $N = N_{\mathrm{h}} \times N_{\mathrm{w}}$, the filter has size $K = K_{\mathrm{h}} \times K_{\mathrm{w}}$, and $\mathrm{w}$ and $\mathrm{h}$ denote the width and height, respectively.} and $L$ is the number of training images.
Note that $D$ is much smaller than $\tilde{N}$ in general.
Using $P_B$ to eliminate boundary artifacts is useful because CDL can be sensitive to the convolution boundary conditions; see Section~\ref{sec:intro} for details.
In sparse data settings, one can generalize $B$ to $\{ {B_l} : | B_l | = S_l  < N, l = 1,\ldots,L \}$, where $B_l$ contains the indices of (randomly) collected samples from $y_l$ \cite[\S4.3]{Heide&eta:15CVPR}, or the indices of the non-zero elements of the $l\rth$ sparse signal, for $l=1,\ldots,L$.

Using Parseval's relation \cite{Bristow&etal:13CVPR, Bristow&Lucey:14arXiv, Kong&Fowkes:14techRep, Heide&eta:15CVPR}, problem \R{eq:sys:synth} is equivalent to the following joint optimization problem (in the frequency domain):
\begingroup
\setlength{\thinmuskip}{1.5mu}
\setlength{\medmuskip}{2mu plus 1mu minus 2mu}
\setlength{\thickmuskip}{2.5mu plus 2.5mu}
\ea{
\begin{split}
\label{eq:sys:synthF}
\min_{\{ d_k \}, \{ \tilde{z}_{l} \}} &~ \sum_{l=1}^L \frac{1}{2} \left\| y_l - P_B \left[  \Phi^{-1} \diag(\Phi P_S^T d_{1}) \Phi \right. \right.
\\
&~ \left. \left. \cdots~ \Phi^{-1} \diag(\Phi P_S^T d_{K}) \Phi \right] \tilde{z}_l  \right\|_2^2 +  \alpha \| \tilde{z}_{l} \|_1
\\
\mbox{s.t.} ~~~&~ \| d_k \|_2^2 \leq 1, \qquad k = 1,\ldots, K,
\end{split}
}
\endgroup
where $\Phi$ denotes the $\tilde{N}$-point 2D (unnormalized) discrete Fourier transform (DFT), $P_S^T \in \bbC^{\tilde{N} \times D}$ is zero-padding matrix, $S$ is a list of indices that correspond to a small support of the filter with $| S | = D$ (again, $D \ll \tilde{N}$), where $\tilde{z}_l = [\tilde{z}_{l,1}^H, \ldots, \tilde{z}_{l,K}^H]^H \in \bbC^{K \tilde{N}}$, and $\{ \tilde{z}_{l,k} \in \bbC^{\tilde{N}} : l = 1,\ldots, L, k = 1,\ldots,K \}$ denotes sparse codes. 
In general, $\tilde{(\cdot)}$ and $\hat{(\cdot)}$ denote a padded signal vector and a transformed vector in frequency domain, respectively.

AL methods are powerful, particularly for non-smooth optimization. 
An AL method was first applied to CDL with Parseval's theorem in \cite{Bristow&etal:13CVPR}, but without handling boundary artifacts.
In \cite{Heide&eta:15CVPR}, ADMM was first applied to solve \R{eq:sys:synth}. A similar spatial domain ADMM framework was introduced in \cite{Wohlberg:16TIP, Wohlberg:16ICIP}.
These AL methods alternate between updating the dictionary $\{ d_k \}$ (the filters) and updating the sparse codes $\{ \tilde{z}_l \}$ (i.e., a two-block update), using AL (or ADMM) methods for each inner update.
In \cite{Heide&eta:15CVPR}, each filter and sparse code update consists of multiple iterations before switching to the other, whereas \cite{Bristow&etal:13CVPR, Wohlberg:16TIP} explored merging all the updates into a single set of iterations.
This single-set-of-iterations scheme based on AL method can be unstable because each filter and sparse code update no longer ensures monotone descent of the cost function.
To improve its stability, one can apply the increasing ADMM parameter scheme \cite[(23)]{Bristow&etal:13CVPR}, the adaptive ADMM parameter selection scheme controlling primal and dual residual norms \cite[\S 3.4.1]{Boyd&Parikh&Chu&Peleato&Eckstein:11FTML}, \cite[\S \Romnum{3}-D]{Wohlberg:16TIP}, or interleaving schemes \cite[Algo.~1]{Bristow&etal:13CVPR}, \cite[\S \Romnum{5}-B]{Wohlberg:16TIP}, \cite{Wohlberg:16ICIP}.
However, it is difficult to obtain theoretical convergence guarantees (even for local convergence) for the AL algorithms using the single-set-of-iterations scheme; in addition, the techniques for reducing instability further complicate theoretical guarantees.

The following section introduces a practical BPG-M method consisting of a single set of updates that guarantees convergence for solving multi-convex problems like CDL.

\section{Convergent Fast BPG-M and Adaptive Restarting} \label{sec:BPG-CDL}

\subsection{BPG-M -- Setup} \label{sec:BPG:setup}

Consider the optimization problem
\be{
\label{eq:multiConvx}
\min_{x \in \cX} ~ F(x_1,\ldots,x_B) := f(x_1,\ldots,x_B) + \sum_{b=1}^B r_b (x_b)
}
where variable $x$ is decomposed into $B$ blocks $x_1,\ldots,x_B$, the set $\cX$ of feasible points is assumed to be closed and \textit{block multi-convex} subset of $\bbR^n$, $f$ is assumed to be a differentiable and \textit{block multi-convex} function, and $r_b$ are extended-value convex functions for $b=1,\ldots,B$. A set $\cX$ is called \textit{block multi-convex} if its projection to each block of variable is convex, i.e., for each $b$ and any fixed $B-1$ blocks $x_1,\ldots,x_{b-1},x_{b+1},\ldots,x_B$, the set
\eas{
& ~ \cX_b (x_1, \ldots, x_{b-1}, x_{b+1}, \ldots, x_B) 
\\
& := \left\{ x_b \in \bbR^{n_b} : (x_1, \ldots, x_{b-1}, x_b, x_{b+1}, \ldots, x_B) \in \cX \right\}
}
is convex. A function $f$ is called \textit{block multi-convex} if for each $b$, $f$ is a convex function of $x_b$, when all the other blocks are fixed. 
In other words, when all blocks are fixed except one block, \R{eq:multiConvx} over the free block is a convex problem.
Extended-value means $r_b (x_b) = \infty$ if $x_b \notin \dom(r_b)$, for $b=1,\ldots,B$. In particular, $r_b$ can be indicator functions of convex sets. We use $r_1,\ldots,r_B$ to enforce individual constraints of $x_1,\ldots,x_B$, when they are present. Importantly, $r_b$ can include nonsmooth functions.

In this paper, we are particularly interested in adopting the following quadratic \textit{majorizer} (i.e., surrogate function) model of the composite function $\varrho(u) = \varrho_1(u) + \varrho_2(u)$ at a given point $v$ to the block multi-convex problem \R{eq:multiConvx}:
\ea{
\label{eq:quadApprox}
\tilde{\varrho}_M (u, v) &= \psi_M (u; v) + \varrho_2 (u),
\nn \\
\psi_M (u; v) &= \varrho_1(v) + \ip{ \nabla \varrho_1(v) }{ u-v } + \frac{1}{2} \nm{ u - v }_{M}^2
}
where $\varrho_1(u)$ and $\varrho_2(u)$ are two convex functions defined on the convex set $\cU$, $\varrho_1(u)$ is differentiable, 
the majorizer $\psi_M (u; v)$ satisfies the following two conditions
\bes{
\varrho_1(v) = \psi_M (v; v) \quad \mathrm{and} \quad \varrho_1(u) \leq \psi_M (u; v), \quad \forall u \in \cU, \forall v,
}
and $M = M^T \succ 0$ is so-called \textit{majorization matrix}. 
The majorizer $\tilde{\rho}_M (u,v)$ has the following unique minimizer
\bes{
u^{\star} = \argmin_{u \in \cU} \, \frac{1}{2} \nm{u - \left( v - M^{-1}  \nabla \varrho_1(v)  \right) }_{M}^2 + \varrho_2(u).
}
Note that decreasing the majorizer $\tilde{\varrho}_M (u, v)$ ensures a monotone decrease of the original cost function $\varrho(u)$.
For example, a majorizer for $\varrho_1(u) = 1/2 \|  y - A u \|_2^2$ is given by
\be{
\label{eq:quad_surrg}
\psi_M (u; v) = \frac{1}{2} \nm{u - \left( v - M_{A}^{-1} A^T \left( A  v - y \right) \right) }_{M_A}^2,
}
where $A \in \bbR^{m \times n}$ and $M_A  \in \bbR^{n \times n}$ is any majorization matrix for the Hessian $A^T A$ (i.e. $M_A \succeq A^T A$).
Other examples include when $\varrho_1$ has Lipschitz-continuous gradient, or $\varrho_1$ is twice continuously differentiable and can be approximated with a majorization matrix $M \succ 0$ for the Hermitian $\nabla^2 \varrho_1 (u) \succeq 0$, $\forall u \in \cU$.

Based on majorizers of the form \R{eq:quadApprox}, the proposed method, BPG-M, is given as follows.
To solve \R{eq:multiConvx}, we minimize $F$ cyclically over each block $x_1,\ldots,x_B$, while fixing the remaining blocks at their previously updated values. Let $x_b^{(i+1)}$ be the value of $x_b$ after its $i\rth$ update, and
\be{
\label{eq:def:f_block}
f_b^{(i)}(x_b) := f(x_1^{(i+1)}, \ldots, x_{b-1}^{(i+1)}, x_b, x_{b+1}^{(i)}, \ldots, x_{B}^{(i)}),
}
for all $b,i$.
At the $b\rth$ step of the $i\rth$ iteration, we consider the updates
\begingroup
\setlength{\thinmuskip}{1.5mu}
\setlength{\medmuskip}{2mu plus 1mu minus 2mu}
\setlength{\thickmuskip}{2.5mu plus 2.5mu}
\eas{
\label{eq:block_update}
& ~ x_b^{(i+1)} \nn
\\
&= \argmin_{ x_b \in \cX_b^{(i)} } \, \ip{ \nabla f_b^{(i)} (\acute{x}_b^{(i)}) }{ x_b - \acute{x}_b^{(i)} } 
+ \frac{1}{2} \nm{ x_b - \acute{x}_b^{(i)} }_{M_b^{(i)}}^2 + r_b (x_b) \nn
\\
&= \argmin_{ x_b \in \cX_b^{(i)} } \, \frac{1}{2} \nm{x_b - \left( \acute{x}_b^{(i)} - \left( M_b^{(i)} \right)^{-1}  \nabla f_b^{(i)} (\acute{x}_b^{(i)}) \right) }_{M_b^{(i)}}^2 \nn
\\
& \qquad \qquad \, + r_b (x_b)
\\
&= \mathrm{Prox}_{r_b} \!\! \left( \acute{x}_b^{(i)} - \left( M_b^{(i)} \right)^{-1}  \nabla f_b^{(i)} (\acute{x}_b^{(i)}) ;  M_b^{(i)} \right), \nn
}
\endgroup
where
\eas{
\acute{x}_b^{(i)} &= x_{b}^{(i)} + W_b^{(i)} \left( x_b^{(i)} - x_b^{(i-1)} \right),
\\
\cX_b^{(i)} &= \cX_b ( x_1^{(i+1)}, \ldots, x_{b-1}^{(i+1)}, x_{b+1}^{(i)}, \ldots, x_{B}^{(i)} ),
}
$\nabla f_b^{(i)} (\acute{x}_b^{(i)})$ is the block-partial gradient of $f$ at $\acute{x}_b^{(i)}$,
$M_b^{(i)} \in \bbR^{n_b \times n_b}$ is a symmetric positive definite majorization matrix for $\nabla^2 f_b^{(i)}(x_b)$, and the proximal operator is defined by
\bes{
\mathrm{Prox}_r (y; M) := \argmin_{x} \, \frac{1}{2} \nm{x - y}_M^2 + r(x).
}
The $\bbR^{n_b \times n_b}$ matrix $W_b^{(i)} \succeq 0$, upper bounded by \R{eq:p:BPG_sq_sum:extraM} below, is an \textit{extrapolation matrix} that significantly accelerates convergence, in a similar manner to the extrapolation weight introduced in \cite{Xu&Yin:13SIAM}. Algorithm \ref{alg:BPG} summarizes these updates.

\begin{algorithm}[pt]
\caption{Block proximal gradient method using a majorizer $\{ M_b : b = 1,\ldots,B\}$ (BPG-M)}
\label{alg:BPG}

\begin{algorithmic}
\REQUIRE $\{ x_b^{(1)} = x_b^{(0)} : b = 1,\ldots,B \}$, $i=1$

\WHILE{a stopping criterion is not satisfied}

\FOR{$b = 1,\ldots,B$}

\STATE Calculate $\displaystyle M_b^{(i)}, W_b^{(i)}$ for $f_b^{(i)}(x_b)$ in \R{eq:def:f_block}
\STATE $\displaystyle \acute{x}_b^{(i)} = x_{b}^{(i)} + W_b^{(i)} \left( x_b^{(i)} - x_b^{(i-1)} \right)$
\STATE $\displaystyle x_b^{(i+1)} = \mathrm{Prox}_{r_b} \!\! \left( \acute{x}_b^{(i)} - \left( M_b^{(i)} \right)^{-1} \nabla f_b^{(i)} (\acute{x}_b^{(i)}); M_b^{(i)} \right)$

\ENDFOR

\STATE  $i = i+1$

\ENDWHILE

\end{algorithmic}
\end{algorithm}

\subsection{BPG-M -- Convergence Analysis} \label{sec:BPG_convg}

This section analyzes the convergence of Algorithm \ref{alg:BPG} under the following assumptions.
\begin{itemize}
\item[] {\em Assumption 1)} $F$ in \R{eq:multiConvx} is continuous in $\dom(F)$ and $\inf_{x \in \dom(F)} F(x) > -\infty$, and \R{eq:multiConvx} has a Nash point (see Definition \ref{d:Nash}).

\item[] {\em Assumption 2)} The majorization matrix $M_b^{(i)}$ obeys $ \beta I \preceq M_b^{(i)} \preceq M_b$ with $\beta > 0$ and a nonsingular matrix $M_b$, and 
\ea{
\label{eq:assume:update}
f_b^{(i)} (x_b^{(i+1)}) 
& \leq f_b^{(i)} (\acute{x}_b^{(i)}) + \ip{ \nabla f_b^{(i)} (\acute{x}_b^{(i)}) }{ x_b^{(i+1)} - \acute{x}_b^{(i)} }  \nn
\\
& \quad + \frac{1}{2} \nm{ x_b^{(i+1)} - \acute{x}_b^{(i)} }_{M_b^{(i+1)}}^2.
}

\item[] {\em Assumption 3)} The majorization matrices $M_b^{(i)}$ and extrapolation matrices $W_b^{(i)}$ are diagonalized by the same basis, $\forall i$.
\end{itemize}

The CDL problem \R{eq:sys:synth} or \R{eq:sys:synthF} straightforwardly satisfies the continuity and the lower-boundedness of $F$ in Assumption 1.
To show this, consider that \textit{1)} the sequence $\{ d_k^{(i+1)} \}$ is in the bounded set $\cD = \{ d_k : \| d_k \|_2^2 \leq 1, k = 1,\ldots,K  \}$; \textit{2)} the positive regularization parameter $\alpha$ ensures that the sequence $\{ z_{l,k}^{(i+1)} \}$ (or $\{ \tilde{z}_{l,k}^{(i+1)} \}$) is bounded (otherwise the cost would diverge). This applies to both the two-block and the multi-block BPG-M frameworks; see Section~\ref{sec:BPGM-twoBlk} and \ref{sec:BPGM-multiBlk}, respectively.
Note that one must carefully design $M_b^{(i+1)}$ to ensure that Assumption 2 is satisfied; Sections~\ref{sec:MajDesign:filter} and \ref{sec:MajDesign:spCd} describe our designs for CDL.
Using a tighter majorization matrix $M_b^{(i)}$ to approximate $\nabla^2 f_b^{(i)}$ is expected to accelerate the algorithm \cite[Lem.~1]{Fessler&etal:93TNS}. 
Some examples that satisfy Assumption 3 include diagonal and circulant matrices (that are decomposed by canonical and Fourier basis, respectively).
Assumptions 1--2 guarantee sufficient decrease of the objective function values.

We now recall the definition of a Nash point (or block coordinate-wise minimizer):
\defn{[A Nash point \mbox{\cite[(2.3)--(2.4)]{Xu&Yin:13SIAM}}] \label{d:Nash}
A Nash point (or block coordinate-wise minimizer) $\bar{x}$ is a point satisfying the Nash equilibrium condition.
The Nash equilibrium condition of \R{eq:multiConvx} is
\eas{
&~ F ( \bar{x}_1 ,\ldots, \bar{x}_{b-1}, \bar{x}_b, \bar{x}_{b+1}, \ldots, \bar{x}_B )
\\
& \leq F ( \bar{x}_1 ,\ldots, \bar{x}_{b-1}, x_b, \bar{x}_{b+1}, \ldots, \bar{x}_B ), \quad \forall x_b \in \bar{\cX}_b, b \in [B],
}
which is equivalent to the following condition:
\ea{
\begin{split}
\label{eq:Nash_EqCond} 
& \ip{ \nabla_{x_b} f(\bar{x}) + \bar{g}_b }{x_b - \bar{x}_b} \geq 0, 
\\
& \mbox{for all $x_b \in \bar{\cX}_b$ and for some $\bar{g}_b \in \partial r_b (\bar{x}_b)$}, 
\end{split}
}
where $\bar{\cX}_b = \cX_b (\bar{x}_1,\ldots,\bar{x}_{b-1},\bar{x}_{b+1},\ldots,\bar{x}_B)$ and $\partial r(x_b)$ is the limiting subdifferential (see \cite[\S 1.9]{Kruger:03JMS}, \cite[\S 8]{Rockafellar&Wets:book}) of $r$ at $x_b$. 
}

In general, the Nash equilibrium condition \R{eq:Nash_EqCond} is weaker than the first-order optimality condition.
For problem \R{eq:multiConvx}, a Nash point is not necessarily a critical point, but a critical point must be a Nash point \cite[Rem.~2.2]{Xu&Yin:13SIAM}.\footnote{Given a feasible set $\cX$, a point $\bar{x} \in \dom(f) \cup \cX$ is a critical point (or stationary point) of $f$ if $f'(\bar{x};d) \geq 0$ for any feasible direction $d$ at $\bar{x}$, where $f'(\bar{x};d)$ denotes directional derivate ($f'(x;d) = d^T \nabla f(x)$ for differentiable $f$).
If $x$ is an interior point of $\cX$, then the condition is equivalent to $0 \in \partial F(\bar{x})$.} This property is particularly useful to show convergence of limit points to a critical point, if one exists; see Remark \ref{r:global_convg}.

\prop{[Square summability of $\|  x^{(i+1)} - x^{(i)} \|_2$]
\label{p:BPG_sq_sum}
Under Assumptions 1--3, let $\{ x^{(i+1)} \}$ be the sequence generated by Algorithm \ref{alg:BPG} with 
\be{
\label{eq:p:BPG_sq_sum:extraM}
0 \preceq W_b^{(i)} \preceq \delta \left( M_b^{(i)} \right)^{-1/2} \left( M_b^{(i-1)} \right)^{1/2}
} 
for $\delta < 1$ for all $b=1,\ldots,B$ and $i$. Then 
\bes{
\sum_{i=1}^{\infty} \nm{x^{(i+1)} - x^{(i)}}_2^2 < \infty.
}
}
\prf{\renewcommand{\qedsymbol}{}
See Section~\ref{sec:prf:p:BPG_sq_sum} of the supplementary material.
}

Proposition \ref{p:BPG_sq_sum} implies that
\be{
\label{eq:p:BPG_sq_sum:imply}
\left\| x^{(i+1)} - x^{(i)} \right\|_2^2 \rightarrow 0.
}

\thm{[A limit point is a Nash point]
\label{t:Nash_convg}
If the assumptions in Proposition \ref{p:BPG_sq_sum} hold, then any limit point of $\{ x^{(i)} \}$ is a Nash point, i.e., it satisfies \R{eq:Nash_EqCond}.
}
\prf{\renewcommand{\qedsymbol}{}
See Section~\ref{sec:prf:t:Nash_convg} of the supplementary material.
}

\rem{
\label{r:global_convg}
Theorem \ref{t:Nash_convg} implies that, if there exists a stationary point for \R{eq:multiConvx}, then any limit point of $\{ x^{(i)} \}$ is a stationary point.
One can further show global convergence under some conditions: if $\{ x^{(i)} \}$ is bounded and the stationary points are isolated, then $\{ x^{(i)} \}$ converges to a stationary point \cite[Cor.~2.4]{Xu&Yin:13SIAM}.\footnote{Due to the difficulty of checking the isolation condition, Xu \& Yin in \cite{Xu&Yin:13SIAM} introduced a better tool to show global convergence based on Kurdyka-{\L}ojasiewicz property.}
}

We summarize some important properties of the proposed BPG-M in CDL:
\begin{summary}
\label{s:BPG-M}
\normalfont
The proposed BPG-M approach exploits a majorization matrix rather than using a Lipschitz constant; therefore, it can be practically applied to CDL without any parameter tuning process (except the regularization parameter). The BPG-M guarantees the local convergence in \R{eq:sys:synth} or \R{eq:sys:synthF}, i.e., if there exists a critical point, any limit point of the BPG-M sequence is a critical point (it also guarantees the global convergence if some further conditions are satisfied; see Remark \ref{r:global_convg} for details). Note that this is the first convergence guarantee in CDL.
The convergence rate of the BPG-M method depends on the tightness of the majorization matrix in \R{eq:quadApprox}; see, for example, Fig.~\ref{fig:Comp_MConvg_BPG}. 
The next section describes variants of BPG-M that further accelerate its convergence.
\end{summary}

\subsection{Restarting Fast BPG-M} \label{sec:FBPG}

This section proposes a technique to accelerate BPG-M. 
By including \textit{1)} a momentum coefficient formula similar to those used in FPG methods \cite{Beck&Teboulle:09SIAM, Nesterov:07CORE, Tseng:08techRep}, and \textit{2)} an adaptive momentum restarting scheme \cite{ODonoghue&Candes:15FCM, Giselsson&Boyd:14CDC}, this section focuses on computationally efficient majorization matrices, e.g., diagonal or circulant majorization matrices.

Similar to \cite{Xu&Yin:16IPI}, we apply some increasing momentum-coefficient formulas $w^{(i)}$ to the extrapolation matrix updates $W_b^{(i)}$ in Algorithm \ref{alg:BPG}:
\begingroup
\setlength{\thinmuskip}{1.5mu}
\setlength{\medmuskip}{2mu plus 1mu minus 2mu}
\setlength{\thickmuskip}{2.5mu plus 2.5mu}
\ea{
& w^{(i)} = \frac{\theta^{(i-1)}  - 1}{\theta^{(i)}}, \quad \theta^{(i)} = \frac{1 + \sqrt{1 + 4 (\theta^{(i-1)})^2}}{2}, \quad \mathrm{or} \label{eq:step_size1}
\\
& w^{(i)} = \frac{\theta^{(i-1)}  - 1}{\theta^{(i)}}, \quad \theta^{(i)} = \frac{i+2}{2}. \label{eq:step_size2}
}
\endgroup
These choices guarantee fast convergence of FPG in \cite{Tseng:08techRep, Beck&Teboulle:09SIAM}. The momentum coefficient update rule in \R{eq:step_size1} was applied to block coordinate updates in \cite{Xu&Yin:14arXiv, Xu&Yin:13SIAM}.
For diagonal majorization matrices $M_b^{(i)}$, $M_b^{(i-1)}$, the extrapolation matrix update is given by
\be{
\label{eq:extraM_update_diag}
\left( W_b^{(i)} \right)_{j,j} = \delta \cdot \min \! \left\{ w^{(i)}, \left( \left( M_b^{(i)} \right)^{-1} M_b^{(i-1)} \right)_{j,j}^{1/2} \right\},
}
where $\delta<1$ appeared in \R{eq:p:BPG_sq_sum:extraM}, for $j=1,\ldots,n$.
(Alternatively, $( W_b^{(i)} )_{j,j} =  \min \{ w^{(i)}, \delta ( ( M_b^{(i)} )^{-1} M_b^{(i-1)} )_{j,j}^{1/2} \}$.)
For circulant majorization matrices $M_b^{(i)} = \Phi^H_{n_b} \diag(\hat{m}_b^{(i)}) \Phi_{n_b}$, $M_b^{(i-1)} = \Phi^H_{n_b} \diag(\hat{m}_b^{(i-1)}) \Phi_{n_b}$, we have the extrapolation matrix updates as follows:
\be{
\label{eq:extraM_update_circ}
W_b^{(i)}  = \left( \Phi^H_{n_b} \right)^{1/2} \widehat{W}_b^{(i)} \Phi^{1/2}_{n_b}
}
where $\Phi_{n_b}$ is a unitary DFT matrix of size $n_b \times n_b$ and $\widehat{W}_b^{(i)} \in \bbR^{n_b \times n_b}$ is a diagonal matrix with entries
\bes{
\left( \widehat{W}_b^{(i)} \right)_{j,j} = \delta \cdot \min \! \left\{ w^{(i)}, \left( \left( \hat{m}_{b,j}^{(i)} \right)^{-1} \hat{m}_{b,j}^{(i-1)} \right)^{1/2} \right\},
}
for $j=1,\ldots,n$.
We refer to BPG-M combined with the modified extrapolation matrix updates \R{eq:extraM_update_diag}--\R{eq:extraM_update_circ} using momentum coefficient formulas \R{eq:step_size1}--\R{eq:step_size2} as \textit{Fast BPG-M} (FBPG-M).
Note that convergence of FBPG-M is guaranteed because \R{eq:p:BPG_sq_sum:extraM} in Proposition \ref{p:BPG_sq_sum} still holds.

\begin{algorithm}[pt!]
\caption{Restarting fast block proximal gradient using a diagonal majorizer $\{ M_b : b = 1,\ldots,B\}$ and gradient-mapping scheme (reG-FBPG-M) } 
\label{alg:FBPG}

\begin{algorithmic}
\REQUIRE $\{ x_b^{(1)} = x_b^{(0)} : b = 1,\ldots,B \}$, $\theta^{(1)} = \theta^{(0)} = 1$, $\delta \in [0,1), \omega \in [-1,0]$, $i=1$

\WHILE{a stopping criterion is not satisfied}

\STATE Update $\displaystyle w^{(i)}$ using either \R{eq:step_size1} or \R{eq:step_size2}

\FOR{$b = 1,\ldots,B$}

\STATE Calculate $\displaystyle M_b^{(i)}$ for $f_b^{(i)}(x_b)$ in \R{eq:def:f_block}
\STATE Calculate $\displaystyle  W_b^{(i)}$ by \R{eq:extraM_update_diag} with $\displaystyle M_b^{(i)}, M_b^{(i-1)}, w^{(i)}$
\STATE $\displaystyle \acute{x}_b^{(i)} = x_{b}^{(i)} + W_b^{(i)} \left( x_b^{(i)} - x_b^{(i-1)} \right)$
\STATE $\displaystyle x_b^{(i+1)} = \mathrm{Prox}_{r_b} \!\! \left( \acute{x}_b^{(i)} - \left( M_b^{(i)} \right)^{-1} \nabla f_b^{(i)} (\acute{x}_b^{(i)}); M_b^{(i)} \right)$

\IF{{\thickmuskip=0.5\thickmuskip $\cos \! \left( \Theta \! \left( M_b^{(i)} \!\! \left( \acute{x}_b^{(i)} - x_b^{(i+1)} \right),  x_b^{(i+1)} - x_b^{(i)} \right) \right) > \omega$}}

\STATE $\displaystyle \acute{x}_b^{(i)} = x_{b}^{(i)}$
\STATE \hspace{-0.5em}$\displaystyle x_b^{(i+1)} = \mathrm{Prox}_{r_b} \!\! \left( \! \acute{x}_b^{(i)} - \left( M_b^{(i)} \right)^{-1} \! \nabla f_b^{(i)} (\acute{x}_b^{(i)}); M_b^{(i)} \! \right)$

\ENDIF

\ENDFOR

\STATE $i = i+1$

\ENDWHILE

\end{algorithmic}

\end{algorithm}

To further accelerate FBPG-M, we apply the adaptive momentum restarting scheme introduced in \cite{ODonoghue&Candes:15FCM, Giselsson&Boyd:14CDC}.
This technique restarts the algorithm by resetting the momentum back to zero and taking the current iteration as the new starting point, when a restarting criterion is satisfied.
The \textit{non-monotonicity} restarting scheme (referred to \textit{reO}) can be used to make whole objective non-increasing \cite{ODonoghue&Candes:15FCM, Giselsson&Boyd:14CDC,Xu&Yin:16IPI}. The restarting criterion for this method is given by
\ea{
\label{eq:restart:func}
\begin{split}
&~ F( x_1^{(i+1)}, \ldots, x_{b-1}^{(i+1)}, x_b^{(i+1)}, x_{b+1}^{(i)}, \ldots, x_B^{(i)} )  
\\
& > F( x_1^{(i+1)}, \ldots, x_{b-1}^{(i+1)}, x_b^{(i)}, x_{b+1}^{(i)}, \ldots, x_B^{(i)} ).
\end{split}
}
However, evaluating the objective in each iteration is computationally expensive and can become an overhead as one increases the number of filters and the size of training datasets.
Therefore, we introduce a \textit{gradient-mapping} scheme (referred to \textit{reG}) that restarts the algorithm when the following criterion is met:
\be{
\label{eq:restart:grad}
\cos \! \left( \Theta \! \left( M_b^{(i)} \!\! \left( \acute{x}_b^{(i)} - x_b^{(i+1)} \right),  x_b^{(i+1)} - x_b^{(i)} \right) \right) > \omega,
}
where the angle between two nonzero real vectors $\vartheta$ and $\vartheta'$ is
\bes{
\Theta (\vartheta, \vartheta') := \frac{ \ip{\vartheta}{\vartheta'} }{\nm{\vartheta}_2 \nm{\vartheta'}_2},
}
and $\omega \in [-1, 0]$.
The gradient-mapping scheme restarts the algorithm whenever the momentum, i.e., $x_b^{(i+1)} - x_b^{(i)}$, is likely to lead the algorithm in a bad direction, as measured by the gradient mapping (which is a generalization of the gradient, i.e.,  $M_b^{(i)} ( \acute{x}_b^{(i)} - x_b^{(i+1)} )$) at the $x_b^{(i+1)}$-update.
The gradient-mapping criterion \R{eq:restart:grad} is a relaxed version of the gradient-based restarting technique introduced in \cite{ODonoghue&Candes:15FCM, Giselsson&Boyd:14CDC}. Compared to those in \cite{ODonoghue&Candes:15FCM, Giselsson&Boyd:14CDC}, the relaxed criterion often provides a faster convergence at the early iterations in practice \cite{Muckley&Noll&Fessler:15TMI}.

To solve the multi-convex optimization problem \R{eq:sys:synthF}, we apply Algorithm \ref{alg:FBPG}, promoting stable and fast convergence. 
We minimize \R{eq:sys:synthF} by the proposed BPG-M using the two-block and multi-block schemes; see Section~\ref{sec:BPGM-twoBlk} and Section~\ref{sec:BPGM-multiBlk}, respectively---each section presents efficiently computable separable majorizers and introduces efficient proximal mapping methods.

\section{Convergent CDL: FBPG-M with Two-Block Update} \label{sec:BPGM-twoBlk}

Based on the FBPG-M method in the previous section, we first solve \R{eq:sys:synthF} by the two-block scheme, i.e., similar to the AL methods, we alternatively update filters $\{ d_{k} : k = 1,\ldots,K \}$ and sparse codes $\{ \tilde{z}_{l} : l = 1,\ldots,L \}$.
The two-block scheme is particularly useful with parallel computing, because proximal mapping problems are separable (see Sections~\ref{sec:prox:filter} and \ref{sec:prox:spCd}) and some majorization matrices computations are parallelizable.

\subsection{Dictionary (Filter) Update} \label{sec:BPGM:filter}

\subsubsection{Separable Majorizer Design} \label{sec:MajDesign:filter}

Using the current estimates of the $\{ \tilde{z}_l : l = 1,\ldots,L \}$, the filter update problem for \R{eq:sys:synthF} is given by
\eas{
\begin{split}
\min_{\{ d_k \}} &~ \frac{1}{2} \sum_{l=1}^L  \left\| y_l - P_B \left[ \Phi^{-1} \diag(\Phi P_S^T d_{1}) \Phi \right. \right.
\\
&~ \left. \left. \cdots~ \Phi^{-1} \diag(\Phi P_S^T d_{K}) \Phi \right] \tilde{z}_l  \right\|_2^2
\\
\mbox{s.t.} ~&~ \| d_k \|_2^2 \leq 1, \qquad k = 1,\ldots, K,
\end{split}
}
which can be rewritten as follows:
\ea{
\label{eq:sys:BPGM:synthF_kernel}
\min_{\{ d_k \}} &~ \frac{1}{2} \left\| \left[ \begin{array}{c} y_1 \\ \vdots \\ y_L \end{array} \right] - \Psi \left[ \begin{array}{c} d_1 \\ \vdots \\ d_K \end{array} \right] \right\|_2^2 
\\ 
\mbox{s.t.} ~&~ \| d_k \|_2^2 \leq 1, \qquad k = 1,\ldots, K, \nn
}
where 
\ea{
\begin{split}
\label{eq:sysMat:BPGM:synthF_kernel}
\Psi & := \left( I_L \otimes  P_B \Phi^{-1} \right) \widehat{Z} \left( I_K \otimes \Phi P_S^T \right),
\end{split}
\\
\begin{split}
\label{eq:def:Ztilde}
\widehat{Z} & := \left[ \arraycolsep=3pt \begin{array}{ccc} \diag(\hat{z}_{1,1}) & \cdots & \diag(\hat{z}_{1,K}) \\ \vdots & \ddots & \vdots \\ \diag(\hat{z}_{L,1}) & \cdots & \diag(\hat{z}_{L,K}) \end{array} \right].
\end{split}
}
and $\{ \hat{z}_{l,k} = \Phi \tilde{z}_{l,k} : l=1,\ldots,L, k=1,\ldots,K  \}$. We now design block separable majorizer for the Hessian matrix $\Psi^H \Psi \in \bbR^{KD \times KD}$ of the cost function in \R{eq:sys:BPGM:synthF_kernel}.
Using $\Phi^{-H} P_B^T P_B \Phi^{-1} \!\preceq\! \tilde{N}^{-1} I$ and $\Phi^H \!=\! \tilde{N} \Phi^{-1}$, $\Psi^H \Psi$ is bounded by
\ea{
\label{eq:H_bound1}
\Psi^H \Psi & \preceq \left( I_K \otimes P_S \Phi^{-1} \right) \widehat{Z}^H \widehat{Z} \left( I_K \otimes \Phi P_S^T \right) \nn
\\
& = \left( I_K \otimes P_S \right) Q_{\Psi}^H Q_{\Psi} \left( I_K \otimes P_S^T \right)
}
where $\widehat{Z}^H \widehat{Z}$ is given according to \R{eq:def:Ztilde}, $Q_{\Psi}^H Q_{\Psi} \in \bbC^{\tilde{N}K \times \tilde{N}K}$ is a block matrix with submatrices $\{ [ Q_{\Psi}^H Q_{\Psi} ]_{k,k'} \in \bbC^{\tilde{N} \times \tilde{N}} :  k, k' = 1,\ldots,K \}$:
\be{
\label{eq:def:QH}
[ Q_{\Psi}^H Q_{\Psi} ]_{k,k'} :=  \Phi^{-1} \sum_{l=1}^L \diag( \hat{z}_{l,k}^* \odot \hat{z}_{l,k'} ) \Phi.
}

Based on the bound \R{eq:H_bound1}, our first diagonal majorization matrix for $\Psi^H \Psi$ is given as follows:

\lem{[Block diagonal majorization matrix $M_{\Psi}$ with diagonals \Romnum{1}] \label{l:Kernel:diag1}
The following block diagonal matrix $M_{\Psi} \in \bbR^{KD \times KD}$ with diagonal blocks satisfies $M_{\Psi} \succeq \Psi^H \Psi$:
\bes{
M_{\Psi}  = \diag \! \left( \left( I_K \otimes P_S \right) | Q_{\Psi}^H Q_{\Psi} | \left( I_K \otimes P_S^T \right) 1_{KD} \right),
}
where $Q_{\Psi}^H Q_{\Psi}$ is defined in \R{eq:def:QH} and $|A|$ denotes the matrix consisting of the absolute values of the elements of $A$.
}
\prf{\renewcommand{\qedsymbol}{}
See Section~\ref{sec:prf:l:Kernel-spCD:diag1} of the supplementary material.
}

We compute $|Q_{\Psi}^H Q_{\Psi}|$ by taking the absolute values of elements of the first row (or column) of each circulant submatrix $[ Q_{\Psi}^H Q_{\Psi} ]_{k,k'}$ for $k,k'=1,\ldots,K$. Throughout the paper, we apply this simple trick to efficiently compute the element-wise absolute value of the circulant matrices (because circulant matrices can be fully specified by a single vector).
The computational complexity for the majorization matrix in Lemma \ref{l:Kernel:diag1} involves $\cO(K^2 L \tilde{N})$ operations for $\widehat{Z}^H \widehat{Z}$ and approximately $\cO(K^2 \tilde{N} \log \tilde{N})$ operations for $Q_{\Psi}^H Q_{\Psi}$.
The permutation trick for a block matrix with diagonal blocks in \cite{Bristow&etal:13CVPR} (see details in \cite[Rem.~3]{Chun&etal:15TCI}) allows parallel computation of $\widehat{Z}^H \widehat{Z}$ over $j=1,\ldots,\tilde{N}$, i.e., each thread requires $\cO(K^2 L)$ operations.
Using Proposition \ref{p:Kernel:MajorQ} below, we can substantially reduce the latter number of operations at the cost of looser bounds (i.e., slower convergence).

\prop{
\label{p:Kernel:MajorQ}
The following block diagonal matrix $M_{Q_{\Psi}} \in \bbR^{\tilde{N}K \times \tilde{N}K}$ satisfies $M_{Q_{\Psi}} \succeq Q_{\Psi}^H Q_{\Psi}$:
\ea{
M_{Q_{\Psi}} &= \bigoplus_{k=1}^K \Phi^{-1} \Sigma_k \Phi, \label{eq:MQ}
\\
\Sigma_k &=  \sum_{l=1}^L \diag( \left| \hat{z}_{l,k} \right|^2 ) + \sum_{k' \neq k} \left| \sum_{l=1}^L  \diag( \hat{z}_{l,k}^* \odot \hat{z}_{l,k'} ) \ \right|, \label{eq:Sigmak}
}
for $k=1,\ldots,K$.
}
\prf{\renewcommand{\qedsymbol}{}
See Section~\ref{sec:prf:p:Kernel:MajorQ} of the supplementary material.
}

We now substitute \R{eq:MQ} into \R{eq:H_bound1}.
Unfortunately, the resulting $KD \times KD$ block-diagonal matrix below is inconvenient for inverting:
\bes{
\Psi^H \Psi \preceq  
\left[ \arraycolsep=2pt \begin{array}{ccc} P_S \Phi^{-1} \Sigma_1 \Phi P_S^T & & \\ & \ddots & \\ & & P_S \Phi^{-1} \Sigma_K \Phi P_S^T \end{array} \right].
}
Using some bounds for the block diagonal matrix intertwined with $P_S$ and $P_S^T$ above, the following two lemmas propose two separable majorization matrices for $\Psi^H \Psi$.

\lem{[Block diagonal majorization matrix $M_{\Psi}$ with scaled identities] \label{l:Kernel:identity}
The following block diagonal matrix $M_{\Psi} \in \bbR^{KD \times KD}$ with scaled identity blocks satisfies $M_{\Psi} \succeq \Psi^H \Psi$:
\eas{
M_{\Psi} &= \bigoplus_{k=1}^K [M_{\Psi}]_{k,k},
\\
[M_{\Psi}]_{k,k} & = \max_{j=1,\ldots,\tilde{N}} \left\{ ( \Sigma_k )_{j,j} \right\} \cdot I_K, \quad k \in [K],
}
where diagonal matrices $\{ \Sigma_k \}$ are as in \R{eq:Sigmak}.
}
\prf{\renewcommand{\qedsymbol}{}
See Section~\ref{sec:prf:l:Kernel:identity} of the supplementary material.
}

\lem{[Block diagonal majorization matrix $M_{\Psi}$ with diagonals \Romnum{2}] \label{l:Kernel:diag2}
The following block diagonal matrix $M_{\Psi} \in \bbR^{KD \times KD}$ with diagonal blocks satisfies $M_{\Psi} \succeq \Psi^H \Psi$:
\eas{
M_{\Psi} &= \bigoplus_{k=1}^K [ M_{\Psi} ]_{k,k},
\\
[ M_{\Psi} ]_{k,k} &= \diag \big( P_S \left| \Phi^{-1} \Sigma_k \Phi \right| P_S^T 1_{K} \big), \quad k \in [K],
}
where diagonal matrices $\{ \Sigma_k \}$ are as in \R{eq:Sigmak}.
}
\prf{\renewcommand{\qedsymbol}{}
See Section~\ref{sec:prf:l:Kernel-spCd:diag2} of the supplementary material.
}

\begin{figure}[!t]
\centering
\begin{tabular}{c}
\includegraphics[scale=0.5, trim=0 0.2em 2.2em 1em, clip]{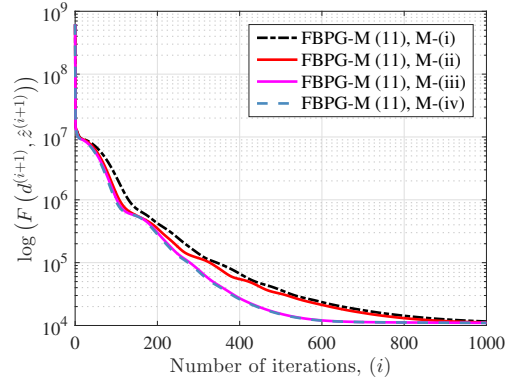} \\
\specialcell[c]{\small (a) Cost minimization with FBPG-M \R{eq:step_size1} \vspace{-0.25em} \\ \small using different majorizers} \\
\includegraphics[scale=0.5, trim=0 0.2em 2.2em 1em, clip]{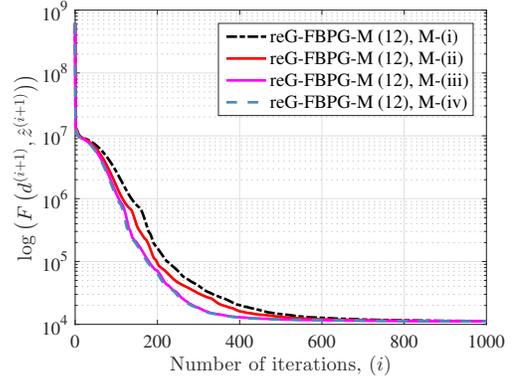}  \\
\specialcell[c]{\small (b) Cost minimization with reG-FBPG-M \R{eq:step_size2} \vspace{-0.25em} \\ \small using different majorizers} \\
\end{tabular}

\caption{Cost minimization behavior for different majorizer designs (the fruit dataset). 
As the majorizer changes from M-(\romnum{1}) to M-(\romnum{4}), we expect to have a tighter majorizer of the Hessian. (See Table \ref{tab:name_BPG-M_Mtype} for details of majorization matrix design.)
As expected, tighter majorizers lead to faster convergence.
}
\label{fig:Comp_MConvg_BPG}
\end{figure}

The majorization matrix designs in Lemma \ref{l:Kernel:identity} and \ref{l:Kernel:diag2} reduce the number of operations $\cO (K^2 \tilde{N} \log \tilde{N})$ to $\cO (K \tilde{N})$ and $\cO (K \tilde{N} \log \tilde{N})$, respectively. If parallel computing is applied over $k=1,\ldots,K$, each thread requires $\cO (\tilde{N})$ and $\cO (\tilde{N} \log N)$ operations for Lemma \ref{l:Kernel:identity} and \ref{l:Kernel:diag2}, respectively.
However, the majorization matrix in Lemma \ref{l:Kernel:diag1} is tighter than those in Lemma \ref{l:Kernel:identity}--\ref{l:Kernel:diag2} because those in Lemma \ref{l:Kernel:identity}--\ref{l:Kernel:diag2} are designed based on another bound. Fig.~\ref{fig:Comp_MConvg_BPG} verifies that the tighter majorizer leads to faster convergence.
Table~\ref{tab:comput_Mtype} summarizes these results.

\subsubsection{Proximal Mapping} \label{sec:prox:filter}

Because all of our majorization matrices are block diagonal, using \R{eq:quad_surrg} the proximal mapping problem \R{eq:sys:BPGM:synthF_kernel} simplifies to separate problems for each filter:
\ea{
\begin{split}
\label{eq:prox_map:blk:BPGM:synthF_kernel}
d_k^{(i+1)} = \argmin_{d_k} &\, \frac{1}{2} \nm{ d_k - \nu_k^{(i)} }_{\left[ M_{\Psi}^{(i)} \right]_{k,k}}^2 
\\
\mbox{s.t.} ~~\,&\, \nm{d_k}_2^2 \leq 1, \qquad k = 1,\ldots,K,
\end{split}
}
where
\bes{
\nu^{(i)} = \acute{d}^{(i)} - \left( M_{\Psi}^{(i)} \right)^{-1} \left( \Psi^{(i)} \right)^H \left( \Psi^{(i)} \acute{d}^{(i)} - y \right),
}
we construct $\Psi^{(i)}$ using \R{eq:sysMat:BPGM:synthF_kernel} with updated sparse codes $\{ \tilde{z}_{l}^{(i)} : l=1,\ldots,L \}$, $M_{\Psi}^{(i)}$ is a designed block diagonal majorization matrix for $( \Psi^{(i)} )^H \Psi^{(i)}$, $y$ is a concatenated vector with $\{ y_l \}$, and $\nu^{(i)}$ is a concatenated vector with $\{ \nu_{k}^{(i)} \in \bbR^K : k = 1,\ldots,K \}$. 
When $\{ [ M_{\Psi}^{(i)} ]_{k,k} \}$ is a scaled identity matrix (i.e., Lemma \ref{l:Kernel:identity}), the optimal solution is simply the projection of $\nu_k^{(i)}$ onto the $\ell^2$ unit ball.
If $\{ [ M_{\Psi}^{(i)} ]_{k,k} \}$ is a diagonal matrix (Lemma \ref{l:Kernel:diag1} and \ref{l:Kernel:diag2}), the proximal mapping requires an iterative scheme.
We apply accelerated Newton's method to efficiently obtain the optimal solution to \R{eq:prox_map:blk:BPGM:synthF_kernel}; see details in Section~\ref{sec:AccNewton}.

\begin{table}[!t]

\caption{Name Conventions for BPG-M Algorithms and Majorization Matrix Designs}
\vspace{-1em}
\label{tab:name_BPG-M_Mtype}
\renewcommand*
\arraystretch{1.1}
\centering

\vspace{-0.1em}
\subtable{
\centering
\begin{tabular}{c|cc}
\hline \hline
{BPG-M names$^\dagger$} & {Momentum coeff.} & {Restarting scheme} \\
\hline \hline
FBPG-M & \R{eq:step_size1} or \R{eq:step_size2} & $\cdot$ \\
reO-BPG-M & $\cdot$ & Objective -- \R{eq:restart:func} \\
reG-BPG-M & $\cdot$ & Gradient -- \R{eq:restart:grad} \\
reG-FBPG-M & \R{eq:step_size1} or \R{eq:step_size2} & Gradient -- \R{eq:restart:grad} \\
\hline \hline
\end{tabular}
}

\subtable{
\centering
\begin{tabular}{c|cc}
\hline \hline
\specialcell[c]{Majorizer \\ names} & \specialcell[c]{Majorization matrix \\ for filter updates} & \specialcell[c]{Majorization matrix \\ for sparse coding} \\
\hline \hline
M-(\romnum{1}), two-block & Lemma \ref{l:Kernel:identity} & Lemma \ref{l:spCd:diag2} \\
M-(\romnum{2}), two-block & Lemma \ref{l:Kernel:diag2} & Lemma \ref{l:spCd:diag2} \\
M-(\romnum{3}), two-block & Lemma \ref{l:Kernel:diag2} & Lemma \ref{l:spCd:diag1} \\
M-(\romnum{4}), two-block & Lemma \ref{l:Kernel:diag1} & Lemma \ref{l:spCd:diag1} \\
\hline
M-(\romnum{5}), multi-block & Lemma \ref{l:Kernel:diag:mltBlk} & Lemma \ref{l:spCd:diag:mltBlk} \\
\hline \hline
\end{tabular}
}

\smallskip
\begin{myquote}{0.1in}
$^\dagger$The non-monotonicity restarting scheme based on the objective is referred to as \textit{reO}.
The gradient-mapping restarting scheme is referred to as \textit{reG}.
\end{myquote}
\end{table}

\subsection{Sparse Code Update} \label{sec:BPGM:spCd}

\subsubsection{Separable Majorizer Design} \label{sec:MajDesign:spCd}

Given the current estimates of the $\{ \lambda_{k} = \Phi P_S^T d_{k}: k = 1,\ldots,K \}$, the sparse code update problem for \R{eq:sys:synthF} becomes $L$ separate optimization problems:
\be{
\label{eq:sys:BPGM:synthF_spCode}
\min_{\tilde{z}_{l}} ~ \frac{1}{2} \left\| y_l - \Gamma \tilde{z}_l  \right\|_2^2 + \alpha \| \tilde{z}_{l} \|_1,
}
for $l=1,\ldots,L$, where
\be{
\label{eq:sysMat:BPGM:synthF_spCode}
\Gamma := P_B \left[ \arraycolsep=2pt \begin{array}{ccc} \Phi^{-1} \diag(\lambda_1) \Phi & \cdots & \Phi^{-1} \diag(\lambda_K) \Phi \end{array} \right].
}
We now seek a block separable majorizer for the Hessian matrix $\Gamma^H \Gamma \in \bbC^{\tilde{N}K \times \tilde{N}K}$ of the quadratic term in \R{eq:sys:BPGM:synthF_spCode}.
Using $\Phi^{-H} P_B^T P_B \Phi^{-1} \!\preceq\! \tilde{N}^{-1} I$ and $\Phi^H \!=\! \tilde{N} \Phi^{-1}$, $\Gamma^H \Gamma$ is bounded as follows:
\be{
\label{eq:G_bound1}
\Gamma^H \Gamma  \preceq \left( I_K \otimes \Phi^{-1} \right) \Lambda^H \Lambda \left( I_K \otimes \Phi \right) = Q_{\Gamma}^H Q_{\Gamma}
}
where $\Lambda^H \Lambda$ is given according to 
\be{
\label{eq:def:Lambda}
\Lambda := \left[ \diag(\lambda_1), \cdots, \diag(\lambda_K) \right],
}
and $Q_{\Gamma}^H Q_{\Gamma} \in \bbC^{\tilde{N}K \times \tilde{N}K}$ is a block matrix with submatrices $\{ [ Q_{\Gamma}^H Q_{\Gamma} ]_{k,k'} \in \bbC^{\tilde{N} \times \tilde{N}} :  k, k' = 1,\ldots,K \}$:
\be{
\label{eq:def:QG}
[ Q_{\Gamma}^H Q_{\Gamma} ]_{k,k'} =  \Phi^{-1} \diag( \lambda_{k}^* \odot \lambda_{k'} ) \Phi.
}

The following lemma describes our first diagonal majorization matrix for $\Gamma^H \Gamma$.

\lem{[Block diagonal majorization matrix $M_{\Gamma}$ with diagonals \Romnum{1}]\label{l:spCd:diag1}
The following block diagonal matrix $M_{\Gamma} \in \bbR^{\tilde{N}K \times \tilde{N}K}$ with diagonal blocks satisfies $M_{\Gamma} \succeq \Gamma^H \Gamma$:
\bes{
M_{\Gamma}  = \diag \! \left(  | Q_{\Gamma}^H Q_{\Gamma} | 1_{\tilde{N}K} \right),
}
where $Q_{\Gamma}^H Q_{\Gamma}$ is defined in \R{eq:def:QG}.
}
\prf{\renewcommand{\qedsymbol}{}
See Section~\ref{sec:prf:l:Kernel-spCD:diag1} of the supplementary material.
}

\begin{table}

\caption{Comparison of Computational Complexity in Computing Different Majorization Matrices}
\vspace{-1em}
\label{tab:comput_Mtype}
\renewcommand*
\arraystretch{1.1}
\centering

\vspace{-0.1em}
\subtable{
\centering
\begin{tabular}{c|l}
\hline \hline
{Majorizer$^\dagger$} & \specialcell[c]{A. Computations for majorization matrix \\ in filter updates$^\ddagger$} \\
\hline \hline
\multirow{2}{*}{M-(\romnum{1}), two-block} & Multi-thread: $\cO(K^2 L) + \cO(\tilde{N})$ \\ & Single-thread: $\cO(K^2 L \tilde{N}) + \cO(K \tilde{N})$ \\
\multirow{2}{*}{M-(\romnum{2}), two-block} & Multi-thread: $\cO(K^2 L) + \cO(\tilde{N} \log \tilde{N})$ \\ & Single-thread: $\cO(K^2 L \tilde{N}) + \cO(K \tilde{N} \log \tilde{N})$ \\
\multirow{2}{*}{M-(\romnum{3}), two-block} & Multi-thread: $\cO(K^2 L) + \cO(\tilde{N} \log \tilde{N})$ \\ & Single-thread: $\cO(K^2 L \tilde{N}) + \cO(K \tilde{N} \log \tilde{N})$ \\
\multirow{2}{*}{M-(\romnum{4}), two-block} & Multi-thread: $\cO(K^2 L) + \cO(K^2 \tilde{N} \log \tilde{N})$ \\ & Single-thread: $\cO(K^2 L \tilde{N}) + \cO(K^2 \tilde{N} \log \tilde{N})$ \\
\hline
M-(\romnum{5}), multi-block & $\cO ( K (L \tilde{N} + \tilde{N} \log \tilde{N}) )$ \\
\hline \hline
\end{tabular}
}

\subtable{
\centering
\begin{tabular}{c|l}
\hline \hline
{Majorizer$^\dagger$} & \specialcell[c]{B. Computations for majorization matrix \\ in sparse code updates$^\ddagger$} \\
\hline \hline
\multirow{2}{*}{M-(\romnum{1}), two-block} & Multi-thread: $\cO(K^2) + \cO(\tilde{N} \log \tilde{N})$ \\ & Single-thread: $\cO(K^2 \tilde{N}) + \cO(K \tilde{N} \log \tilde{N})$ \\
\multirow{2}{*}{M-(\romnum{2}), two-block} & Multi-thread: $\cO(K^2) + \cO(\tilde{N} \log \tilde{N})$ \\ & Single-thread: $\cO(K^2 \tilde{N}) + \cO(K \tilde{N} \log \tilde{N})$ \\
\multirow{2}{*}{M-(\romnum{3}), two-block} & Multi-thread: $\cO(K^2) + \cO(K^2 \tilde{N} \log \tilde{N})$ \\ & Single-thread: $\cO(K^2 \tilde{N}) + \cO(K^2 \tilde{N} \log \tilde{N})$ \\
\multirow{2}{*}{M-(\romnum{4}), two-block} & Multi-thread: $\cO(K^2) + \cO(K^2 \tilde{N} \log \tilde{N})$ \\ & Single-thread: $\cO(K^2 \tilde{N}) + \cO(K^2 \tilde{N} \log \tilde{N})$ \\
\hline
M-(\romnum{5}), multi-block & $\cO(K \tilde{N} D)$ or $\cO( K ( \tilde{N} \log \tilde{N} ) )$ \\
\hline \hline
\end{tabular}
}

\smallskip
\begin{myquote}{0.1in}
$^\dagger$The majorizer name follows the name convention in Table~\ref{tab:name_BPG-M_Mtype}. 
\\
$^\ddagger$For two-block BPG-M, the values in \textit{multi-thread} denote computational costs at each thread when parallel computing is applied.
The computational costs for multi-block BPG-M are estimated when multi-core processing is not used.
\end{myquote}
\end{table}

Computing the majorization matrix in Lemma \ref{l:spCd:diag1} involves $\cO(K^2 \tilde{N})$ operations for $\Lambda^H \Lambda$ and approximately $\cO(K^2 \tilde{N} \log \tilde{N})$ operations for $Q_{\Gamma}^H Q_{\Gamma}$.
Again, applying the permutation trick in \cite{Bristow&etal:13CVPR}, \cite[Rem.~3]{Chun&etal:15TCI} allows computing $\Lambda^H \Lambda$ by parallelization over $j=1,\ldots,\tilde{N}$, i.e., each thread requires $\cO(K^2)$ operations.
Similar to the filter update case, Proposition \ref{p:spcd:MajorQ} below substantially reduces the computational cost $\cO(K^2 \tilde{N} \log \tilde{N})$ at the cost of slower convergence.

\prop{
\label{p:spcd:MajorQ}
The following block diagonal matrix $M_{Q_{\Gamma}} \in \bbR^{\tilde{N}K \times \tilde{N}K}$ satisfies $M_{Q_{\Gamma}} \succeq Q_{\Gamma}^H Q_{\Gamma}$:
\ea{
M_{Q_{\Gamma}} &= \bigoplus_{k=1}^K \Phi^{-1} \Sigma'_k \Phi, \label{eq:MP}
\\
\Sigma'_k &= \diag( \left| \lambda_{k} \right|^2 ) + \sum_{k' \neq k} \left| \diag( \lambda_{k}^* \odot \lambda_{k'} ) \right|, \label{eq:Sigma'k}
}
for $k=1,\ldots,K$.
}
\prf{\renewcommand{\qedsymbol}{}
See Section~\ref{sec:prf:p:spcd:MajorQ} of the supplementary material.
}

\lem{[Block diagonal majorization matrix $M_{\Gamma}$ with diagonals \Romnum{2}] \label{l:spCd:diag2}
The following block diagonal matrix $M_{\Gamma} \in \bbR^{\tilde{N}K \times \tilde{N}K}$ with diagonal blocks satisfies $M_{\Gamma} \succeq \Gamma^H \Gamma$:
\eas{
M_{\Gamma} &= \bigoplus_{k=1}^K [M_{\Gamma}]_{k,k},
\\
[ M_{\Gamma} ]_{k,k} &= \diag \big( \left| \Phi^{-1} \Sigma'_k \Phi \right| 1_{\tilde{N}} \big), \quad k \in [K],
}
where diagonal matrices $\{ \Sigma'_k \}$ are as in \R{eq:Sigma'k}.
}
\prf{\renewcommand{\qedsymbol}{}
See Section~\ref{sec:prf:l:Kernel-spCd:diag2} of the supplementary material.
}

The majorization matrix in Lemma \ref{l:spCd:diag2} reduces the cost $\cO (K^2 \tilde{N} \log \tilde{N})$ (of computing that in Lemma~\ref{l:spCd:diag1}) to $\cO (K \tilde{N} \log \tilde{N})$. 
Parallelization can further reduce computational complexity to $\cO (\tilde{N} \log \tilde{N})$.
However, similar to the majorizer designs in the filter update, the majorization matrix in Lemma \ref{l:spCd:diag1} is expected to be tighter than those in Lemma \ref{l:spCd:diag2} because the majorization matrix in Lemma \ref{l:Kernel:diag2} is designed based on another bound. Fig.~\ref{fig:Comp_MConvg_BPG} illustrates that tighter majorizers lead to faster convergence. 
Table~\ref{tab:comput_Mtype} summarizes these results.

\subsubsection{Proximal Mapping} \label{sec:prox:spCd}

Using \R{eq:quad_surrg}, the corresponding proximal mapping problem of \R{eq:sys:BPGM:synthF_spCode} is given by:
\be{
\label{eq:prox_map:BPGM:synthF_spCode}
\tilde{z}_{l}^{(i+1)} = \argmin_{ \tilde{z}_l } \, \frac{1}{2} \nm{\tilde{z}_l - \zeta_l^{(i)} }_{M_{\Gamma}^{(i)}}^2 + \alpha \| \tilde{z}_{l} \|_1
}
where
\bes{
\zeta_l^{(i)} = \acute{z}_l^{(i)} - \left( M_{\Gamma}^{(i)} \right)^{-1} \left( \Gamma^{(i)} \right)^H \left( \Gamma^{(i)} \acute{z}_l^{(i)} - y_l \right),
}
we construct $\Gamma^{(i)}$ using \R{eq:sysMat:BPGM:synthF_spCode} with updated kernels $\{ d_{k}^{(i+1)} : k=1,\ldots,K \}$, $M_{\Gamma}^{(i)}$ is a designed majorization matrix for $(\Gamma^{(i)})^H \Gamma^{(i)}$, and $\zeta_l^{(i)}$ is a concatenated vector with $\{ \zeta_{l,k}^{(i)} \in \bbR^{\tilde{N}}: k = 1,\ldots, K \}$, for $l=1,\ldots,L$.
Using the circulant majorizer in Proposition \ref{p:spcd:MajorQ} would require an iterative method for proximal mapping.
For computational efficiency in proximal mapping, we focus on diagonal majorizers, i.e., Lemma \ref{l:spCd:diag1} and \ref{l:spCd:diag2}.
Exploiting the structure of diagonal majorization matrices, the solution to \R{eq:prox_map:BPGM:synthF_spCode} is efficiently computed by soft-shrinkage:
\bes{
\left( \tilde{z}_{l,k}^{(i+1)} \right)_{j} = \mathrm{softshrink} \! \left( \left( \zeta_{l,k}^{(i)} \right)_j, \alpha \left( \left[ M_{\Gamma}^{(i)} \right]_{k,k} \right)_{j,j}^{-1} \right),
}
for $k=1,\ldots,K$, $j=1,\ldots,\tilde{N}$, where the soft-shrinkage operator is defined by $\mathrm{softshrink}(a,b) := \sgn(a) \max(|a| - b, 0)$.

Note that one does not need to use $\Gamma^{(i)}$ in \R{eq:sysMat:BPGM:synthF_spCode} (or $(\Gamma^{(i)})^H$) directly.
If the filter size $D$ is smaller than $\log \tilde{N}$, it is more efficient to use (circular) convolutions, by considering that the computational complexities for $d_k \circledast z_{l,k}$ and $\Phi^{-1} \diag(\lambda_{k}) \Phi \tilde{z}_{l,k}$ are $\cO (\tilde{N} D)$ and $\cO (\tilde{N} \log \tilde{N})$, respectively. This scheme analogously applies to $\Psi^{(i)}$ in \R{eq:sysMat:BPGM:synthF_kernel} in the filter update.

\section{Accelerated Convergent CDL: FBPG-M with Multi-Block Update} \label{sec:BPGM-multiBlk}

This section establishes a multi-block BPG-M framework for CDL that is particularly useful for single-thread computation mainly due to \textit{1)} more efficient majorization matrix computations and \textit{2)} (possibly) tighter majorizer designs than those in the two-block methods.
In single-thread computing, it is desired to reduce the computational cost for majorizers, by noting that without parallel computing, the computational cost---but disregarding majorizer computation costs---in the two-block scheme is $\cO (K L ( \tilde{N} \log \tilde{N} + D/L + \tilde{N} ) )$ and identical to that of the multi-block approach.
While guaranteeing convergence, the multi-block BPG-M approach accelerates the convergence rate of the two-block BPG-M methods in the previous section, with a possible reason that the majorizers of the multi-block scheme are tighter than those of the two-block scheme.

We update $2 \cdot K$ blocks sequentially; at the $k\rth$ block, we sequentially update the $k\rth$ filter---$d_k$---and the set of $k\rth$ sparse codes for each training image---$\{ z_{l,k} : l=1,\ldots,L \}$ (referred to the \textit{$k\rth$ sparse code set}).
One could alternatively randomly shuffle the $K$ blocks at the beginning of each cycle \cite{Richtarik&Takac:14MP} to further accelerate convergence.
The mathematical decomposing trick used in this section (specifically, \R{eq:sys:BPGM:synthF_kernel:decoup} and \R{eq:sys:BPGM:synthF_spCode:decoup}) generalizes a sum of outer products of two vectors in \cite{Aharon&Elad&Bruckstein:06TSP, Ravishankar&Nadakuditi&Fessler:17TCI}.

\subsection{$k\rth$ Dictionary (Filter) Update}

\begin{figure}[!t]
\centering
\begin{tabular}{c}
\includegraphics[scale=0.5, trim=0 0.2em 2.2em 1em, clip]{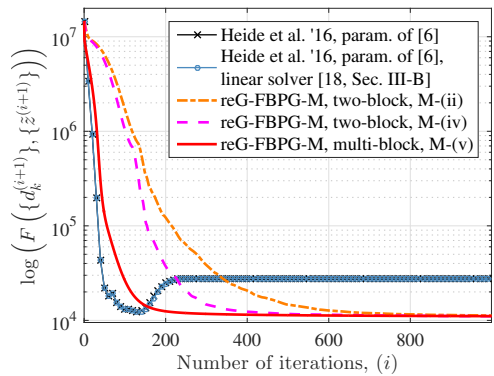} \\
{\small (a) The fruit dataset} \\
\includegraphics[scale=0.5, trim=0 0.2em 2.2em 1em, clip]{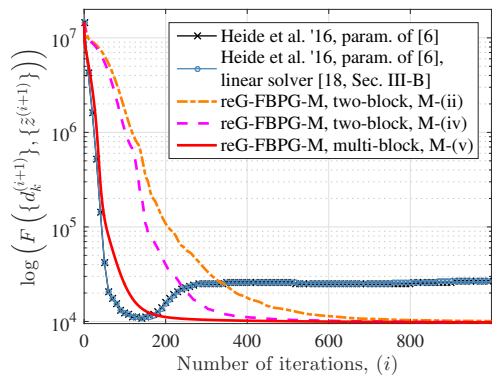}  \\
{\small (b) The city dataset} \\
\end{tabular}

\caption{Comparison of cost minimization between different CDL algorithms (the small datasets; for ADMM \cite{Heide&eta:15CVPR}, the number of whole iterations is the product of the number of inner iterations and that of outer iterations; reG-FBPG-M used the momentum-coefficient formula \R{eq:step_size1}). The multi-block framework significantly improves the convergence rate over the two-block schemes, with a possible reason that the majorizer in the multi-block update, i.e., M-(\romnum{5}), is tighter than those in the two-block update, i.e., M-(\romnum{1})--M-(\romnum{4}).}
\label{fig:Comp_newBPGM}
\end{figure}

We decompose the $\{ d_k : k=1,\ldots,K \}$-update problem \R{eq:sys:BPGM:synthF_kernel} into $K$ $d_k$-update problems as follows:
\ea{
\label{eq:sys:BPGM:synthF_kernel:decoup}
\min_{d_k} &~ \frac{1}{2} \nm{ \left( \left[ \begin{array}{c} y_1 \\ \vdots \\ y_L \end{array} \right] - \sum_{k'\neq k} \Psi_{k'} d_{k'} \right) - \Psi_k d_k }_2^2 
\\ 
\mbox{s.t.} ~&~ \| d_k \|_2^2 \leq 1, \nn
}
where the $k\rth$ submatrix of $\Psi \!=\! [\Psi_1 \!\cdots\! \Psi_K]$ \R{eq:sysMat:BPGM:synthF_kernel} is defined by
\be{
\label{eq:sysMat:BPGM:synthF_kernel:decoup}
\Psi_k := \left[ \arraycolsep=3pt \begin{array}{c} P_B \Phi^{-1} \diag(\hat{z}_{1,k}) \Phi P_S^T \\ \vdots \\ P_B \Phi^{-1} \diag(\hat{z}_{L,k}) \Phi P_S^T \end{array} \right],
}
$\{ \hat{z}_{l,k} = \Phi \tilde{z}_{l,k} : l=1,\ldots,L  \}$, and we use the most recent estimates of all other filters and coefficients in \R{eq:sys:BPGM:synthF_kernel:decoup} and \R{eq:sysMat:BPGM:synthF_kernel:decoup}, for $k=1,\ldots,K$.

\subsubsection{Separable Majorizer Design}

The following lemma introduces a majorization matrix for $\Psi_k^H \Psi_k$:

\lem{[Diagonal majorization matrix $M_{\Psi_k}$]\label{l:Kernel:diag:mltBlk}
The following diagonal matrix $M_{\Psi_k} \in \bbR^{D \times D}$ satisfies $M_{\Psi_k} \succeq \Psi_k^H \Psi_k$:
\bes{
M_{\Psi_k} = \diag \! \left( P_S \left| \Phi^{-1} \sum_{l=1}^L \diag( | \hat{z}_{l,k} |^2 ) \Phi \right| P_S^T 1_{D} \right).
}
}
\prf{\renewcommand{\qedsymbol}{}
See Section~\ref{sec:prf:l:Kernel:diag:mltBlk} of the supplementary material.
}

The design in Lemma~\ref{l:Kernel:diag:mltBlk} is expected to be tighter than those in Lemma~\ref{l:Kernel:identity} and \ref{l:Kernel:diag2}, because we use fewer bounds in designing it. Fig.~\ref{fig:Comp_newBPGM} supports this expectation through convergence rate; additionally, Fig.~\ref{fig:Comp_newBPGM} illustrates that the majorization matrix in Lemma~\ref{l:Kernel:diag:mltBlk} is expected to be tighter than that in Lemma~\ref{l:Kernel:diag1}.
Another benefit of the majorization matrix in Lemma~\ref{l:Kernel:diag:mltBlk} is lower computational complexity than those in the two-block approaches (in single-thread computing). 
As shown in Table~\ref{tab:comput_Mtype}-A, it allows up to $2K$ times faster the majorizer computations in the multi-block scheme (particularly, that in Lemma~\ref{l:Kernel:diag1}).

\subsubsection{Proximal Mapping}

Using \R{eq:quad_surrg}, the corresponding proximal mapping problem of \R{eq:sys:BPGM:synthF_kernel:decoup} is given by
\be{
\label{eq:prox:BPGM:kernel:multiBlk}
d_k^{(i+1)} = \argmin_{d_k} \, \frac{1}{2} \nm{ d_k - \nu_k^{(i)}  }_{M_{\Psi_k}}, \quad \mbox{s.t.} ~ \| d_k \|_2^2 \leq 1,
}
where
\eas{
\nu_k^{(i)} &= \acute{d}_k^{(i)} - \left( M_{\Psi_k}^{(i)} \right)^{-1} \left( \Psi_k^{(i)} \right)^H \left( \Psi_k^{(i)} \acute{d}_k^{(i)} - \check{y} \right),
\\
\check{y} &= \left[ \begin{array}{c} y_1 \\ \vdots \\ y_L \end{array} \right] - \sum_{k'\neq k} \Psi_{k'} d_{k'},
}
we construct $\Psi_k^{(i)}$ using \R{eq:sysMat:BPGM:synthF_kernel:decoup} with the updated $k\rth$ sparse code set $\{ \tilde{z}_{l,k}^{(i)} : l = 1,\ldots,L \}$, and $M_{\Psi_k}^{(i)}$ is a designed diagonal majorization matrix for $( \Psi_k^{(i)} )^H \Psi_k^{(i)}$.
Similar to Section~\ref{sec:prox:filter}, we apply the accelerated Newton's method in Section~\ref{sec:AccNewton} to efficiently solve \R{eq:prox:BPGM:kernel:multiBlk}.

\subsection{$k\rth$ Sparse Code Set Update}

We decompose the $\{ \tilde{z}_{l,k} : k=1,\ldots,K \}$-update problem \R{eq:sys:BPGM:synthF_spCode} into $K$ $\tilde{z}_{l,k}$-update problems as follows:
\ea{
\label{eq:sys:BPGM:synthF_spCode:decoup}
\min_{\tilde{z}_{l,k}} ~ \frac{1}{2} \left\| \left( y_l - \sum_{k' \neq k} \Gamma_{k'} \tilde{z}_{l,k'} \right) - \Gamma_k \tilde{z}_{l,k} \right\|_2^2 + \alpha \| \tilde{z}_{l,k} \|_1,
}
where the $k\rth$ submatrix of $\Gamma \!=\! [\Gamma_1 \!\cdots\! \Gamma_K]$ \R{eq:sysMat:BPGM:synthF_spCode} is defined by
\be{
\label{eq:sysMat:BPGM:synthF_spCode:decoup}
\Gamma_k := P_B \Phi^{-1} \diag(\lambda_k) \Phi,
}
$\{ \lambda_{k} = \Phi P_S^T d_{k} \}$, we use the most recent estimates of all other filters and coefficients in \R{eq:sys:BPGM:synthF_spCode:decoup} and \R{eq:sysMat:BPGM:synthF_kernel:decoup}, for $k=1,\ldots,K$.
Using \R{eq:sys:BPGM:synthF_spCode:decoup}, we update the $k\rth$ set of sparse codes $\{ \tilde{z}_{l,k} :  l=1,\ldots,L \}$, which is easily parallelizable over $l=1,\ldots,L$. Note, however, that this parallel computing scheme does not provide computational benefits over that in the two-block approach. Specifically, in each thread, the two-block scheme requires $\cO( K \tilde{N} ( \log \tilde{N} + 1 ) )$; and the multi-block requires $K$ times the cost $\cO( \tilde{N} ( \log \tilde{N} + 1 ) )$, i.e., $\cO( K \tilde{N} ( \log \tilde{N} + 1 ) )$.

\subsubsection{Separable Majorizer Design}

Applying Lemma~\ref{l:diag(|At||A|1)}, our diagonal majorization matrix for $\Gamma_k^H \Gamma_k$ is given in the following lemma:

\lem{[Diagonal majorization matrix $M_{\Gamma_k}$] \label{l:spCd:diag:mltBlk}
The following diagonal matrix $M_{\Gamma_k} \in \bbR^{\tilde{N} \times \tilde{N}}$ satisfies $M_{\Gamma_k} \succeq \Gamma_k^H \Gamma_k$:
\eas{
M_{\Gamma_k} = \diag \! \left( | \Gamma_k^H | | \Gamma_k | 1_{\tilde{N}} \right).
}
}

The design in Lemma~\ref{l:spCd:diag:mltBlk} is expected to be tighter than those in Lemma~\ref{l:spCd:diag1} and \ref{l:spCd:diag2}, because only a single bound is used in designing it.
Fig.~\ref{fig:Comp_newBPGM} supports our expectation through convergence rate per iteration.
In addition, the design in Lemma~\ref{l:spCd:diag:mltBlk} requires lower computation costs than those in the two-block schemes (in a single processor computing).
Specifically, it reduces complexities of computing those in multi-block scheme (particularly, that in Lemma~\ref{l:spCd:diag1}) by up to a factor of $K(1/D+1)$; see Table~\ref{tab:comput_Mtype}-B.

\subsubsection{Proximal Mapping} \label{sec:BPGM-multiBlk:spCd:Prox}

Using \R{eq:quad_surrg}, the corresponding proximal mapping problem of \R{eq:sys:BPGM:synthF_spCode:decoup} is given by:
\be{
\label{eq:prox:BPGM:spCd:multiBlk}
\tilde{z}_{l,k}^{(i+1)} = \argmin_{\tilde{z}_{l,k}} \, \frac{1}{2} \nm{ \tilde{z}_{l,k} - \zeta_{l,k}^{(i)} }_{\Gamma_k^{(i)}}^2 + \alpha \nm{ \tilde{z}_{l,k} }_1
}
where
\eas{
\zeta_{l,k}^{(i)} &= \acute{z}_{l,k}^{(i)} - \left( M_{\Gamma_k}^{(i)} \right)^{-1} \left( \Gamma_k^{(i)} \right)^H \left( \Gamma_k^{(i)} \acute{z}_{l,k}^{(i)} - \check{y}_l \right),
\\
\check{y}_l &= y_l - \sum_{k' \neq k} \Gamma_{k'} \tilde{z}_{l,k'},
}
we construct $\Gamma_k^{(i)}$ using \R{eq:sysMat:BPGM:synthF_spCode:decoup} with the updated $k\rth$ filter $d_k^{(i+1)}$, and $M_{\Gamma_k}^{(i)}$ is a designed diagonal majorization matrix for $( \Gamma_k^{(i)} )^H \Gamma_k^{(i)}$.
Similar to Section~\ref{sec:prox:spCd}, problem~\R{eq:prox:BPGM:spCd:multiBlk} is solved by the soft-shrinkage operator.

To efficiently compute $\sum_{k' \neq k} \Psi_{k'} d_{k'}$ in \R{eq:sys:BPGM:synthF_kernel:decoup} and $\sum_{k' \neq k} \Gamma_{k'} \tilde{z}_{l,k'}$ in \R{eq:sys:BPGM:synthF_spCode:decoup} at the $k\rth$ iteration, we update and store $\{ \Gamma_k \tilde{z}_{l,k} : l = 1,\ldots,L \}$---which is identical to $\Psi_k d_k$---with newly estimated $d_k^{(i+1)}$ and $\{ \tilde{z}_{l,k}^{(i+1)} :l = 1,\ldots,L \}$, and simply take sum in $\sum_{k' \neq k} \Psi_{k'} d_{k'}$ and $\sum_{k' \neq k} \Gamma_{k'} \tilde{z}_{l,k'}$.
Similar to $\Gamma^{(i)}$ in \R{eq:sysMat:BPGM:synthF_spCode} and $\Psi^{(i)}$ in \R{eq:sysMat:BPGM:synthF_kernel}, one can perform  $\Gamma_k^{(i)}$ in \R{eq:sysMat:BPGM:synthF_spCode:decoup} (or $(\Gamma_k^{(i)})^H$) and $\Psi_k^{(i)}$ in \R{eq:sysMat:BPGM:synthF_kernel:decoup} in a spatial domain---see Section~\ref{sec:prox:filter}.

\section{CDL-ACE: Application of CDL to Image Denoising} \label{sec:CDL-ACE}

Applying learned filters by CDL to some inverse problems is not straightforward due to \textit{model mismatch} between training and testing stages. 
CDL conventionally learns features from preprocessed training datasets (by, for example, the techniques in Section~\ref{sec:exp}); however, such nonlinear preprocessing techniques are not readily incorporated when solving inverse problems \cite{Serrano&etal:16CGF}.

The most straightforward approach in resolving the model mismatch is to learn filters from non-preprocessed training data, as noted in \cite[\S5.2]{Heide&eta:15CVPR}.
An alternative approach is to model (linear) contrast enhancement methods in CDL---similar to CDL-ACE in \cite{Chun&Fessler:17SAMPTA}---and apply them to solving inverse problems.
The CDL-ACE model is given by \cite{Chun&Fessler:17SAMPTA}
\begingroup
\setlength{\thinmuskip}{1.5mu}
\setlength{\medmuskip}{2mu plus 1mu minus 2mu}
\setlength{\thickmuskip}{2.5mu plus 2.5mu}
\ea{
\begin{split}
\label{eq:sys:CDLadt}
\min_{\{ d_k \}, \{ z_{l,k} \}, \{ \rho_l \}} &~ \sum_{l=1}^L \frac{1}{2} \left\| y_l - \left( P_B \sum_{k=1}^K d_{k} \circledast z_{l,k} \right) - \rho_l \right\|_2^2 
\\
& + \alpha \sum_{k=1}^K \left\| z_{l,k} \right\|_1 + \frac{\gamma}{2} \left\| C \rho_l \right\|_2^2
\\
\mathrm{s.t.} \qquad&~ \left\| d_{k} \right\|_2^2 \leq 1, \qquad k = 1,\ldots, K,
\end{split}
}
\endgroup
where $\{ \rho_l \in \bbR^N : l = 1,\ldots,L \}$ is a set of low-frequency component vectors and we design $C \in \bbR^{N' \times N}$ for adaptive contrast enhancement of $\{ y_l \}$ (see below).
Considering particular boundary conditions (e.g. periodic or reflective) for $\{ \rho_l \}$, we rewrite \R{eq:sys:CDLadt}  as follows \cite{Chun&Fessler:17SAMPTA}:
\begingroup
\setlength{\thinmuskip}{1.5mu}
\setlength{\medmuskip}{2mu plus 1mu minus 2mu}
\setlength{\thickmuskip}{2.5mu plus 2.5mu}
\ea{
\! \min_{\{ d_k \}, \{ z_{l,k} \}} & \sum_{l=1}^L \frac{1}{2} \left\| \tilde{y}_l - R \left( P_B \sum_{k=1}^K d_{k} \circledast z_{l,k} \right) \right\|_2^2 
 \! + \alpha \sum_{k=1}^K \left\| z_{l,k} \right\|_1
\nn \\
\! \mathrm{s.t.} ~~~~& \left\| d_{k} \right\|_2^2 \leq 1, \qquad k = 1,\ldots, K, \label{eq:sys:CDLadtRe}
}
\endgroup
where $\{ \tilde{y}_l :=  R y_l : l=1,\ldots,L \}$ and
\begingroup
\setlength{\thinmuskip}{1.5mu}
\setlength{\medmuskip}{2mu plus 1mu minus 2mu}
\setlength{\thickmuskip}{2.5mu plus 2.5mu}
\be{
\label{eq:def:R}
R :=  \left( \gamma C^T C \right)^{1/2} \left( \gamma C^T C + I \right)^{-1/2}.
}
\endgroup
The matrix $R$ in \R{eq:def:R} can be viewed as a simple form of a contrast enhancing transform (without divisive normalization by local variances), e.g., $R^T R y = y - (\gamma C^T C + I)^{-1} y$, where $(\gamma C^T C + I)^{-1}$ is a low-pass filter. 
To solve \R{eq:sys:CDLadtRe}, AL methods would now require six additional AL parameters to tune and consume more memory (than the ADMM approach in \cite{Heide&eta:15CVPR} solving \R{eq:sys:synth}); however, BPG-M methods are free from the additional parameter tuning processes and memory issues.

To denoise a measured image $b \in \bbR^n$ corrupted by AWGN ($\sim \mathcal{N}(0,\sigma^2)$), we solve the following optimization problem with the filters $\{ d_k^\star : k =1,\ldots,K \}$ learned via the CDL models, i.e., \R{eq:sys:synth} or, optimally, \R{eq:sys:CDLadtRe}, \cite{Chun&Fessler:17SAMPTA}:
\ea{
\label{eq:sys:denoising}
\left\{ \{ a_k^\star \}, \rho^\star \right\} = \argmin_{\{ a_{k} \}, \rho} &~ \frac{1}{2} \nm{ b - \left( P_B \sum_{k=1}^K d_{k}^\star \circledast a_k \right) - \rho }_2^2
\nn \\
&~ + \alpha' \sum_{k=1}^K \nm{a_k}_1 + \gamma' \nm{ C \rho }_2^2,
}
and synthesize the denoised image by $P_B \sum_{k=1}^K d_{k}^\star \circledast a_k^\star + \rho^\star$,
where $\{ a_k \in \bbR^n \}$ is a set of sparse codes, $\rho \in \bbR^n$ is a low-frequency component vectors, and $C \in \bbR^{n' \times n}$ is a regularization transform modeled in the CDL model \R{eq:sys:CDLadt}. 
Using the reformulation techniques in \R{eq:sys:CDLadt}--\R{eq:sys:CDLadtRe}, we rewrite \R{eq:sys:denoising} as a convex problem and solve it through FPG method using a diagonal majorizer (designed by a technique similar to Lemma~\ref{l:spCd:diag2}) and adaptive restarting \cite{Chun&Fessler:17SAMPTA}.

\section{Results and Discussion} \label{sec:result}

\subsection{Experimental Setup} \label{sec:exp}

Table~\ref{tab:name_BPG-M_Mtype} gives the naming conventions for the proposed BPG-M algorithms and designed majorizers.

We tested all the introduced CDL algorithms for two types of datasets: preprocessed and non-preprocessed.
The preprocessed datasets include the fruit and city datasets with $L = 10$ and $N = 100 \!\times\! 100$ \cite{Zeiler&etal:10CVPR, Heide&eta:15CVPR}, and the CT dataset with $L = 10$ and $N = 512 \!\times\! 512$ from down-sampled $512 \!\times\! 512$ XCAT phantom slices \cite{Segars&etal:08MP}---referred to the CT-(\romnum{1}) dataset.
The preprocessing includes local contrast normalization \cite[Adaptive Deconvolutional Networks Toolbox]{Zeiler&etal:11CVPR}, \cite[\S 2]{Jarrett&etal:09ICCV}, \cite{Zeiler&etal:10CVPR} and intensity rescaling to $[0,1]$ \cite{Zeiler&etal:10CVPR, Zeiler&etal:11CVPR, Bristow&etal:13CVPR, Heide&eta:15CVPR}.
The non-preprocessed dataset \cite[\S5.2]{Heide&eta:15CVPR} consists of XCAT phantom images of $L = 80$ and $N = 128 \!\times\! 128$, created by dividing down-sampled $512 \!\times\! 512$ XCAT phantom slices \cite{Segars&etal:08MP} into $16$ sub-images \cite{Olshausen&Field:96Nature, Bristow&etal:13CVPR}; we refer this to the CT-(\romnum{2}) dataset. 
Both the preprocessed and non-preprocessed datasets contain zero-mean training images (i.e., by subtracting the mean from each training image \cite[Adaptive Deconvolutional Networks Toolbox]{Zeiler&etal:11CVPR}, \cite[\S 2]{Jarrett&etal:09ICCV}; note that subtracting the mean can be omitted for the preprocessed datasets), as conventionally used in many (convolutional) dictionary learning studies, e.g., \cite{Xu&Yin:16IPI, Jarrett&etal:09ICCV, Zeiler&etal:11CVPR, Zeiler&etal:10CVPR, Bristow&etal:13CVPR, Heide&eta:15CVPR}. 
For image denoising experiments, we additionally trained filters by CDL-ACE \R{eq:sys:CDLadtRe} through the BPG-M method, and the non-preprocessed city datasets (however, note that we do not apply the mean subtraction step because it is not modeled in \R{eq:sys:CDLadtRe}. For all the CDL experiments, we trained filters of $D = 11 \!\times\! 11$ and $K = 100$ \cite{Heide&eta:15CVPR, Wohlberg:16ICIP}.

The parameters for the algorithms were defined as follows.
For CDL \R{eq:sys:synth} using both the preprocessed and non-preprocessed datasets, we set the regularization parameters as $\alpha = 1$ \cite{Heide&eta:15CVPR}. For CDL-ACE \R{eq:sys:CDLadtRe} using the non-preprocessed dataset, we set $\alpha = 0.4$.
We used the same (normally distributed) random initial filters and coefficients for each training dataset to fairly compare different CDL algorithms.
We set the the tolerance value, $\mathrm{tol}$ in \R{eq:stopCrit_relErr}, as $10^{-4}$.
Specific details regarding the algorithms are described below.

Comparing convergence rates in Fig.~\ref{fig:Comp_newBPGM} and execution time in Table~\ref{tab:comput&time&momory_algo}-C, we normalized the initial filters such that $\{ \nm{d_k}_2^2 \leq 1 : k=1,\ldots,K \}$ (we empirically observed that the normalized initial filters improve convergence rates of the multi-block algorithms, but marginally improve convergence rates of the two-block algorithms---for both ADMMs and BPG-M).
The execution time in Table~\ref{tab:comput&time&momory_algo} was recorded by (double precision) MATLAB implementations based on Intel Core i5 with $3.30$ GHz CPU and $32$ GB RAM.

\subsubsection{ADMM \cite{Heide&eta:15CVPR}} \label{sec:exp:ALM}

We first selected ADMM parameters as suggested in the corresponding MATLAB code of \cite{Heide&eta:15CVPR}: the ADMM parameters were selected by considering the maximum value of $\{ y_m : m=1,\ldots,M \}$, similar to \cite{Almeida&Figueiredo:13TIP}. We used $10$ inner iterations (i.e., $\mathrm{Iter}_{\text{ADMM}}$ in Table~\ref{tab:comput&time&momory_algo}-A) for each kernel and sparse code update \cite{Heide&eta:15CVPR} and set the maximum number of outer iterations to $100$.
We terminated the iterations if either of the following stopping criteria are met before reaching the maximum number of iterations \cite{Heide&eta:15CVPR}:
\ea{
\label{eq:stopCrit_obj}
F( d^{(i+1)},  \tilde{z}^{(i)} )  & \geq F( d^{(i)}, \tilde{z}^{(i)} ) \quad \mathrm{and}
\nn \\
F( d^{(i+1)}, \tilde{z}^{(i+1)} ) & \geq F( d^{(i)}, \tilde{z}^{(i)} ),
}
or
\begingroup
\setlength{\thinmuskip}{1.5mu}
\setlength{\medmuskip}{2mu plus 1mu minus 2mu}
\setlength{\thickmuskip}{2.5mu plus 2.5mu}
\be{
\label{eq:stopCrit_relErr}
\frac{ \big\| d^{(i+1)} - d^{(i)} \big\|_2 }{ \big\| d^{(i+1)} \big\|_2 } < \mathrm{tol}  \quad \mathrm{and} \quad
\frac{ \big\| \tilde{z}^{(i+1)} - \tilde{z}^{(i)} \big\|_2 }{ \big\| \tilde{z}^{(i+1)} \big\|_2 } < \mathrm{tol},
}
\endgroup
where $d$ and $\tilde{z}$ are concatenated vectors from $\{ d_k \}$ and $\{ \tilde{z}_{l,k} \}$, respectively. These rules were applied at the outer iteration loop \cite{Heide&eta:15CVPR}.
For the experiments in Figs.~\ref{fig:Comp_newBPGM} and \ref{fig:filters_ADMM}, and Table~\ref{tab:comput&time&momory_algo}-C, we disregarded the objective-value-based termination criterion \R{eq:stopCrit_obj}.
For a memory-efficient variant of ADMM \cite{Heide&eta:15CVPR}, we replaced the solver in \cite[(11)]{Heide&eta:15CVPR} with that in \cite[\S\Romnum{3}-B]{Wohlberg:16TIP} to solve the linear system \cite[(10)]{Heide&eta:15CVPR}, and tested it with the same parameter sets above.

\subsubsection{BPG-M Algorithms}

We first selected the parameter $\delta$ in \R{eq:extraM_update_diag} as $1-\varepsilon$, where $\varepsilon$ is the (double) machine epsilon value, similar to \cite{Xu&Yin:16IPI}.
For the gradient-mapping restarting, we selected the parameter $\omega$ in \R{eq:restart:grad} as $\cos(95^{\circ})$, similar to \cite{Muckley&Noll&Fessler:15TMI}.
For the accelerated Newton's method, we set the initial point $\varphi_k^{(0)}$ to $0$, the tolerance level for $| \varphi_k^{(i'+1)} - \varphi_k^{(i')} |$ to $10^{-6}$, and the maximum number of iterations to $10$, for $k=1,\ldots,K$.
The maximum number of BPG-M iterations was set to $\mathrm{Iter} = 1000$.
We terminated the iterations if the relative error stopping criterion \R{eq:stopCrit_relErr} was met before reaching the maximum number of iterations.

\subsubsection{Image Denoising with Learned Filters via CDL} \label{sec:exp:img_denoise}

For image denoising applications, we corrupted a test image with relatively strong AWGN, i.e., $\textmd{SNR} \!=\! 10$~dB. 
We denoised the noisy image through the following methods (all the parameters were selected as suggested in \cite{Chun&Fessler:17SAMPTA}, giving the best peak signal-to-noise ratio (PSNR) values): \textit{1)} adaptive Wiener filtering with $3 \!\times\! 3$ window size; \textit{2)} total variation (TV) with MFISTA using its regularization parameter $0.8 \sigma$ and maximum number of iterations $200$ \cite{Beck&Teboulle:09TIP}; \textit{3)} image denoiser \R{eq:sys:denoising} with $100$ (empirically) convergent filters trained by CDL model \R{eq:sys:synth} (i.e., Fig.~\ref{fig:filters_ADMM}(b)) and preprocessed training data, $\alpha' \!=\! 2.5\sigma$, the first-order finite difference for $C$ in \R{eq:sys:denoising} \cite{Serrano&etal:16CGF}, and $\gamma' \!=\! 10 \sigma$; and \textit{4)} \R{eq:sys:denoising} with $100$ learned filters by CDL-ACE \R{eq:sys:CDLadt}, $\alpha' \!=\! \alpha \cdot 5.5 \sigma$, and $\gamma' \!=\! \gamma \cdot 5.5 \sigma$.
For \R{eq:sys:denoising}, the stopping criteria is set similar to \R{eq:stopCrit_relErr} (with $\mathrm{tol} \!=\! 10^{-3}$) before reaching the maximum number of iterations $100$.

\subsection{BPG-M versus ADMM \cite{Heide&eta:15CVPR} and Its Memory-Efficient Variant for CDL \R{eq:sys:synth}} \label{sec:result:BPGMvsADMM}

\begin{table}

\caption{Comparison of Computational Complexity, Execution Time, and Memory Requirement from Different Single-Threaded CDL Algorithms}
\vspace{-1em}
\label{tab:comput&time&momory_algo}
\renewcommand*
\arraystretch{1.1}
\centering

\vspace{-0.1em}
\subtable{
\centering
\begin{tabular}{l|l}
\hline \hline
\multicolumn{1}{c|}{Algorithms} & \specialcell[c]{\hspace{0.75em}A. Computations for updating whole \\ \hspace{0.75em}blocks in a single (outer) iteration$^a$} \\
\hline \hline
\multirow{6}{*}{\vspace{-2em}ADMM in Heide et al. \cite{Heide&eta:15CVPR}} 
& Case $K > L$ \cite{Heide&eta:15CVPR}: \\ 
& $\cO \! \left( K L^2 \tilde{N} + ( \mathrm{Iter}_{\text{ADMM}} - 1 ) \cdot K L \tilde{N} \right.$ \\ 
& $\left. ~\quad + \mathrm{Iter}_{\text{ADMM}} \cdot K L ( \tilde{N} \log \tilde{N} + \tilde{N} ) \right) \hspace{-0.5em}$ \\
& Case $K \leq L$ \cite{Heide&eta:15CVPR}: \\ 
& $\cO \! \left( K^3 \tilde{N} + ( \mathrm{Iter}_{\text{ADMM}} - 1 ) \cdot K^2 L \right.$ \\
& $\left. ~\quad + \mathrm{Iter}_{\text{ADMM}} \cdot K L ( \tilde{N} \log \tilde{N} + \tilde{N} )  ) \right) \hspace{-0.5em}$ \\ 
\hline
\multirow{2}{*}{\vspace{-0.75em}\specialcell[c]{ADMM in Heide et al. \cite{Heide&eta:15CVPR} \\ w. linear solver \cite[\S\Romnum{3}-B]{Wohlberg:16TIP}}}
& $\cO \! \left( \mathrm{Iter}_{\text{ADMM}} \cdot \big( K L^2 \tilde{N} \right.$ \\
& $\left. ~\quad + K L ( \tilde{N} \log \tilde{N} + \tilde{N} ) \big) \right) \hspace{-0.5em}$ \\
\hline
{reG-FBPG-M, multi-block} & $\cO \! \left( K \cdot L ( \tilde{N} \log \tilde{N} + D/L + \tilde{N} ) \right)$ \\
\hline \hline
\end{tabular}
}

\subtable{
\centering
\begin{tabular}{c|cccc}
\hline \hline
\multirow{2}{*}{Algorithms} & \multicolumn{4}{c}{\specialcell{B. Execution time$^b$ \\ $(\text{hours} : \text{minutes})$}} \\
& Fruit & City & CT-(\romnum{1}) & CT-(\romnum{2}) \\
\hline \hline
ADMM in Heide et al. \cite{Heide&eta:15CVPR} & $0:45$ & $0:45$ & $\cdot$ & $\cdot$ \\
\hline
ADMM in Heide et al. \cite{Heide&eta:15CVPR} & \multirow{2}{*}{$0:53$} & \multirow{2}{*}{$0:53$} & \multirow{2}{*}{$\cdot$} &  \multirow{2}{*}{$44:12$} \\
w. linear solver \cite[\S\Romnum{3}-B]{Wohlberg:16TIP} & & & & \\
\hline
reG-FBPG-M, multi-block & $0:54$ & $0:55$ & $36:09$ & $17:25$ \\
\hline \hline
\end{tabular}
}

\subtable{
\centering
\begin{tabular}{c|cc}
\hline \hline
\multirow{2}{*}{Algorithms} & \multicolumn{2}{c}{\specialcell{C. Total dimension of \\ variables to be stored$^c$}} \\ 
& Filter update & Sparse code update
\\
\hline \hline
\multirow{2}{*}{ADMM in Heide et al. \cite{Heide&eta:15CVPR}}
& $2 \tilde{N} ( 2 L + K )$  & $2 \tilde{N} L ( 2 K + 1 )$  \\
& \multicolumn{2}{l}{\hspace{5.2em}$+K^2 \tilde{N}$} \\
\hline
ADMM in Heide et al. \cite{Heide&eta:15CVPR} &  $2 \tilde{N} ( 2 L + K )$  & $2 \tilde{N} L ( 2 K + 1 )$ \\
w. linear solver \cite[\S\Romnum{3}-B]{Wohlberg:16TIP} & \multicolumn{2}{l}{\hspace{5.2em}$+K L \tilde{N}$} \\
\hline
\multirow{2}{*}{reG-FBPG-M, multi-block} 
& {$D(3K+2)$} & {$\tilde{N} (2KL + K + 2)$} \\
& \multicolumn{2}{l}{\hspace{5.2em}$+K L N$} \\
\hline \hline
\end{tabular}
}

\smallskip
\begin{myquote}{0.1in}
$^a$Here, $\mathrm{Iter}_{\text{ADMM}}$ denotes the number of (inner) ADMM iterations in each block update in the ADMM framework in \cite{Heide&eta:15CVPR}. For fair comparison with ADMM \cite{Heide&eta:15CVPR}, one should multiply the cost of a single iteration in reG-FBPG-M by $\mathrm{Iter}_{\text{ADMM}}$. 
\\
$^b$The symbol $\cdot$ means that the execution time cannot be recorded due to exceeding available memory.
\\
$^c$For the ADMM approach in \cite{Heide&eta:15CVPR}, one must store previously updated filters $\{ d_k^{(i)} \}$ and sparse codes $\{ \tilde{z}_{l,k}^{(i)} \}$, because it outputs them instead of the current estimates $\{ d_k^{(i+1)} \}$ and $\{ \tilde{z}_{l,k}^{(i+1)} \}$, when it is terminated by the objective function criterion \R{eq:stopCrit_obj}. The additional dimension on the second line of each method corresponds to the following: for ADMM \cite{Heide&eta:15CVPR}, it is the dimension of variables to solve the linear systems \cite[(10)]{Heide&eta:15CVPR} through the Parseval tricks in \cite{Heide&eta:15CVPR}; for ADMM \cite{Heide&eta:15CVPR} using linear solver \cite[\S\Romnum{3}-B]{Wohlberg:16TIP}, it is the largest dimension to apply \cite[\S\Romnum{3}-B]{Wohlberg:16TIP} to solve \cite[(10)]{Heide&eta:15CVPR}; for multi-block reG-FBPG-M,  it is the dimension of $\{ \Gamma_k \tilde{z}_{l,k} : l = 1,\ldots,L \}$ or $\Psi_k d_k$ for faster computation of $\sum_{k' \neq k} \Psi_{k'} d_{k'}$ in \R{eq:sys:BPGM:synthF_kernel:decoup} and $\sum_{k' \neq k} \Gamma_{k'} \tilde{z}_{l,k'}$ in \R{eq:sys:BPGM:synthF_spCode:decoup} (see details in Section~\ref{sec:BPGM-multiBlk:spCd:Prox}).
\end{myquote}

\end{table}

\begin{table}
\caption{Comparisons of Objective Values with Different Convolutional Dictionary Learning Algorithms}
\label{tab:Comp:obj:ALMvsBPG-M}
\renewcommand*
\arraystretch{1.1}
\centering

\begin{tabular}{c|ccc}
\hline \hline
\multirow{2}{*}{Algorithms$^\dagger$} & \multicolumn{2}{c}{Objective values} \\
& Fruit & City \\
\hline \hline
ADMM \cite{Heide&eta:15CVPR}, param. of \cite{Heide&eta:15CVPR}  & \multicolumn{1}{r}{$12594$} & \multicolumn{1}{r}{$11057$}   \\
\specialcell[c]{ADMM \cite{Heide&eta:15CVPR}, param. of \cite{Heide&eta:15CVPR}, \\ no termination by \R{eq:stopCrit_obj}}  & \multicolumn{1}{r}{$27415$} & \multicolumn{1}{r}{$26375$} \\
ADMM \cite{Heide&eta:15CVPR}, param. $1$  & \multicolumn{1}{r}{$53504$} & \multicolumn{1}{r}{$53907$} \\
\specialcell[c]{ADMM \cite{Heide&eta:15CVPR} w. linear solver \cite[\S\Romnum{3}-B]{Wohlberg:16TIP}, \\ param. of \cite{Heide&eta:15CVPR}, no termination by \R{eq:stopCrit_obj}}  & \multicolumn{1}{r}{$27405$} & \multicolumn{1}{r}{$26372$} \\
\hline
FBPG-M, two-block, \R{eq:step_size1} & \multicolumn{1}{r}{$11013$} & \multicolumn{1}{r}{$9585$}  \\
reO-BPG-M, two-block & \multicolumn{1}{r}{$11039$} & \multicolumn{1}{r}{$9606$} \\
reG-BPG-M, two-block & \multicolumn{1}{r}{$11081$} & \multicolumn{1}{r}{$9674$} \\
reG-FBPG-M, two-block, \R{eq:step_size1} & \multicolumn{1}{r}{$11068$} & \multicolumn{1}{r}{$9644$} \\
reG-FBPG-M, two-block, \R{eq:step_size2} & \multicolumn{1}{r}{$11149$} & \multicolumn{1}{r}{$9648$} \\
reG-FBPG-M, multi-block, \R{eq:step_size1} & \multicolumn{1}{r}{$10980$} & \multicolumn{1}{r}{$9698$} \\
\hline \hline
\end{tabular}

\medskip
\begin{myquote}{0.1in}
$^\dagger$The two-block BPG-M algorithms used the M-(\romnum{4}) majorizer.
\end{myquote}
\end{table}

The BPG-M methods guarantee convergence without difficult parameter tuning processes. Figs.~\ref{fig:Comp_newBPGM}, \ref{fig:filters_BPGM}, and \ref{fig:filters_ADMM} show that the BPG-M methods converge more stably than ADMM \cite{Heide&eta:15CVPR} and the memory-efficient variant of ADMM \cite{Heide&eta:15CVPR}. 
When the ADMM parameters are poorly chosen, for example simply using $1$, the ADMM algorithm fails (see Table~\ref{tab:Comp:obj:ALMvsBPG-M}).
The objective function termination criterion \R{eq:stopCrit_obj} can stabilize ADMM; however, note that terminating the algorithm with \R{eq:stopCrit_obj} is not a natural choice, because the monotonic decrease in objective function values is not guaranteed \cite{Xu&Yin:13SIAM}.
For the small datasets (i.e., the fruit and city datasets), the execution time of reG-FBPG-M using the multi-block scheme is comparable to that of the ADMM approach \cite{Heide&eta:15CVPR} and its memory-efficient variant; see Table~\ref{tab:comput&time&momory_algo}-B.
Based on the numerical experiments in \cite{Wohlberg:16ICIP}, for small datasets particularly with the small number of training images, the state-of-the-art ADMM approach in \cite[AVA-MD]{Wohlberg:16ICIP} using the single-set-of-iterations scheme \cite{Wohlberg:16TIP} (or \cite{Bristow&etal:13CVPR}) can be faster than multi-block reG-FBPG-M; however, it lacks theoretical convergence guarantees and can result in non-monotone minimization behavior---see Section~\ref{sec:CDLmodel} and \cite[Fig.~2,~AVA-MD]{Wohlberg:16ICIP}.

The proposed BPG-M-based CDL using the multi-block scheme is especially useful to large datasets having large image size or many images (compared to ADMM \cite{Heide&eta:15CVPR} and its memory-efficient variant applying linear solver \cite[\S\Romnum{3}-B]{Wohlberg:16TIP}):
\bulls{
\item The computational complexity of BPG-M depends mainly on the factor $K \cdot L \cdot \tilde{N} \log \tilde{N}$; whereas that of ADMM \cite{Heide&eta:15CVPR} depends not only on the factor $K \cdot L \cdot \tilde{N} \log \tilde{N}$, but also on the approximated factors $K \cdot L^2 \cdot \tilde{N} \cdot \mathrm{Iter}_{\text{ADMM}}^{-1}$ (for $K > L$) or $K^3 \cdot \tilde{N} \cdot \mathrm{Iter}_{\text{ADMM}}^{-1}$ (for $K \leq L$). The memory-efficient variant of ADMM \cite{Heide&eta:15CVPR} requires even higher computational complexity than ADMM \cite{Heide&eta:15CVPR}: it depends both on the factors $K \cdot L \cdot \tilde{N} \log \tilde{N}$ and $K \cdot L^2 \cdot \tilde{N}$.
See Table~\ref{tab:comput&time&momory_algo}-A--B.
\item The multi-block reG-FBPG-M method requires much less memory than ADMM \cite{Heide&eta:15CVPR} and its variant.
In the filter updates, it only depends on the parameter dimensions of filters (i.e., $K, D$); however, ADMM requires the amount of memory depending on the dimensions of training images and the number of filters (i.e., $\tilde{N}, L, K$).
In the sparse code updates, the multi-block reG-FBPG-M method requires about half the memory of ADMM.  Additionally, there exists no $K^2$ factor dependence in multi-block reG-FBPG-M. 
The memory-efficient variant of ADMM \cite{Heide&eta:15CVPR} removes the $K^2$ factor dependence, but still requires higher memory than multi-block reG-FBPG-M.
See Table~\ref{tab:comput&time&momory_algo}-C.
}
Table~\ref{tab:comput&time&momory_algo}-B shows that the ADMM approach in \cite{Heide&eta:15CVPR} and/or its memory-efficient variant fail to run CDL for the larger datasets (i.e., CT-(\romnum{1}) and CT-(\romnum{2})), due to its high memory usage. 
By not caching the inverted matrices \cite[(11)]{Heide&eta:15CVPR} computed at the beginning of each block update, the memory-efficient variant of ADMM \cite{Heide&eta:15CVPR} avoids the $K^2$ factor dependence in memory requirement. However, its computational cost now depends on the factor $L^2$ multiplied with $K$ and $\tilde{N}$; this product becomes a serious computational bottleneck as $L$---the number of training images---grows. See the CT-(\romnum{2}) column in Table~\ref{tab:comput&time&momory_algo}-B. 
(Note that single-set-of-iterations ADMM \cite[AVA-MD]{Wohlberg:16ICIP} obeys the same trends.)
Heide et al.'s report that their ADMM can handle the large dataset (of $L=10$, $N = 1000 \!\times\! 1000$) that the patch-based method (i.e., K-SVD \cite{Aharon&Elad&Bruckstein:06TSP} using all patches) cannot, due to its high memory usage \cite[\S3.2, using $132$ GB RAM machine]{Heide&eta:15CVPR}.
Combining these results, the BPG-M-based CDL algorithm (particularly using the multi-block scheme) is a reasonable choice to learn dictionary from large datasets. Especially, the multi-block BPG-M method is well-suited to CDL with large datasets consisting of a large number of (relatively small-dimensional) signals---for example, the datasets are often generated by dividing (large) images into many sub-images \cite{Heide&eta:15CVPR, Bristow&etal:13CVPR, Wohlberg:16TIP}.

For the non-preproccsed dataset (i.e., CT-(\romnum{2})), the proposed BPG-M algorithm (specifically, reG-FBPG-M using the multi-block scheme) properly converges to desirable solutions, i.e., the resultant filters and sparse codes (of sparsity $5.25$\%) properly synthesize training images. However, the memory-efficient variant of ADMM \cite{Heide&eta:15CVPR} does not converge to the desirable solutions. Compare the results in Fig.~\ref{fig:filters_BPGM}(d) to those in Fig.~\ref{fig:filters_ADMM}(c).

\begin{figure}[!t]
\centering
\begin{tabular}{c}
\includegraphics[scale=0.5, trim=0 0.2em 2.2em 1em, clip]{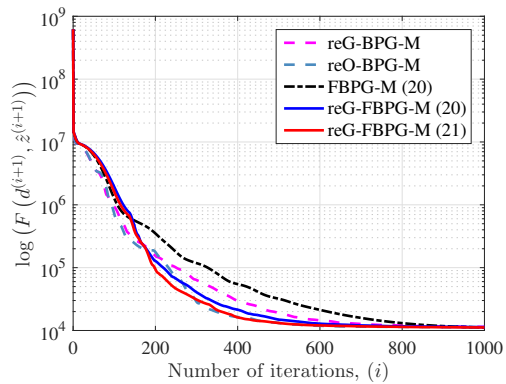} \\
\specialcell[c]{\small (a) Cost minimization with different accelerated \vspace{-0.25em} \\ \small BPG-M algorithms using M-(\romnum{2})} \\
\includegraphics[scale=0.5, trim=0 0.2em 2.2em 1em, clip]{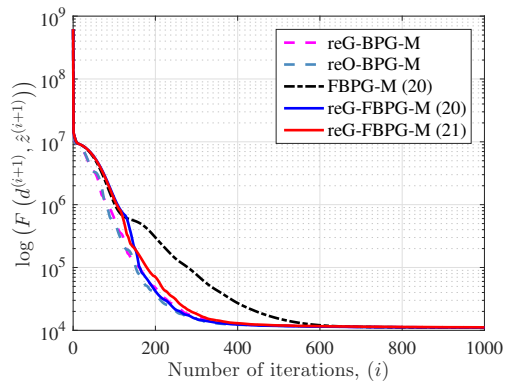} \\
\specialcell[c]{\small (b) Cost minimization with different accelerated \vspace{-0.25em} \\ \small BPG-M algorithms using M-(\romnum{3})}
\end{tabular}

\caption{Comparison of cost minimization between different accelerated BPG-M CDL algorithms (the fruit dataset).}
\label{fig:Comp_Convg_BPG}
\end{figure}

\begin{figure*}[!t]
\small\addtolength{\tabcolsep}{-2pt}
\renewcommand{\arraystretch}{1.5}
\centering

\begin{tabular}{cccc}
{\small Learned filters} & \multicolumn{2}{c}{\small Examples of synthesized images} & {\small Convergence behavior} \\
\multirow{2}{*}[5.9em]{
\includegraphics[scale=0.52, trim=4em 6em 1.4em 5.2em, clip]{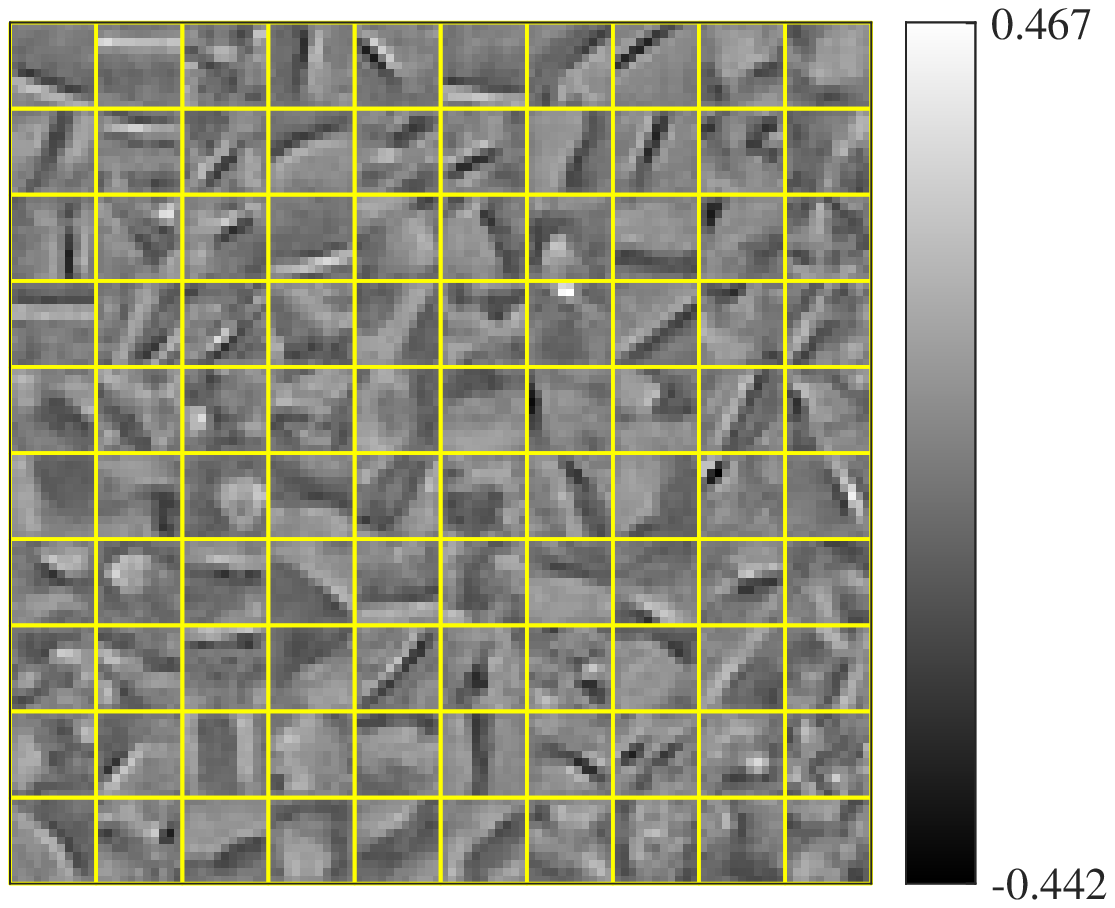} 
}
& \includegraphics[width=2.1cm,height=2.1cm,keepaspectratio]{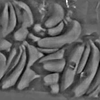}
& \includegraphics[width=2.1cm,height=2.1cm,keepaspectratio]{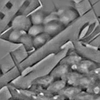}
& \multirow{2}{*}[5.75em]{
\includegraphics[scale=0.54, trim=0.2em 0.2em 1.8em 1.5em, clip]{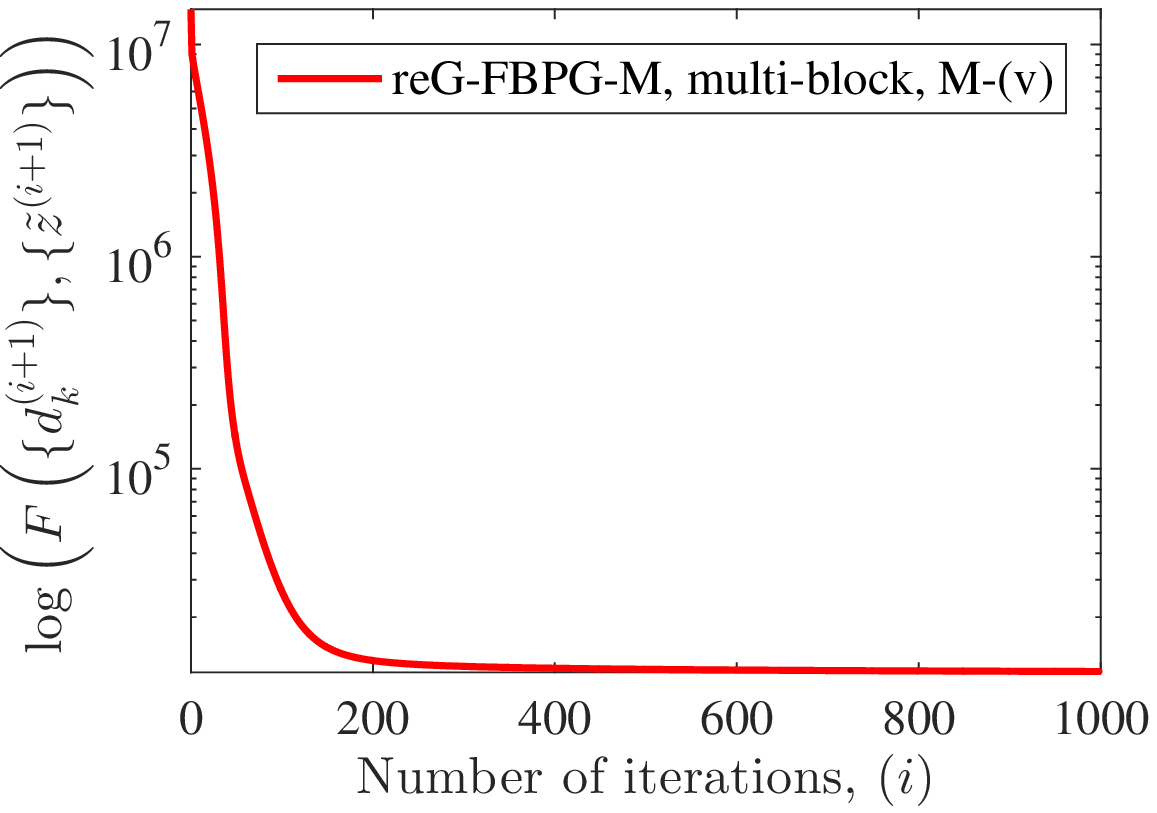} 
} 
\\
& \includegraphics[width=2.1cm,height=2.1cm,keepaspectratio]{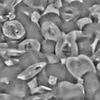} 
& \includegraphics[width=2.1cm,height=2.1cm,keepaspectratio]{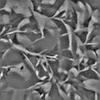}
\\
& {\small $\cdots$} & {\small $\cdots$} & \vspace{-0.5em} \\
\multicolumn{4}{c}{\small (a) The fruit dataset ($L = 10$, $N = 100 \!\times\! 100$)} 
\vspace{0.5em} \\
\multirow{2}{*}[5.9em]{
\includegraphics[scale=0.52, trim=4em 6em 1.4em 5.2em, clip]{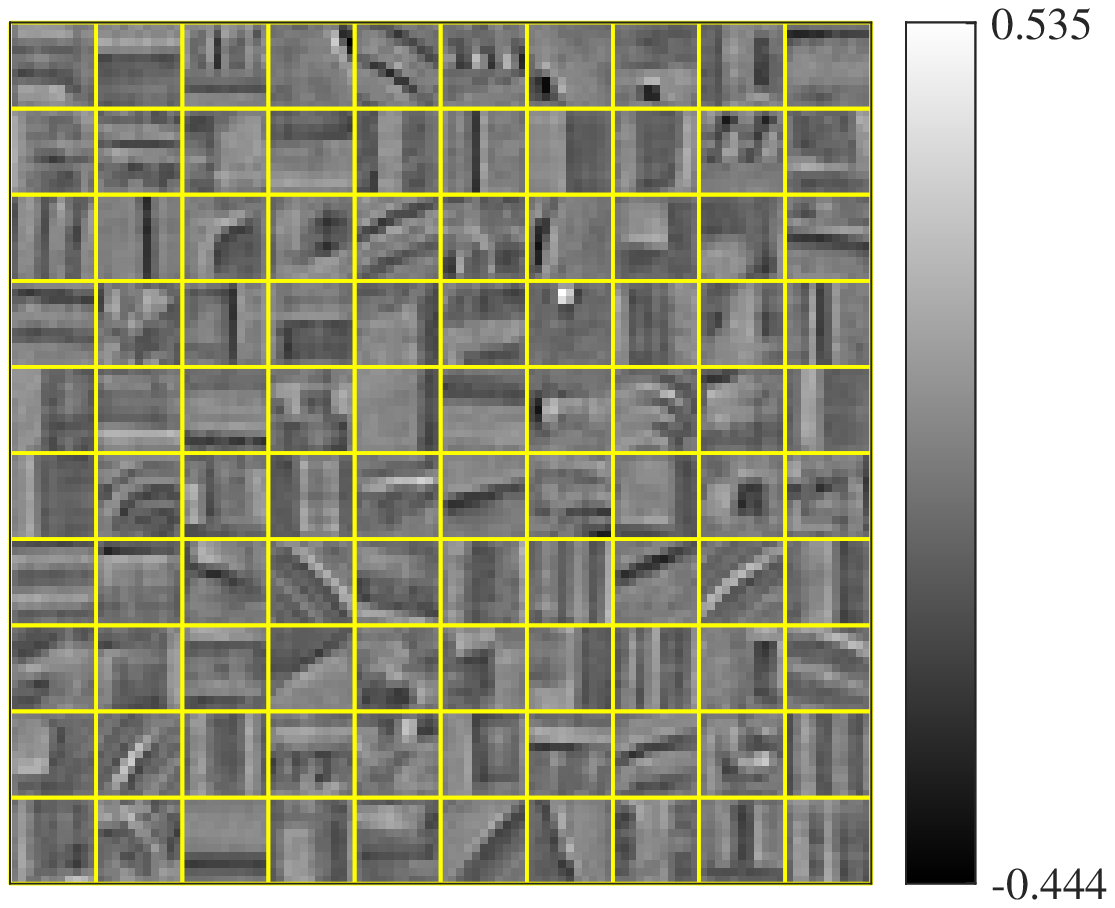} 
}
& \includegraphics[width=2.1cm,height=2.1cm,keepaspectratio]{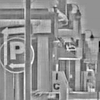}
& \includegraphics[width=2.1cm,height=2.1cm,keepaspectratio]{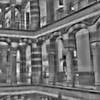}
& \multirow{2}{*}[5.75em]{
\includegraphics[scale=0.54, trim=0.2em 0.2em 1.8em 1.5em, clip]{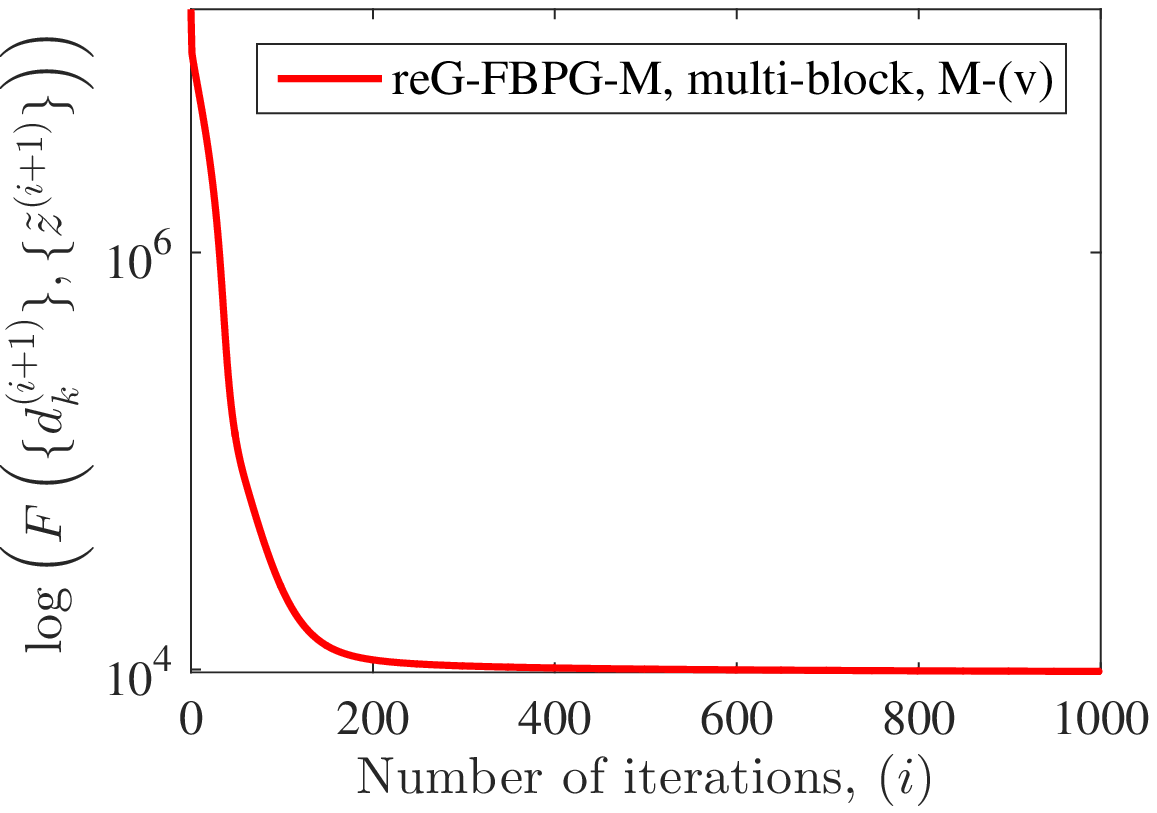} 
} 
\\
& \includegraphics[width=2.1cm,height=2.1cm,keepaspectratio]{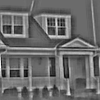} 
& \includegraphics[width=2.1cm,height=2.1cm,keepaspectratio]{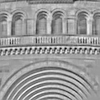}
\\
& {\small $\cdots$} & {\small $\cdots$} & \vspace{-0.5em} \\
\multicolumn{4}{c}{\small (b) The city dataset ($L = 10$, $N = 100 \!\times\! 100$)} 
\vspace{0.5em} \\
\multirow{2}{*}[5.9em]{
\includegraphics[scale=0.52, trim=4em 6em 1.4em 5.2em, clip]{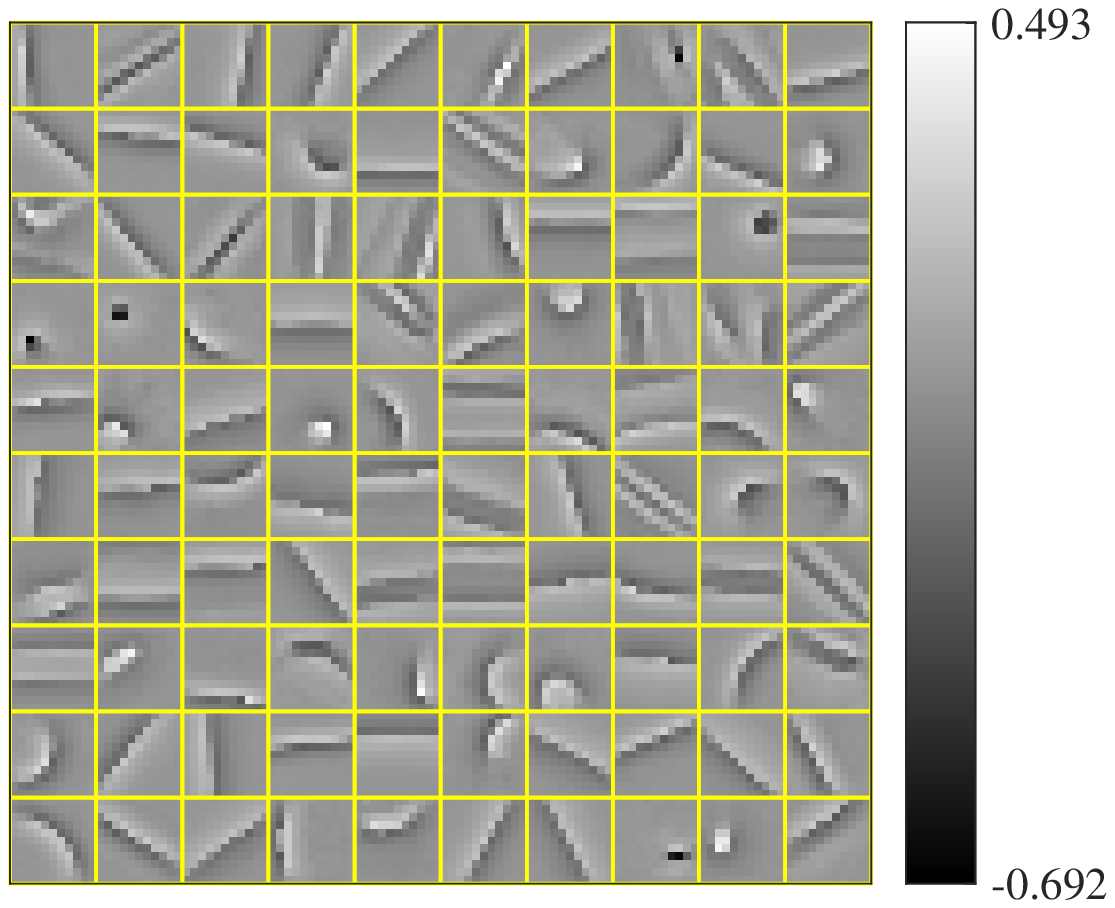} 
}
& \includegraphics[width=2.1cm,height=2.1cm,keepaspectratio]{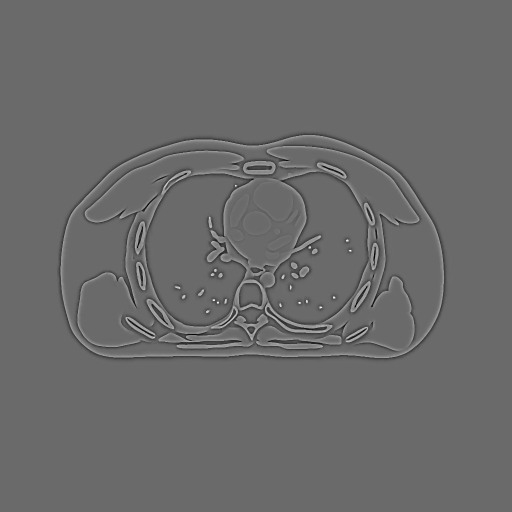}
& \includegraphics[width=2.1cm,height=2.1cm,keepaspectratio]{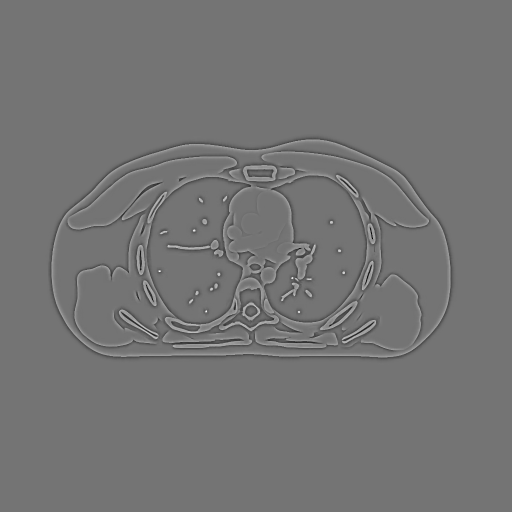}
& \multirow{2}{*}[5.75em]{
\includegraphics[scale=0.54, trim=0.2em 0.2em 1.8em 1.5em, clip]{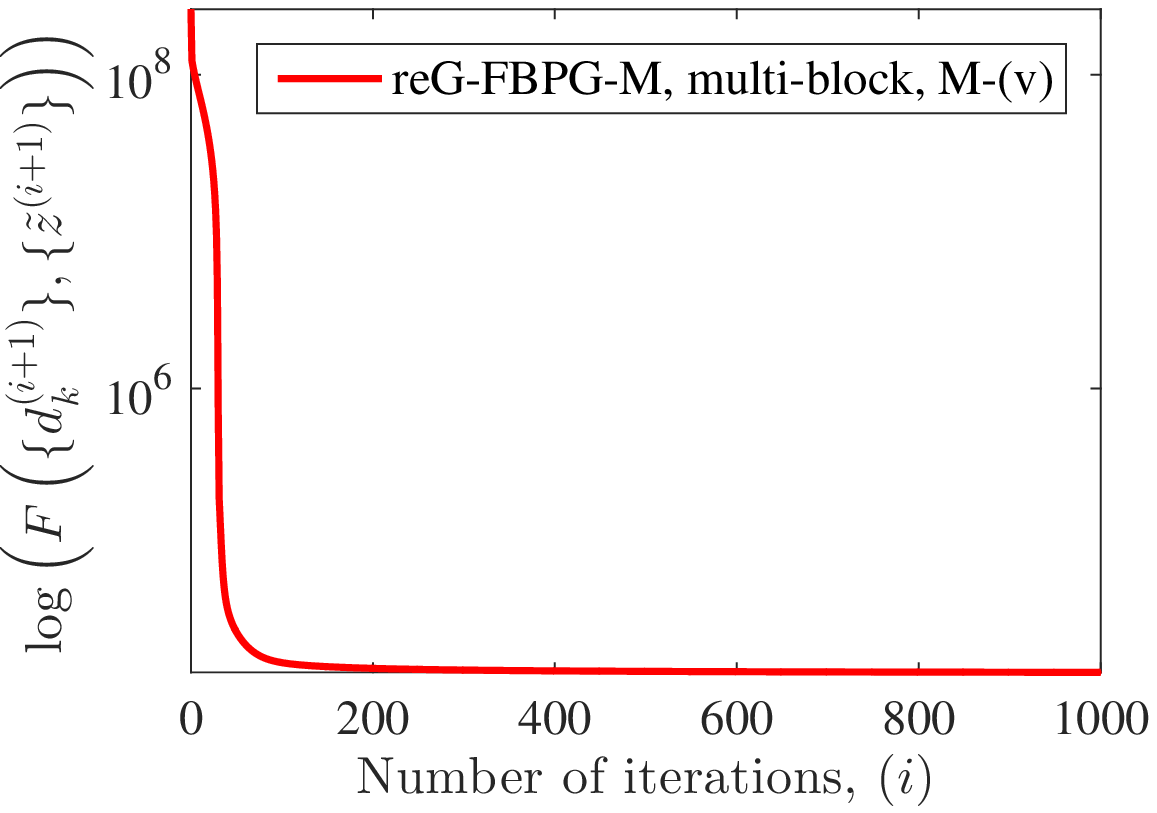} 
} 
\\
& \includegraphics[width=2.1cm,height=2.1cm,keepaspectratio]{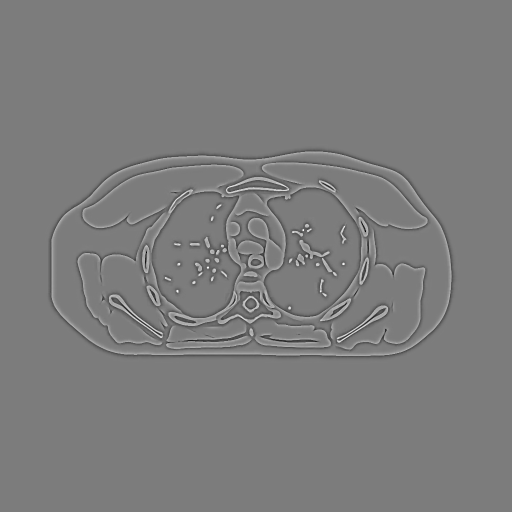} 
& \includegraphics[width=2.1cm,height=2.1cm,keepaspectratio]{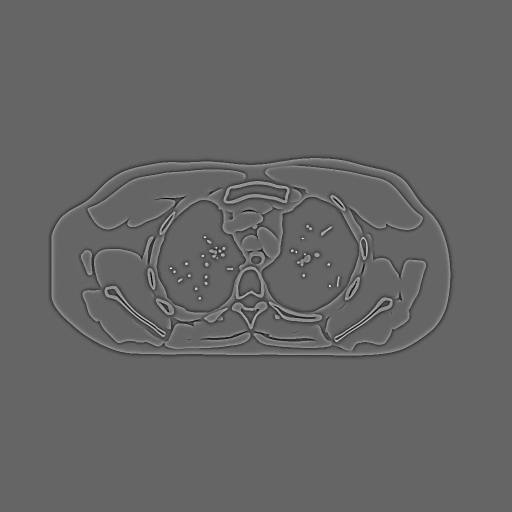}
\\
& {\small $\cdots$} & {\small $\cdots$} & \vspace{-0.5em} \\
\multicolumn{4}{c}{\small (c) The CT-(\romnum{1}) dataset ($L = 10$, $N = 512 \!\times\! 512$)} 
\vspace{0.5em} \\
\multirow{2}{*}[5.9em]{
\includegraphics[scale=0.52, trim=4em 6em 1.4em 5.2em, clip]{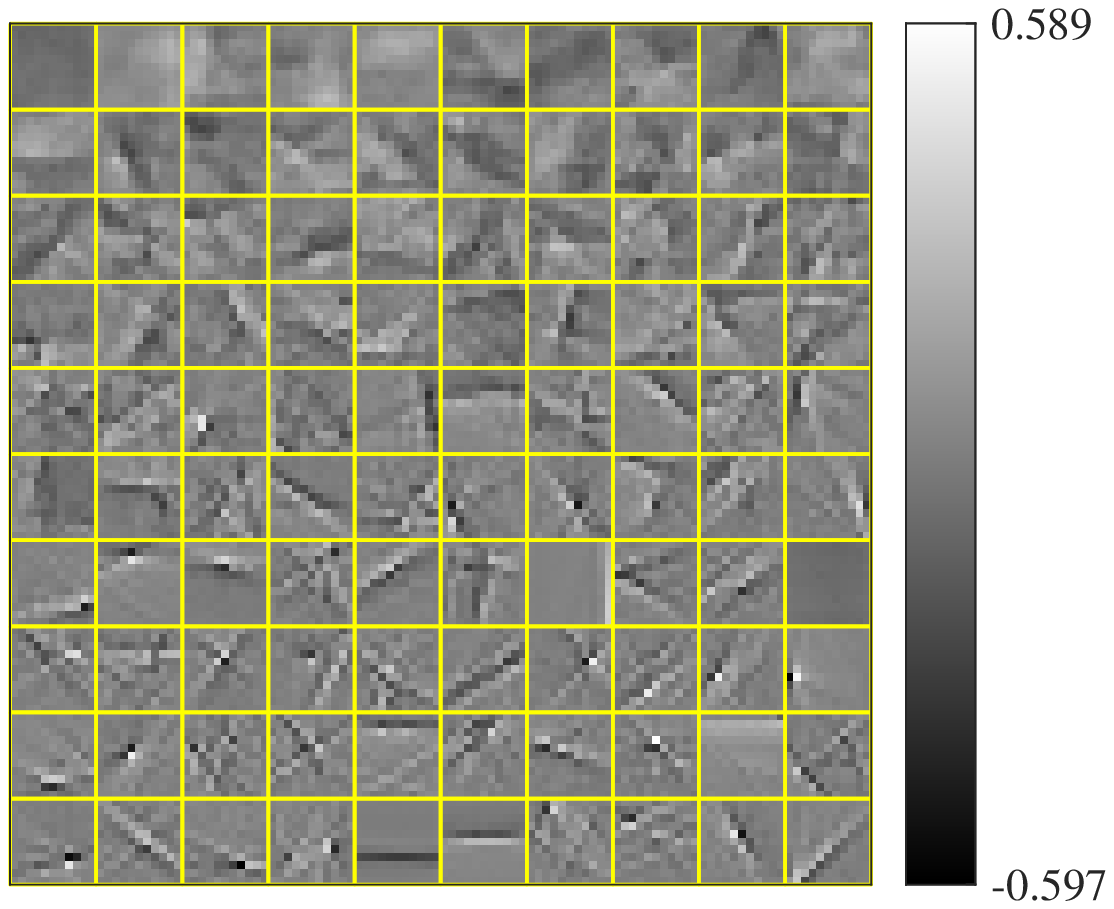} 
}
& \includegraphics[width=2.1cm,height=2.1cm,keepaspectratio]{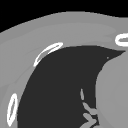}
& \includegraphics[width=2.1cm,height=2.1cm,keepaspectratio]{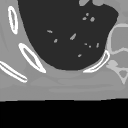}
& \multirow{2}{*}[5.75em]{
\includegraphics[scale=0.54, trim=0.2em 0.2em 1.8em 1.5em, clip]{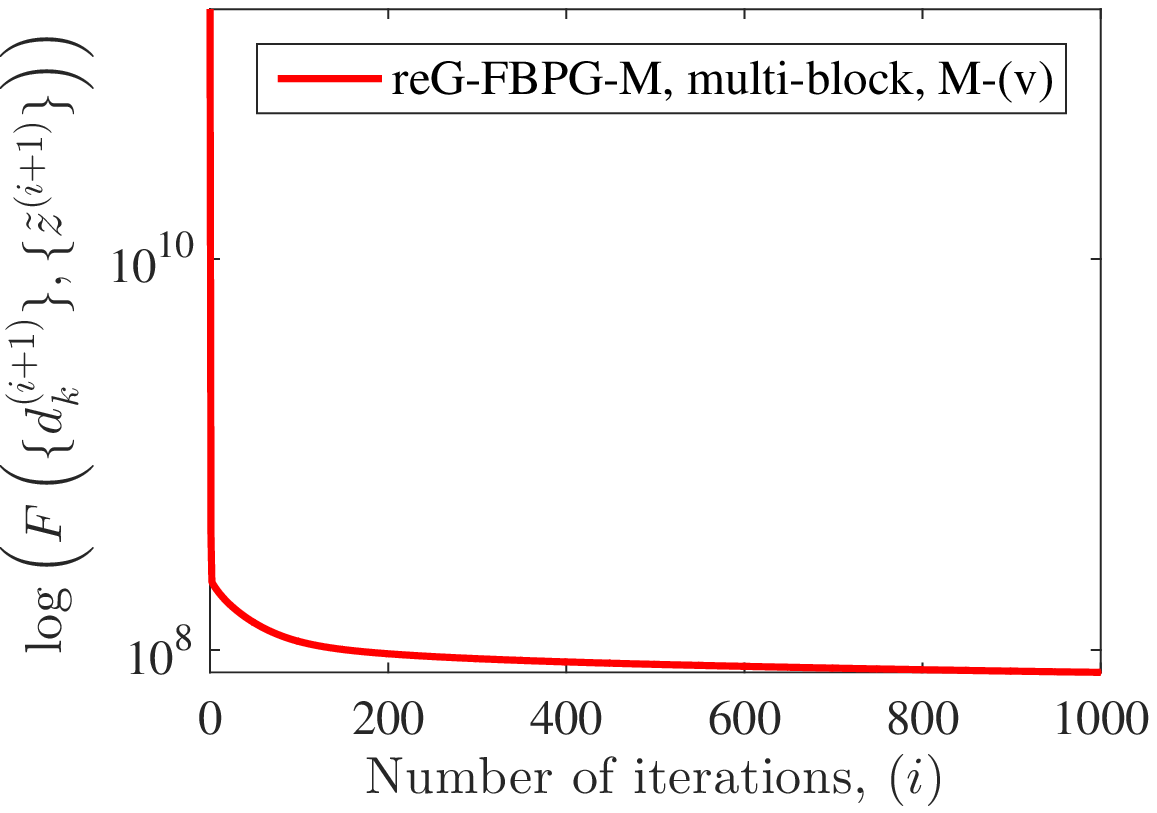} 
} 
\\
& \includegraphics[width=2.1cm,height=2.1cm,keepaspectratio]{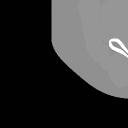} 
& \includegraphics[width=2.1cm,height=2.1cm,keepaspectratio]{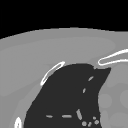}
\\
& {\small $\cdots$} & {\small $\cdots$} & \vspace{-0.5em} \\
\multicolumn{4}{c}{\small (d) The CT-(\romnum{2}) dataset ($L = 80$, $N = 128 \!\times\! 128$)} 
\end{tabular}

\caption{Examples of CDL results by the proposed reG-FBPG-M algorithm (using the multi-block scheme and M-(\romnum{5})) from different datasets. 
While guaranteeing convergence, the BPG-M methods provide desirable solutions: \textit{1)} the learned (Gabor-like) filters capture structures of training images; \textit{2)} the corresponding sparse codes have sparsity less than $1$\% (for (d), approximately $5$\%); \textit{3)} the resultant filters and sparse codes properly synthesize the training images. The sparsity is measured by $\sum_{l=1}^L ( \| \tilde{z}_l \| / \tilde{N} K )$ in percentages. The synthesized images mean $\{ P_B \sum_{k=1}^K d_{k}^\star \circledast z_{l,k}^\star : l=1,\ldots,L \}$.}
\label{fig:filters_BPGM}
\end{figure*}

\begin{figure*}[!pt]
\centering
\small\addtolength{\tabcolsep}{-2pt}
\renewcommand{\arraystretch}{0.9}

    \begin{tabular}{C{5cm}C{5cm}C{5cm}}
        
        \begin{tikzpicture}
            \begin{scope}[spy using outlines={rectangle,yellow,magnification=1.75,size=18mm, connect spies}]
                \node {\includegraphics[scale=0.5]{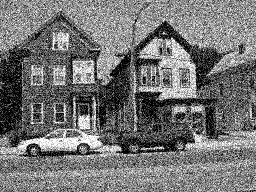}};
                \spy on (-1.2,-0.6) in node [left] at (2.41,-0.95);
                \node [yellow] at (-1.15,-1.5) {\scriptsize $\textmd{PSNR} = 16.04$~dB};
            \end{scope}
        \end{tikzpicture} &
        
        \begin{tikzpicture}
            \begin{scope}[spy using outlines={rectangle,yellow,magnification=1.75,size=18mm, connect spies}]
                \node {\includegraphics[scale=0.5]{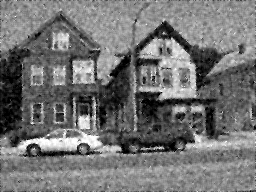}};
                \spy on (-1.2,-0.6) in node [left] at (2.41,-0.95);
	        \node [yellow] at (-1.15,-1.5) {\scriptsize $\textmd{PSNR} = 22.02$~dB};
            \end{scope}
        \end{tikzpicture} &

        \begin{tikzpicture}
            \begin{scope}[spy using outlines={rectangle,yellow,magnification=1.75,size=18mm, connect spies}]
                \node {\includegraphics[scale=0.5]{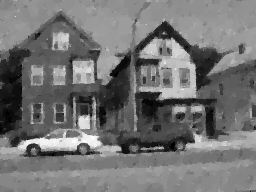}};
                \spy on (-1.2,-0.6) in node [left] at (2.41,-0.95);
	        \node [yellow] at (-1.15,-1.5) {\scriptsize $\textmd{PSNR} = 23.66$~dB};
            \end{scope}
        \end{tikzpicture} \\
       
        \small{(a) Noisy image} & \small{(b) Wiener filtering} & \small{(c) TV denoiser} \\
        
         \begin{tikzpicture}
            \begin{scope}[spy using outlines={rectangle,yellow,magnification=1.75,size=18mm, connect spies}]
                \node {\includegraphics[scale=0.5]{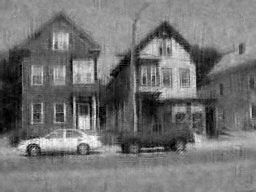}};
                \spy on (-1.2,-0.6) in node [left] at (2.41,-0.95);
                \node [yellow] at (-1.15,-1.5) {\scriptsize $\textmd{PSNR} = 22.36$~dB};
            \end{scope}
        \end{tikzpicture} &
        
        \begin{tikzpicture}
            \begin{scope}[spy using outlines={rectangle,yellow,magnification=1.75,size=18mm, connect spies}]
                \node {\includegraphics[scale=0.5]{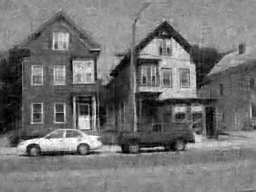}};
                \spy on (-1.2,-0.6) in node [left] at (2.41,-0.95);
                \node [yellow] at (-1.15,-1.5) {\scriptsize $\textmd{PSNR} = 23.94$~dB};
            \end{scope}
        \end{tikzpicture} &
         
        \begin{tikzpicture}
            \begin{scope}[spy using outlines={rectangle,yellow,magnification=1.75,size=18mm, connect spies}]
                \node {\includegraphics[scale=0.5]{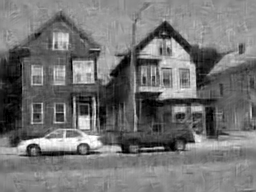}};
                \spy on (-1.2,-0.6) in node [left] at (2.41,-0.95);
                \node [yellow] at (-1.15,-1.5) {\scriptsize $\textmd{PSNR} = 24.11$~dB};
            \end{scope}
        \end{tikzpicture} \\
        
        \specialcell[c]{\small{(d) CDL image denoiser \R{eq:sys:denoising} using} \\ \small{learned filters via ADMM CDL \R{eq:sys:synth}}} & 
        \specialcell[c]{\small{(e) CDL image denoiser \R{eq:sys:denoising} using} \\ \small{learned filters via BPG-M CDL \R{eq:sys:synth}}} & 
        \specialcell[c]{\small{(f) CDL image denoiser \R{eq:sys:denoising} using} \\ \small{learned filters via BPG-M CDL \R{eq:sys:CDLadtRe}}} \\

    \end{tabular}
    
\caption{Comparison of denoised images from different image denoising models (image is corrupted by AWGN with $\textmd{SNR} \!=\! 10$~dB; for ADMM \cite{Heide&eta:15CVPR}, we used (empirically) convergent learned filters; for BPG-M, we used the two-block reG-FBPG-M method using \R{eq:step_size2}). 
The image denoising model \R{eq:sys:denoising} using the learned filters by BPG-M-based CDL---(e)---shows better image denoising performance compared to (b) Wiener filtering, (c) TV denoising, and (d) that using the learned filters by ADMM-based CDL.
The filters trained by CDL-ACE further improves (e)-image denoiser.}
\label{fig:denoise}

\end{figure*}

Figs.~\ref{fig:Comp_MConvg_BPG}--\ref{fig:filters_BPGM} and \ref{fig:filters_ADMM} illustrate that all the proposed BPG-M algorithms converge to desirable solutions and reach lower objective function values than the ADMM approach in \cite{Heide&eta:15CVPR} and its memory-efficient variant (see Table~\ref{tab:Comp:obj:ALMvsBPG-M}  and compare Fig.~\ref{fig:filters_BPGM}(d) to Fig.~\ref{fig:filters_ADMM}(c)).
In particular, the tighter majorizer enables the BPG-M algorithms to converge faster; see Figs.~\ref{fig:Comp_MConvg_BPG}--\ref{fig:Comp_newBPGM}.
Interestingly, the restarting schemes \R{eq:restart:func}--\R{eq:restart:grad} provide significant convergence acceleration over the momentum coefficient formula \R{eq:step_size1}.  The combination of reG \R{eq:restart:grad} and momentum coefficient formulas \R{eq:step_size1}--\R{eq:step_size2}, i.e., reG-FBPG-M, can be useful in accelerating the convergence rate of reG-BPG-M, particularly when majorizers are insufficiently tight. 
Fig.~\ref{fig:Comp_Convg_BPG} supports these assertions.
Most importantly, all the numerical experiments regarding the BPG-M methods are in good agreement with our theoretical results on the convergence analysis, e.g., Theorem \ref{t:Nash_convg} and Remark \ref{r:global_convg}.
Finally, the results in Table~\ref{tab:Comp:obj:ALMvsBPG-M} concur with the existing empirical results of comparison between BPG and BCD in \cite{Xu&Yin:13SIAM, Xu&Yin:16IPI}, noting that ADMM in \cite{Heide&eta:15CVPR} is BCD-type method.

\subsection{Application of Learned Filters by CDL to Image Denoising} \label{sec:result:img_denoise}

The filters learned via convergent BPG-M-based CDL \R{eq:sys:synth} show better image denoising performance than the (empirically) convergent ones trained by ADMM-based CDL \R{eq:sys:synth} in \cite{Heide&eta:15CVPR}; it improves PSNR by approximately $1.6$~dB.
Considering that the BPG-M methods reach lower objective values than ADMM of \cite{Heide&eta:15CVPR}, this implies that the filters of lower objective values can improve the CDL-based image denoiser \R{eq:sys:denoising}.
The learned filters by CDL-ACE \R{eq:sys:CDLadtRe} further improve image denoising compared to those trained by BPG-M-based CDL \R{eq:sys:synth}, by resolving the model mismatch; it improves PSNR by approximately $0.2$~dB.
Combining these, the CDL-based image denoiser using the learned filters by CDL-ACE \R{eq:sys:CDLadtRe} outperforms the TV denoising model. All these assertions are supported by Fig.~\ref{fig:denoise}.
Finally, the CDL-ACE model \R{eq:sys:CDLadt} better captures structures of non-preprocessed training images than the CDL model \R{eq:sys:synth}; see \cite[Fig.~2]{Chun&Fessler:17SAMPTA}.

\section{Conclusion} \label{sec:conclusion}

Developing convergent and stable algorithms for non-convex problems is important and challenging.
In addition, parameter tuning is a known challenge for AL methods.
This paper has considered both algorithm acceleration and the above two important issues for CDL.

The proposed BPG-M methods have several benefits over the ADMM approach in \cite{Heide&eta:15CVPR} and its memory-efficient variant.
First, the BPG-M algorithms guarantee local convergence (or global convergence if some conditions are satisfied) without additional parameter tuning (except regularization parameter). The BPG-M methods converge stably and empirically to a \dquotes{desirable} solution regardless of the datasets.
Second, particularly with the multi-block framework, they are useful for larger datasets due to their lower memory requirement and no polynomial computational complexity (specifically, no $\cO (L^2 K \tilde{N})$, $\cO (K^2 L)$, and $\cO(K^3 \tilde{N})$ complexity).
Third, they empirically achieve lower objective values. 
Among the proposed BPG-M algorithms, the reG-FBPG-M scheme---i.e., BPG-M using gradient-mapping-based restarting and momentum coefficient formulas---is practically useful by due to its fast convergence rate and no requirements in objective value evaluation.
The CDL-based image denoiser using learned filters via BPG-M-based CDL-ACE \cite{Chun&Fessler:17SAMPTA} outperforms Wiener filtering, TV denoising, and filters trained by the conventional ADMM-based CDL \cite{Heide&eta:15CVPR}.
The proposed BPG-M algorithmic framework is a reasonable choice towards stable and fast convergent algorithm development in CDL with big data (i.e., training data with the large number of signals or high-dimensional signals).

There are a number of avenues for future work. First, in this paper, the global convergence guarantee in Remark \ref{r:global_convg} requires a stringent condition in practice. 
Future work will explore the more general global convergence guarantee based on the Kurdyka-{\L}ojasiewicz property.
Second, we expect to further accelerate BPG-M by using the stochastic gradient method while guaranteeing its convergence (the stochastic ADMM \cite{Zhong&Kwok:14ICML} improves the convergence rate of ADMM on convex problems, and is applied to convolutional sparse coding for image super-resolution \cite{Gu&etal:15ICCV}). Applying the proposed CDL algorithm to multiple-layer setup is an interesting topic for future work \cite{Papyan&Romano&Elad:16arXiv-CNN}.
On the application side, we expect that incorporating normalization by local variances into CDL-ACE will further improve solutions to inverse problems.

\section*{Appendix: Notation}

We use $\nm{\cdot}_{p}$ to denote the $\ell^p$-norm and write $\ip{\cdot}{\cdot}$ for the standard inner product on $\bbC^N$.  
The weighted $\ell^2$-norm with a Hermitian positive definite matrix $A$ is denoted by $\nm{\cdot}_{A} = \nm{ A^{1/2} (\cdot) }_2$.
$\nm{\cdot}_{0}$ denotes the $\ell^0$-norm, i.e., the number of nonzeros of a vector.  
$( \cdot )^T$, $( \cdot )^H$, and $( {\cdot} )^*$ indicate the transpose, complex conjugate transpose (Hermitian transpose), and complex conjugate, respectively. 
$\diag(\cdot)$ and $\sgn(\cdot)$ denote the conversion of a vector into a diagonal matrix or diagonal elements of a matrix into a vector and the sign function, respectively.
$\otimes$, $\odot$, and $\bigoplus$ denote Kronecker product for two matrices, element-wise multiplication in a vector or a matrix, and the matrix direct sum of square matrices, respectively. 
$[C]$ denotes the set $\{1,2,\ldots,C\}$.  
For self-adjoint matrices $A,B \in \bbC^{N \times N}$, the notation $B \preceq A$ denotes that $A-B$ is a positive semi-definite matrix.

\section*{Acknowledgment}

We thank Dr. Donghwan Kim and Dr. Jonghoon Jin for constructive feedback.

\bibliographystyle{IEEEtran}
\bibliography{referenceBibs_Bobby}

\begin{thebibliography}{10}
\providecommand{\url}[1]{#1}
\csname url@samestyle\endcsname
\providecommand{\newblock}{\relax}
\providecommand{\bibinfo}[2]{#2}
\providecommand{\BIBentrySTDinterwordspacing}{\spaceskip=0pt\relax}
\providecommand{\BIBentryALTinterwordstretchfactor}{4}
\providecommand{\BIBentryALTinterwordspacing}{\spaceskip=\fontdimen2\font plus
\BIBentryALTinterwordstretchfactor\fontdimen3\font minus
  \fontdimen4\font\relax}
\providecommand{\BIBforeignlanguage}[2]{{%
\expandafter\ifx\csname l@#1\endcsname\relax
\typeout{** WARNING: IEEEtran.bst: No hyphenation pattern has been}%
\typeout{** loaded for the language `#1'. Using the pattern for}%
\typeout{** the default language instead.}%
\else
\language=\csname l@#1\endcsname
\fi
#2}}
\providecommand{\BIBdecl}{\relax}
\BIBdecl

\bibitem{Bruckstein&Dohono&Elad:09SIAM}
A.~M. Bruckstein, D.~L. Donoho, and M.~Elad, ``From sparse solutions of systems
  of equations to sparse modeling of signals and images,'' \emph{SIAM Rev.},
  vol.~51, no.~1, pp. 34--81, Feb. 2009.

\bibitem{Coates&NG:bookCh}
A.~Coates and A.~Y. Ng, ``Learning feature representations with {K}-means,'' in
  \emph{Neural Networks: Tricks of the Trade, 2nd ed., LNCS 7700}, G.~Montavon,
  G.~B. Orr, and K.-R. M\"{u}ller, Eds.\hskip 1em plus 0.5em minus 0.4em\relax
  Berlin: Springer Verlag, 2012, ch.~22, pp. 561--580.

\bibitem{Mairal&Bach&Ponce:14FTCGV}
J.~Mairal, F.~Bach, and J.~Ponce, ``Sparse modeling for image and vision
  processing,'' \emph{Found. \& Trends in Comput. Graph. Vis.}, vol.~8, no.
  2--3, pp. 85--283, Dec. 2014.

\bibitem{Aharon&Elad&Bruckstein:06TSP}
M.~Aharon, M.~Elad, and A.~Bruckstein, ``\textit{K}-{SVD}: An algorithm for
  designing overcomplete dictionaries for sparse representation,'' \emph{IEEE
  Trans. Signal Process.}, vol.~54, no.~11, pp. 4311--4322, Nov. 2006.

\bibitem{Xu&Yin:16IPI}
Y.~Xu and W.~Yin, ``A fast patch-dictionary method for whole image recovery,''
  \emph{Inverse Probl. Imag.}, vol.~10, no.~2, pp. 563--583, May 2016.

\bibitem{Heide&eta:15CVPR}
F.~Heide, W.~Heidrich, and G.~Wetzstein, ``Fast and flexible convolutional
  sparse coding,'' in \emph{Proc. $2015$ IEEE CVPR}, Boston, MA, Jun. 2015, pp.
  5135--5143.

\bibitem{Olshausen&Field:96Nature}
B.~A. Olshausen and D.~J. Field, ``Emergence of simple-cell receptive field
  properties by learning a sparse code for natural images,'' \emph{Nature},
  vol. 381, no. 6583, pp. 607--609, Jun. 1996.

\bibitem{Olshausen&Field:97VR}
------, ``Sparse coding with an overcomplete basis set: A strategy employed by
  {V}1?'' \emph{Vision Res.}, vol.~37, no.~23, pp. 3311--3325, Dec. 1997.

\bibitem{LeCun&Bengio&Hinton:15Nature}
Y.~LeCun, Y.~Bengio, and G.~Hinton, ``Deep learning,'' \emph{Nature}, vol. 521,
  no. 7553, pp. 436--444, 2015.

\bibitem{LeCun&etal:98ProcIEEE}
Y.~LeCun, L.~Bottou, Y.~Bengio, and P.~Haffner, ``Gradient-based learning
  applied to document recognition,'' \emph{Proc. IEEE}, vol.~86, no.~11, pp.
  2278--2324, Nov. 1998.

\bibitem{Krizhevsky&etal:12NIPS}
A.~Krizhevsky, I.~Sutskever, and G.~E. Hinton, ``Image{N}et classification with
  deep convolutional neural networks,'' in \emph{Proc. NIPS $2012$, Advances in
  Neural Information Processing Systems $25$}, 2012, pp. 1097--1105.

\bibitem{Zeiler&etal:10CVPR}
M.~D. Zeiler, D.~Krishnan, G.~W. Taylor, and R.~Fergus, ``Deconvolutional
  networks,'' in \emph{Proc. $2010$ IEEE CVPR}, San Francisco, CA, Jun. 2010,
  pp. 2528--2535.

\bibitem{Kavukcuoglu&etal:10NIPS}
K.~Kavukcuoglu, P.~Sermanet, Y.-L. Boureau, K.~Gregor, M.~Mathieu, and Y.~L.
  Cun, ``Learning convolutional feature hierarchies for visual recognition,''
  in \emph{Proc. NIPS $2010$, Advances in Neural Information Processing Systems
  $23$}, 2010, pp. 1090--1098.

\bibitem{Bristow&etal:13CVPR}
H.~Bristow, A.~Eriksson, and S.~Lucey, ``Fast convolutional sparse coding,'' in
  \emph{Proc. $2013$ IEEE CVPR}, Portland, OR, Jun. 2013, pp. 391--398.

\bibitem{Wohlberg:14ICASSP}
B.~Wohlberg, ``Efficient convolutional sparse coding,'' in \emph{Proc.
  $39^{\text{th}}$ IEEE ICASSP}, Florence, Italy, May 2014, pp. 7173--7177.

\bibitem{Kong&Fowkes:14techRep}
B.~Kong and C.~C. Fowlkes, ``Fast convolutional sparse coding ({FCSC}),''
  \emph{Tech. Rep, Department of Computer Science, University of California,
  Irvine, Available:
  \url{http://vision.ics.uci.edu/papers/KongF_TR_2014/KongF_TR_2014.pdf}},
  vol.~3, 2014.

\bibitem{Bristow&Lucey:14arXiv}
H.~Bristow and S.~Lucey, ``Optimization methods for convolutional sparse
  coding,'' \emph{arXiv preprint cs.CV:1406.2407}, Jun. 2014.

\bibitem{Wohlberg:16TIP}
B.~Wohlberg, ``Efficient algorithms for convolutional sparse representations,''
  \emph{IEEE Trans. Image Process.}, vol.~25, no.~1, pp. 301--315, Jan. 2016.

\bibitem{Wohlberg:16ICIP}
------, ``Boundary handling for convolutional sparse representations,'' in
  \emph{Proc. $23^{\text{rd}}$ IEEE ICIP}, Phoenix, AZ, Sep. 2016, pp.
  1833--1837.

\bibitem{Sorel&Michal:16DSP}
M.~{\v{S}}orel and F.~{\v{S}}roubek, ``Fast convolutional sparse coding using
  matrix inversion lemma,'' \emph{Digit. Signal Process.}, vol.~55, pp. 44--51,
  Aug. 2016.

\bibitem{Papyan&Sulam&Elad:16arXiv-part2}
V.~Papyan, J.~Sulam, and M.~Elad, ``Working locally thinking globally -- part
  {II}: Stability and algorithms for convolutional sparse coding,'' \emph{arXiv
  preprint cs.IT:1607.02009}, 2016.

\bibitem{Chun&Fessler:17SAMPTA}
I.~Y. Chun and J.~A. Fessler, ``Convergent convolutional dictionary learning
  using adaptive contrast enhancement ({CDL-ACE}): Application of cdl to image
  denoising,'' in \emph{\emph{to appear in} Proc. $12^{\textmd{th}}$ Sampling
  Theory and Appl. (SampTA)}, Tallinn, Estonia, Jul. 2017.

\bibitem{Papyan&Romano&Elad:16arXiv-CNN}
V.~Papyan, Y.~Romano, and M.~Elad, ``Convolutional neural networks analyzed via
  convolutional sparse coding,'' \emph{arXiv preprint cs.ML:1607.08194}, 2016.

\bibitem{Boyd&Parikh&Chu&Peleato&Eckstein:11FTML}
S.~Boyd, N.~Parikh, E.~Chu, B.~Peleato, and J.~Eckstein, ``Distributed
  optimization and statistical learning via the alternating direction method of
  multipliers,'' \emph{Found. \& Trends in Machine Learning}, vol.~3, no.~1,
  pp. 1--122, Jan. 2011.

\bibitem{Parikh&Boyd:14FTO}
N.~Parikh and S.~P. Boyd, ``Proximal algorithms.'' \emph{Found. Trends in
  Optim.}, vol.~1, no.~3, pp. 127--239, Jan. 2014.

\bibitem{Chalasani&etal:13IJCNN}
R.~Chalasani, J.~C. Principe, and N.~Ramakrishnan, ``A fast proximal method for
  convolutional sparse coding,'' in \emph{Proc. $2013$ IEEE IJCNN}, Dallas, TX,
  Aug. 2013, pp. 1--5.

\bibitem{Beck&Teboulle:09SIAM}
A.~Beck and M.~Teboulle, ``A fast iterative shrinkage-thresholding algorithm
  for linear inverse problems,'' \emph{SIAM J. Imaging Sci.}, vol.~2, no.~1,
  pp. 183--202, Mar. 2009.

\bibitem{Chen&etal:16ICIP}
B.~Chen, J.~Li, B.~Ma, and G.~Wei, ``Convolutional sparse coding classification
  model for image classification,'' in \emph{Proc. $23^{\text{rd}}$ IEEE ICIP},
  Phoenix, AZ, Sep. 2016, pp. 1918--1922.

\bibitem{Matakos&etal:13TIP}
A.~Matakos, S.~Ramani, and J.~A. Fessler, ``Accelerated edge-preserving image
  restoration without boundary artifacts,'' \emph{IEEE Trans. Image Process.},
  vol.~22, no.~5, pp. 2019--2029, May 2013.

\bibitem{Almeida&Figueiredo:13TIP}
M.~S. Almeida and M.~Figueiredo, ``Deconvolving images with unknown boundaries
  using the alternating direction method of multipliers,'' \emph{IEEE Trans.
  Image Process.}, vol.~22, no.~8, pp. 3074--3086, Aug. 2013.

\bibitem{Wohlberg:sporco}
B.~Wohlberg, ``{SP}arse {O}ptimization {R}esearch {CO}de ({SPORCO}),''
  \emph{\emph{[Online]. Available:
  \url{http://purl.org/brendt/software/sporco}}}, 2017.

\bibitem{Xu&Yin:13SIAM}
Y.~Xu and W.~Yin, ``A block coordinate descent method for regularized
  multiconvex optimization with applications to nonnegative tensor
  factorization and completion,'' \emph{SIAM J. Imaging Sci.}, vol.~6, no.~3,
  pp. 1758--1789, Sep. 2013.

\bibitem{Fessler&etal:93TNS}
J.~A. Fessler, N.~H. Clinthorne, and W.~L. Rogers, ``On complete-data spaces
  for {PET} reconstruction algorithms,'' \emph{IEEE Trans. Nucl. Sci.},
  vol.~40, no.~4, pp. 1055--1061, Aug. 1993.

\bibitem{Kruger:03JMS}
A.~Y. Kruger, ``On {F}r{\'e}chet subdifferentials,'' \emph{J. Math Sci.}, vol.
  116, no.~3, pp. 3325--3358, Jul. 2003.

\bibitem{Rockafellar&Wets:book}
R.~T. Rockafellar and R.~J.-B. Wets, \emph{Variational analysis}.\hskip 1em
  plus 0.5em minus 0.4em\relax Berlin: Springer Verlag, 2009, vol. 317.

\bibitem{Nesterov:07CORE}
Y.~Nesterov, ``Gradient methods for minimizing composite objective function,''
  \emph{\emph{CORE Discussion Papers - 2007/76, UCL, Available:
  \url{http://www.uclouvain.be/cps/ucl/doc/core/documents/Composit.pdf}}},
  2007.

\bibitem{Tseng:08techRep}
P.~Tseng, ``On accelerated proximal gradient methods for convex-concave
  optimization,'' \emph{Tech. Rep., Available:
  \url{http://www.mit.edu/~dimitrib/PTseng/papers/apgm.pdf}}, May 2008.

\bibitem{ODonoghue&Candes:15FCM}
B.~O'Donoghue and E.~Cand\`{e}s, ``Adaptive restart for accelerated gradient
  schemes,'' \emph{Found. Comput. Math.}, vol.~15, no.~3, pp. 715--732, Jun.
  2015.

\bibitem{Giselsson&Boyd:14CDC}
P.~Giselsson and S.~Boyd, ``Monotonicity and restart in fast gradient
  methods,'' in \emph{Proc. $53^{\text{rd}}$ IEEE CDC}, Los Angeles, CA, Dec.
  2014, pp. 5058--5063.

\bibitem{Xu&Yin:14arXiv}
Y.~Xu and W.~Yin, ``A globally convergent algorithm for nonconvex optimization
  based on block coordinate update,'' \emph{arXiv preprint math.OC:1410.1386},
  2014.

\bibitem{Muckley&Noll&Fessler:15TMI}
M.~J. Muckley, D.~C. Noll, and J.~A. Fessler, ``Fast parallel {MR} image
  reconstruction via {B}1-based, adaptive restart, iterative soft thresholding
  algorithms ({BARISTA}),'' \emph{IEEE Trans. Med. Imag.}, vol.~34, no.~2, pp.
  578--588, Feb. 2015.

\bibitem{Chun&etal:15TCI}
I.~Y. Chun, S.~Noh, D.~J. Love, T.~M. Talavage, S.~Beckley, and S.~J.~K.
  Kisner, ``Mean squared error ({MSE})-based excitation pattern design for
  parallel transmit and receive {SENSE} {MRI} image reconstruction,''
  \emph{IEEE Trans. Comput. Imag.}, vol.~2, no.~1, pp. 424--439, Dec. 2016.

\bibitem{Richtarik&Takac:14MP}
P.~Richt{\'a}rik and M.~Tak{\'a}{\v{c}}, ``Iteration complexity of randomized
  block-coordinate descent methods for minimizing a composite function,''
  \emph{Math. Program.}, vol. 144, no. 1-2, pp. 1--38, Apr. 2014.

\bibitem{Ravishankar&Nadakuditi&Fessler:17TCI}
S.~Ravishankar, R.~R. Nadakuditi, and J.~Fessler, ``Efficient sum of outer
  products dictionary learning ({SOUP}-{DIL}) and its application to inverse
  problems,'' \emph{IEEE Trans. Comput. Imag.}, vol.~PP, no.~99, pp. 1--1,
  2017.

\bibitem{Serrano&etal:16CGF}
A.~Serrano, F.~Heide, D.~Gutierrez, G.~Wetzstein, and B.~Masia, ``Convolutional
  sparse coding for high dynamic range imaging,'' \emph{Comput. Graph. Forum},
  vol.~35, no.~2, pp. 153--163, May 2016.

\bibitem{Segars&etal:08MP}
W.~P. Segars, M.~Mahesh, T.~J. Beck, E.~C. Frey, and B.~M. Tsui, ``Realistic
  {CT} simulation using the {4D} {XCAT} phantom,'' \emph{Med. Phys.}, vol.~35,
  no.~8, pp. 3800--3808, Jul. 2008.

\bibitem{Zeiler&etal:11CVPR}
M.~D. Zeiler, G.~W. Taylor, and R.~Fergus, ``Adaptive deconvolutional networks
  for mid and high level feature learning,'' in \emph{Proc. $2011$ IEEE CVPR},
  Colorado Springs, CO, Jun. 2011, pp. 2018--2025.

\bibitem{Jarrett&etal:09ICCV}
K.~Jarrett, K.~Kavukcuoglu, Y.~LeCun \emph{et~al.}, ``What is the best
  multi-stage architecture for object recognition?'' in \emph{Proc. $2009$
  ICCV}, Kyoto, Japan, Sep. 2009, pp. 2146--2153.

\bibitem{Beck&Teboulle:09TIP}
A.~Beck and M.~Teboulle, ``Fast gradient-based algorithms for constrained total
  variation image denoising and deblurring problems,'' \emph{IEEE Trans. Image
  Process.}, vol.~18, no.~11, pp. 2419--2434, Nov. 2009.

\bibitem{Zhong&Kwok:14ICML}
W.~Zhong and J.~T.-Y. Kwok, ``Fast stochastic alternating direction method of
  multipliers.'' in \emph{Proc. $2014$ IEEE ICML}, Beijing, China, Jun. 2014,
  pp. 46--54.

\bibitem{Gu&etal:15ICCV}
S.~Gu, W.~Zuo, Q.~Xie, D.~Meng, X.~Feng, and L.~Zhang, ``Convolutional sparse
  coding for image super-resolution,'' in \emph{Proc. $2015$ IEEE ICCV},
  Santiago, Chile, Dec. 2015, pp. 1823--1831.

\end{thebibliography}


\begin{thebibliography}{10}
\providecommand{\url}[1]{#1}
\csname url@samestyle\endcsname
\providecommand{\newblock}{\relax}
\providecommand{\bibinfo}[2]{#2}
\providecommand{\BIBentrySTDinterwordspacing}{\spaceskip=0pt\relax}
\providecommand{\BIBentryALTinterwordstretchfactor}{4}
\providecommand{\BIBentryALTinterwordspacing}{\spaceskip=\fontdimen2\font plus
\BIBentryALTinterwordstretchfactor\fontdimen3\font minus
  \fontdimen4\font\relax}
\providecommand{\BIBforeignlanguage}[2]{{%
\expandafter\ifx\csname l@#1\endcsname\relax
\typeout{** WARNING: IEEEtran.bst: No hyphenation pattern has been}%
\typeout{** loaded for the language `#1'. Using the pattern for}%
\typeout{** the default language instead.}%
\else
\language=\csname l@#1\endcsname
\fi
#2}}
\providecommand{\BIBdecl}{\relax}
\BIBdecl

\bibitem{Chun&Fessler:supp}
I.~Y. Chun and J.~Fessler, ``Convolutional dictionary learning: Acceleration
  and convergence,'' \emph{\emph{submitted}}, Nov. 2016.

\bibitem{Xu&Yin:supp}
Y.~Xu and W.~Yin, ``A block coordinate descent method for regularized
  multiconvex optimization with applications to nonnegative tensor
  factorization and completion,'' \emph{SIAM J. Imaging Sci.}, vol.~6, no.~3,
  pp. 1758--1789, Sep. 2013.

\bibitem{Rockafellar:supp}
R.~T. Rockafellar, ``Monotone operators and the proximal point algorithm,''
  \emph{SIAM J. Control Optim.}, vol.~14, no.~5, pp. 877--898, Aug. 1976.

\bibitem{Boyd&Vandenberghe:supp}
S.~Boyd and L.~Vandenberghe, \emph{Convex Optimization}.\hskip 1em plus 0.5em
  minus 0.4em\relax New York, NY: Cambridge University Press, 2004.

\bibitem{Reinsch:supp}
C.~H. Reinsch, ``Smoothing by spline functions. {II},'' \emph{Numer. Math.},
  vol.~16, no.~5, pp. 451--454, Feb. 1971.

\bibitem{Heide&eta:supp}
F.~Heide, W.~Heidrich, and G.~Wetzstein, ``Fast and flexible convolutional
  sparse coding,'' in \emph{Proc. $2015$ IEEE CVPR}, Boston, MA, Jun. 2015, pp.
  5135--5143.

\bibitem{Boyd&Parikh&Chu&Peleato&Eckstein:supp}
S.~Boyd, N.~Parikh, E.~Chu, B.~Peleato, and J.~Eckstein, ``Distributed
  optimization and statistical learning via the alternating direction method of
  multipliers,'' \emph{Found. \& Trends in Machine Learning}, vol.~3, no.~1,
  pp. 1--122, Jan. 2011.

\bibitem{Wohlberg:supp}
B.~Wohlberg, ``Efficient algorithms for convolutional sparse representations,''
  \emph{IEEE Trans. Image Process.}, vol.~25, no.~1, pp. 301--315, Jan. 2016.

\bibitem{Goldstein&Osher:supp}
T.~Goldstein and S.~Osher, ``The split {B}regman method for {L}1-regularized
  problems,'' \emph{SIAM J. Imaging Sci.}, vol.~2, no.~2, pp. 323--343, Apr.
  2009.

\bibitem{Ramani&Fessler:supp}
S.~Ramani and J.~A. Fessler, ``Parallel {MR} image reconstruction using
  augmented {L}agrangian methods,'' \emph{IEEE Trans. Med. Imag.}, vol.~30,
  no.~3, pp. 694--706, Mar. 2011.

\end{thebibliography}


\renewcommand{\thefigure}{S.\arabic{figure}}
\renewcommand{\thetable}{S.\Roman{table}}
\renewcommand{\thesection}{S.\Roman{section}}

\renewcommand\thetheorem{S.\arabic{theorem}}
\renewcommand{\theequation}{S.\arabic{equation}}

\renewcommand{\thealgorithm}{S.\arabic{algorithm}} 

\setcounter{section}{0}
\setcounter{equation}{0}
\setcounter{theorem}{0}
\setcounter{footnote}{0}

{
\twocolumn[
\begin{center}
 \Huge Convolutional Dictionary Learning: Acceleration and Convergence: Supplementary Material
\vspace{0.2in}
\end{center}]
}

In this supplementary material for \citeSupp{Chun&Fessler:supp}, we provide mathematical proofs or detailed descriptions to support several arguments in the main manuscript.  We use the prefix ``S'' for the numbers in section, equation, figure, and table in the supplementary material.\let\thefootnote\relax\footnote{Supplementary material updated August $22^\text{nd}$, 2017.}

\section{Useful Lemmas and Their Proofs}

\lem{
\renewcommand\thetheorem{\Alph{subsection}.\arabic{theorem}}
\label{l:obj_ineq}
Let $\varrho_1(u)$ and $\varrho_2(u)$ be two convex functions defined on the convex set $\cU$, $\varrho_1(u)$ be differentiable, and $M \succ 0$. 
Let $\varrho(u) = \varrho_1(u) + \varrho_2(u)$ and $u^{\star} = \argmin_{u \in \cU}  \, \ip{ \nabla \varrho_1(v) }{ u-v } + \frac{1}{2} \nm{ u - v }_{M}^2 + \varrho_2(u)$. If
\be{
\label{eq:lem_conK}
\varrho_1 (u^{\star}) \leq \varrho_1 (u) +  \ip{ \nabla \varrho_1(v) }{ u^{\star}-v }  + \frac{1}{2} \nm{ u^{\star} - v }_{M}^2,
}
then we have
\be{
\label{eq:lem_result}
\varrho(u) - \varrho(u^{\star}) \leq \frac{1}{2} \nm{ u^{\star} - v }_{M}^2 + ( v - u )^T M (u^{\star} - v).
}
}
\prf{
The following proof is an extension of that given in \citeSupp{Xu&Yin:supp}.
The first-order optimality condition for $u^{\star} = \argmin_{u \in \cU} \,  \ip{ \nabla \varrho_1(v) }{ u-v } + \frac{1}{2} \nm{ u - v }_{M}^2 + \varrho_2(u)$ is given by
\be{
\label{eq:first_order_conK}
\ip{\nabla \varrho_1(v) + M (u^{\star} - v) + g^{\star}}{u - u^{\star}} \geq 0, \quad \mbox{for any $u \in \cU$}
}
and for some $g \in \partial \varrho_2 (u^{\star})$.
For any $u \in \cU$, we obtain
\eas{
& ~ \varrho(u) - \varrho(u^{\star}) 
\\
& \geq \varrho(u) - \left( \varrho_1(v) + \ip{ \nabla \varrho_1(v) }{ u^{\star} - v } + \frac{1}{2} \nm{u^{\star} - v}_L^2 \right) 
\\
& \quad - \varrho_2(u^{\star})
\\
&= \varrho_1 (u) - \varrho_1 (v) - \ip{ \nabla \varrho_1(v) }{ u - v } + \ip{ \nabla \varrho_1(v) }{ u - u^{\star} } 
\\
& \quad + \varrho_2 (u) - \varrho_2 (u^{\star}) - \frac{1}{2} \nm{ u^{\star} - v }_2^2
\\
&\geq \varrho_2(u) - \varrho_2(u^{\star}) - \ip{g^{\star}}{ u - u^{\star}} - ( u^{\star} - v )^T M ( u - u^{\star} ) 
\\
& \quad - \frac{1}{2} (u^{\star} - v)^T M (u^{\star} - v)
\\
&\geq - ( u^{\star} - v )^T M ( u - u^{\star} ) - \frac{1}{2} (u^{\star} - v)^T M (u^{\star} - v)
\\
&= \frac{1}{2} \nm{u^{\star} - v}_L^2 + (v - u)^T M (u^{\star} - v)
}
where the first inequality comes from \R{eq:lem_conK}, the second inequality is obtained by convexity of $\varrho_1$ (i.e., $\ip{ \nabla \varrho_1(v) }{ u - v } \leq \varrho_1(u) - \varrho_1(v)$) and \R{eq:first_order_conK}, and the last inequality is obtained by the convexity of $\varrho_2$ and the fact $g^{\star} \in \partial \varrho_2 (u^{\star})$ (i.e., $\ip{g^{\star}}{u - u^{\star}} \leq \varrho_2(u) - \varrho_2(u^{\star}) \leq 0$). This completes the proof.
}

\lem{
\label{l:diag(|psdA|1)}
If the diagonal elements of a Hermitian matrix $A$ are nonnegative (e.g., if $A$ is positive semidefinite), then $A \preceq \diag(|A| 1)$, where $|A|$ denotes the matrix consisting of the absolute values of the elements of $A$.
}
\prf{
Let $E = \diag(|A| 1)-A$. 
We seek to apply the property that, if a Hermitian matrix is diagonally dominant with nonnegative diagonal entries, then it is positive semidefinite.
We first show that $E$ is diagonally dominant.
For $j = k$, we have
\be{
\label{eq:prf:l:diag(|psdA|1)}
E_{j,j} = \sum_{k} | A_{j,k} | - A_{j,j} = \sum_{k \neq j} | A_{j,k} |
}
due to the assumption of $A_{j,j} \geq 0$ in Lemma \ref{l:diag(|psdA|1)}. For $j \neq k$, $\sum_{k \neq j} \left| E_{j,k} \right| = \sum_{k \neq j} | A_{j,k} | = E_{j,j}$ where the first equality uses $E_{j,k}  = - A_{j,k}$ and the second equality uses \R{eq:prf:l:diag(|psdA|1)}.
It is straightforward to show that $E$ is a Hermitian matrix (due to $E_{j,k}  = - A_{j,k}$) with nonnegative diagonal entries (due to \R{eq:prf:l:diag(|psdA|1)}).
Combining these results completes the proof.
}

\lem{
\label{l:diag(|At||A|1)}
For a complex-valued matrix $A$, $A^H A  \preceq \diag( | A^H | | A | 1 )$.
}
\prf{
Let $E = \diag( | A^H | | A | 1 ) - A^H A$. The $j\rth$ diagonal element of $\diag( | A^H | | A | 1 )$ is $\sum_{l} | A_{l,j} | \sum_{k} | A_{l,k} |$.
We again seek to apply the property that, if a Hermitian matrix is diagonally dominant with nonnegative diagonal entries, then it is positive semidefinite.
For $j = k$, observe that
\ea{
\label{eq:prf:l:diag(|At||A|1)_diag}
E_{j,j} &=  \sum_{l} | A_{l,j} | \sum_{k} | A_{l,k} | - \sum_{l} | A_{l,j} |^2
\nn \\
&= \sum_{l} | A_{l,j} | \left( | A_{l,j} | + \sum_{k \neq j} | A_{l,k} | \right) - | A_{l,j} |^2
\nn \\
&= \sum_{l} | A_{l,j} | \sum_{k \neq j} | A_{l,k} |
 \\
&\geq 0 \nn
}
establishing nonnegative diagonal elements.
For $j \neq k$, it follows from the triangle inequality that
\be{
\label{eq:prf:l:diag(|At||A|1)_offdiag}
\sum_{k \neq j} \left| E_{j,k} \right| = \sum_{k \neq j} \left| \sum_{l} A_{l,k}^* A_{l,j} \right| \leq \sum_{k \neq j} \sum_{l}  | A_{l,k} | | A_{l,j} |.
}
Combining \R{eq:prf:l:diag(|At||A|1)_diag} and \R{eq:prf:l:diag(|At||A|1)_offdiag} gives $E_{j,j} - \sum_{k \neq j} \left| E_{j,k} \right| \geq 0$, $\forall j$.
Combining these results completes the proof.
}

\section{Proof of Proposition \ref{p:BPG_sq_sum}} \label{sec:prf:p:BPG_sq_sum}
The following proof extends that given in \citeSupp{Xu&Yin:supp}.
Let $F_b^{(i)} := f_b^{(i)} + r_b$ and $\varrho_1 = f_b^{(i)}$, $\varrho_2 = r_b$, $u = x_b^{(i)}$, $v = \acute{x}_b^{(i+1)}$, $u^{\star} = x_b^{(i+1)}$, and $M = M_b^{(i)}$, by applying Lemma \ref{l:obj_ineq} to \R{eq:assume:update}.
We first obtain the following bounds for $F_b^{(i)} (x_b^{(i)}) - F_b^{(i)} (x_b^{(i+1)})$:
\eas{
& ~ F_b^{(i)} (x_b^{(i)}) - F_b^{(i)} (x_b^{(i+1)}) 
\\
& \geq \frac{1}{2} \nm{ \acute{x}_b^{(i)} - x_b^{(i+1)}  }_{M_b^{(i)}}^2 
\\
& \quad + \left( \acute{x}_b^{(i)} - x_b^{(i+1)} \right)^T M_b^{(i)} \left( x_b^{(i)} - \acute{x}_b^{(i)} \right)
\\
& = \frac{1}{2} \nm{ x_b^{(i)} - x_b^{(i+1)} }_{M_b^{(i)}}^2 - \frac{1}{2} \nm{ W_b^{(i)} \left( x_b^{(i-1)} - x_b^{(i)} \right)  }_{M_b^{(i)}}^2
\\
& \geq \frac{1}{2} \nm{ x_b^{(i)} - x_b^{(i+1)} }_{M_b^{(i)}}^2 - \frac{\delta^2}{2} \nm{ x_b^{(i-1)} - x_b^{(i)} }_{M_b^{(i-1)}}^2 
}
where the first inequality is obtained by using \R{eq:lem_result} in Lemma \ref{l:obj_ineq}, the first equality uses the symmetry of $M_b^{(i)}$, and the second inequality holds by
\bes{
\delta^2 \left( W_b^{(i)} \right)^T M_b^{(i)} W_b^{(i)} \preceq \delta^2 M_b^{(i-1)}
}
due to Assumption 3.
Summing the following inequality of $F(x^{(i)}) - F(x^{(i+1)})$
\begingroup
\setlength{\thinmuskip}{1.5mu}
\setlength{\medmuskip}{2mu plus 1mu minus 2mu}
\setlength{\thickmuskip}{2.5mu plus 2.5mu}
\eas{
& ~ F(x^{(i)}) - F(x^{(i+1)}) 
\\
& = \sum_{b=1}^B F_b^{(i)} (x_b^{(i)}) - F_b^{(i)} (x_b^{(i+1)}) 
\\
& \geq \sum_{b=1}^B \frac{1}{2} \nm{ x_b^{(i)} - x_b^{(i+1)} }_{M_b^{(i)}}^2 - \frac{\delta^2}{2} \nm{ x_b^{(i-1)} - x_b^{(i)} }_{M_b^{(i-1)}}^2
}
\endgroup
over $i = 1,\ldots,\mathrm{Iter}$, we have
\begingroup
\setlength{\thinmuskip}{1.5mu}
\setlength{\medmuskip}{2mu plus 1mu minus 2mu}
\setlength{\thickmuskip}{2.5mu plus 2.5mu}
\eas{
& ~ F(x^{(0)}) - F(x^{(\mathrm{Iter}+1)}) 
\\
& \geq \sum_{i=1}^{\mathrm{Iter}} \sum_{b=1}^B \frac{1}{2} \nm{ x_b^{(i)} - x_b^{(i+1)} }_{M_b^{(i)}}^2 - \frac{\delta^2}{2} \nm{ x_b^{(i-1)} - x_b^{(i)} }_{M_b^{(i-1)}}^2
\\ 
& \geq  \sum_{i=1}^{\mathrm{Iter}} \sum_{b=1}^B \frac{1 - \delta^2}{2} \nm{ x_b^{(i)} - x_b^{(i+1)} }_{M_b^{(i)}}^2
\\
& \geq  \sum_{i=1}^{\mathrm{Iter}} \frac{\left( 1 - \delta^2 \right) \beta}{2} \nm{ x^{(i)} - x^{(i+1)} }_2^2.
}
\endgroup
Due to the lower boundedness of $F$ in Assumption 1, taking $\mathrm{Iter} \rightarrow \infty$ completes the proof.

\section{Proof of Theorem \ref{t:Nash_convg}} \label{sec:prf:t:Nash_convg}
Let $\bar{x}$ be a limit point of $\{ x^{(i)} \}$ and $\{ x^{(i_j)} \}$ be the subsequence converging to $\bar{x}$.
Note that $\bar{x} \in \cX$ (due to the closedness of $\cX$), and $M_b^{(i_j)} \rightarrow M_b^{(i)}$ (taking another
subsequence if necessary) as $j \rightarrow \infty$ since $\{ M_b^{(i)} \}$ is bounded, for $b\in[B]$.
Using \R{eq:p:BPG_sq_sum:imply}, $\{ x^{(i_j+\iota)} \}$ converges to $\bar{x}$ for any $\iota \geq 0$.

Now we observe that 
\ea{
\label{eq:t:prf:nonconvg}
x_b^{(i_j+1)} 
& = \argmin_{ x_b \in \cX_b^{(i_j)} } \, \ip{ \nabla f_b^{(i_j)} (\acute{x}_b^{(i_j)}) }{ x_b - \acute{x}_b^{(i_j)} } 
\nn \\
& \qquad\qquad ~ + \frac{1}{2} \nm{ x_b - \acute{x}_b^{(i_j)} }_{M_b^{(i_j)}}^2 + r_b (x_b).
}
Note that the convex proximal minimization is continuous in the sense that the output point $x_b^{(i_j+1)}$ continuously depends on the input point $\acute{x}_b^{(i_j)}$ \citeSupp{Rockafellar:supp}.
Using the fact that $x_b^{(i_j+1)} \rightarrow \bar{x}_b$ and $\acute{x}_b^{(i_j)} \rightarrow \bar{x_b}$ as $j \rightarrow \infty$, \R{eq:t:prf:nonconvg} becomes
\be{
\label{eq:t:prf:convg}
\bar{x}_b = \argmin_{ x_b \in \bar{\cX}_b } \, \ip{ \nabla_{x_b} f_b (\bar{x}) }{ x_b - \bar{x}_b } + \frac{1}{2} \nm{ x_b - \bar{x}_b }_{\overbar{M}_b}^2 + r_b (x_b).
}
Thus, $\bar{x}_b$ satisfies the first-order optimality condition (see \R{eq:first_order_conK}) of \R{eq:t:prf:convg}:
\bes{
\ip{\nabla_{x_b} f_b (\bar{x}) + \bar{g}_b }{ x_b - \bar{x}_b } \geq 0, \quad \mbox{for any $x_b \in \bar{\cX}_b$}
}
and for some $\bar{g} \in \partial r_b (\bar{x}_b)$, which is equivalent to the Nash equilibrium condition \R{eq:Nash_EqCond}. This completes the proof.

\section{Proofs of Propositions \ref{p:Kernel:MajorQ} and \ref{p:spcd:MajorQ}}

\subsection{Proof of Proposition \ref{p:Kernel:MajorQ}} \label{sec:prf:p:Kernel:MajorQ}

To show that $M_{Q_{\Psi}} \succeq Q_{\Psi}^H Q_{\Psi}$, we use $\Sigma \succeq \widehat{Z}^H \widehat{Z}$ satisfying
\bes{
( I_K \otimes \Phi^{-1} ) \Sigma ( I_K \otimes \Phi ) \succeq ( I_K \otimes \Phi^{-1} ) \widehat{Z}^H \widehat{Z} ( I_K \otimes \Phi )
}
where $\Sigma \in \bbC^{\tilde{N}D \times \tilde{N}K}$ is a block diagonal matrix with diagonal matrices $\{ \Sigma_k : k = 1,\ldots,K \}$.
We seek to apply the property that, if a Hermitian matrix is diagonally dominant with nonnegative diagonal entries, then it is positive semidefinite.
Noting that $\Sigma - \widehat{Z}^H \widehat{Z} \succeq 0$ is a Hermitian matrix, this property can be applied to show $\Sigma \succeq  \widehat{Z}^H \widehat{Z}$.
Observe that the diagonal elements of $[ \Sigma - \widehat{Z}^H \widehat{Z} ]_{k,k}$ are nonnegative because
\bes{
\sum_{k' \neq k} \left| \sum_{l=1}^L \overline{ \left( \hat{z}_{l,k} \right)_i } \cdot \left( \hat{z}_{l,k'} \right)_i \right| \geq 0, \qquad k \in [K].
}
It now suffices to show that 
\bes{
\left| \left( [ \Sigma - \widehat{Z}^H \widehat{Z} ]_{k,k} \right)_{i,i} \right| \geq \sum_{k' \neq k} \left| \left( [ \Sigma - \widehat{Z}^H \widehat{Z} ]_{k,k'} \right)_{i,i} \right|.
}
This is true because the left terms and the right terms are identical, given by
\bes{
\sum_{k' \neq k} \left| \sum_{l=1}^L \overline{ \left( \hat{z}_{l,k} \right)_i } \cdot \left( \hat{z}_{l,k'} \right)_i \right|
}
for all $k = 1,\ldots,K$ and $i=1,\ldots,\tilde{N}$.
Combining these results completes the proof.

\subsection{Proofs of Proposition \ref{p:spcd:MajorQ}} \label{sec:prf:p:spcd:MajorQ}

Using the similar technique in Section \ref{sec:prf:p:Kernel:MajorQ}, the majorization matrix for $\Lambda^H \Lambda$ is given by a block diagonal matrix with diagonal blocks $\{ \Sigma'_k \}$ given in \R{eq:Sigma'k}.
Substituting this majorization matrix into \R{eq:G_bound1} completes the proof.

\section{\hspace{-0.2em}Proofs of Lemmas \ref{l:Kernel:diag1} \!\&\! \ref{l:spCd:diag1}, \ref{l:Kernel:identity}, \ref{l:Kernel:diag2} \!\&\! \ref{l:spCd:diag2}, and \ref{l:Kernel:diag:mltBlk}}

\subsection{Proofs of Lemmas \ref{l:Kernel:diag1} \& \ref{l:spCd:diag1}} \label{sec:prf:l:Kernel-spCD:diag1}

Note that $Q_{\Psi}^H Q_{\Psi}$ is a Hermitian matrix with nonnegative diagonal entries because its diagonal submatrices are given by
\bes{
[ Q_{\Psi}^H Q_{\Psi} ]_{k,k} =  \Phi^{-1} \sum_{l=1}^L \diag( | \hat{z}_{l,k} |^2 ) \Phi, \qquad k \in [K],
}
which imply that $[ Q_{\Psi}^H  Q ]_{k,k}$ is a circulant matrix with (identical) positive diagonal entries, and 
\bes{
[ Q_{\Psi}^H Q_{\Psi} ]_{k,k'} = [ Q_{\Psi}^H Q_{\Psi} ]_{k',k}^H, \qquad k \neq k' \in [K]
}
where $[ Q_{\Psi}^H Q_{\Psi} ]_{k,k'}$ is given as \R{eq:def:QH}. 
Applying Lemma \ref{l:diag(|psdA|1)} to the Hermitian matrix $Q_{\Psi}^H Q_{\Psi}$ completes the proof.

Repeating the similar procedure leads to a result in Lemma \ref{l:spCd:diag1}.

\subsection{Proof of Lemma \ref{l:Kernel:identity}} \label{sec:prf:l:Kernel:identity}

Observe that for any $x \in \bbC^{\tilde{N}}$
\eas{
x^H \Phi^{-1} \Sigma_k \Phi x  & = x^H \widetilde{\Phi}^H \Sigma_k \widetilde{\Phi} x =  \sum_{i=1}^{\tilde{N}} ( \Sigma_k )_{i,i} | y_i |^2 
\\
& \leq \max_{i=1,\ldots,\tilde{N}} \! \left\{ ( \Sigma_k )_{i,i} \right\} \sum_{i=1}^{\tilde{N}} | y_i |^2 
\\
& =  \max_{i=1,\ldots,\tilde{N}} \! \left\{ ( \Sigma_k )_{i,i} \right\} \cdot x^H I_{\tilde{N}} x
}
where we use $y = \widetilde{\Phi} x$ and $\| y \|_2^2 = \| x \|_2^2$ and $\widetilde{\Phi}$ denotes unitary DFT. 
This completes the proof.

\subsection{Proofs of Lemmas \ref{l:Kernel:diag2} \& \ref{l:spCd:diag2}} \label{sec:prf:l:Kernel-spCd:diag2}

The results directly follow by noting that $\{ \Phi^{-1}  \Sigma_k \Phi \}$ and $\{ \Phi^{-1}  \Sigma'_k \Phi \}$ are Hermitian (circulant) matrices with nonnegative diagonal entries, and applying Lemma \ref{l:diag(|psdA|1)}.

\subsection{Proof of Lemma \ref{l:Kernel:diag:mltBlk}} \label{sec:prf:l:Kernel:diag:mltBlk}

Using $P_B^T P_B \preceq I$, we have
\bes{
\Psi_k^H \Psi_k \preceq \Phi^{-1} \sum_{l=1}^L \diag( | \hat{z}_{l,k} |^2 ) \Phi.
}
Observe that the Hermitian matrix $\Phi^{-1} \sum_{l=1}^L \diag( | \hat{z}_{l,k} |^2 ) \Phi$ is positive semidefinite. Applying Lemma~\ref{l:diag(|psdA|1)} completes the proof.

\section{Accelerated Newton's Method to Solve \R{eq:prox_map:blk:BPGM:synthF_kernel}} \label{sec:AccNewton}

The optimal solution to \R{eq:prox_map:blk:BPGM:synthF_kernel} can be obtained by the classical approach for solving a quadratically constrained quadratic program (see, for example, \citeSupp[Ex.~4.22]{Boyd&Vandenberghe:supp}):\footnote{
Note that \R{eq:soln:prox_map:blk:BPGM:synthF_kernel} can be also applied to a circulant majorizer $M_{\Psi}$, e.g.,
\bes{
\left[ M_{\Psi} \right]_{k,k} = \Phi_K^{-1} E_k \Phi_K,
}
where $E_k = \diag \big( \big| \Phi_K P_S \Phi^{-1} \Sigma_k \Phi P_S^T \Phi_K^{-1} \big| 1_{K} \big)$ and $\Phi_K$ is a (unnormalized) DFT matrix of size $K \times K$. Unfortunately, an efficient scheme to compute the circulant majorizer is unknown, due to the difficulty in deriving the symbolic expression for the matrix of $| \cdot |$ (i.e., the matrix inside $| \cdot |$ is no longer circulant).
}
\be{
\label{eq:soln:prox_map:blk:BPGM:synthF_kernel}
d_k^{(i+1)} = \left( \left[ M_{\Psi}^{(i)} \right]_{k,k} + \varphi_k I_K \right)^{-1} \left[ M_{\Psi}^{(i)} \right]_{k,k} \nu_k^{(i)}
}
where the Lagrangian parameter is determined by $\varphi_k = \max \{ 0 , \varphi_k^{\star} \}$ and $\varphi_k^{\star}$ is the largest solution of the nonlinear equation $f (\varphi_k) = 1$, where
\be{
\label{eq:secular1}
f (\varphi_k) = \nm{ \left( \left[ M_{\Psi}^{(i)} \right]_{k,k} + \varphi_k I_K \right)^{-1}  \left[ M_{\Psi}^{(i)} \right]_{k,k} \nu_k^{(i)} }_2^2,
}
which is the so-called \textit{secular} equation, for $k=1,\ldots,K$. 
More specifically, the algorithm goes as follows. If $\| \nu_k^{(i)} \|_2 \leq 1$, then $d_k^{(i+1)} = \nu_k^{(i)}$ is the optimal solution.
Otherwise, one can obtain the optimal solution $d_k^{(i+1)}$ through \R{eq:soln:prox_map:blk:BPGM:synthF_kernel} with the Lagrangian parameter $\varphi_k = \varphi_k^{\star}$, where $\varphi_k^{\star}$ is optimized by solving the secular equation $f (\varphi_k) = 1$ and $f (\varphi_k)$ is given as \R{eq:secular1}.
To solve $f (\varphi_k) = 1$, we first rewrite \R{eq:secular1} by
\be{
\label{eq:secular2}
f (\varphi_k) = \sum_{j=1}^K \frac{ \left( \left[ M_{\Psi}^{(i)} \right]_{k,k} \right)_{j,j}^2 \left( \nu_k^{(i)} \right)_{j}^2 }{ \left(  \varphi_k + \left( \left[ M_{\Psi}^{(i)} \right]_{k,k} \right)_{j,j} \right)^2 }.
}
where $\{ ( [ M_{\Psi}^{(i)} ]_{k,k} )_{j,j} > 0 : j=1,\ldots,K \}$.
Noting that $f(0) > 1$ and $f(\varphi_k)$ monotonically decreases to zero as $\varphi_k \rightarrow \infty$, the nonlinear equation $f (\varphi_k) = 1$ has exactly one nonnegative solution $\varphi_k^{\star}$.  The optimal solution $\varphi_k^\star$ can be determined by using the classical Newton's method. 
To solve the secular equation $f (\varphi_k) = 1$ faster, we apply the accelerated Newton's method in \citeSupp{Reinsch:supp}:
\be{
\label{eq:acc_newton}
\varphi_k^{(\iota+1)} = \varphi_k^{(\iota)} - 2 \frac{f (\varphi_k^{(\iota)})}{f' (\varphi_k^{(\iota)})} \left( \sqrt{f (\varphi_k^{(\iota)})}  - 1 \right)
}
where $f(\varphi_k)$ is given as \R{eq:secular2},
\bes{
f'(\varphi_k) =  -2 \sum_{j=1}^K  \frac{ \left( \left[ M_{\Psi}^{(i)} \right]_{k,k} \right)_{j,j}^2 \left( \nu_k^{(i)} \right)_{j}^2 }{ \left(  \varphi_k + \left( \left[ M_{\Psi}^{(i)} \right]_{k,k} \right)_{j,j} \right)^3 },
}
and $\varphi_k^{(0)} = 0$. Note that \R{eq:acc_newton} approaches the optimal solution $\varphi_k^\star$ faster than the classical Newton's method.

\begin{figure*}[t!]
\small\addtolength{\tabcolsep}{-2pt}
\renewcommand{\arraystretch}{1.5}
\centering

\begin{tabular}{cccc}
{\small Learned filters} & \multicolumn{2}{c}{\small Examples of synthesized images} & {\small Convergence behavior} \\
\multirow{2}{*}[5.9em]{
\includegraphics[scale=0.52, trim=4em 6em 1.4em 5.2em, clip]{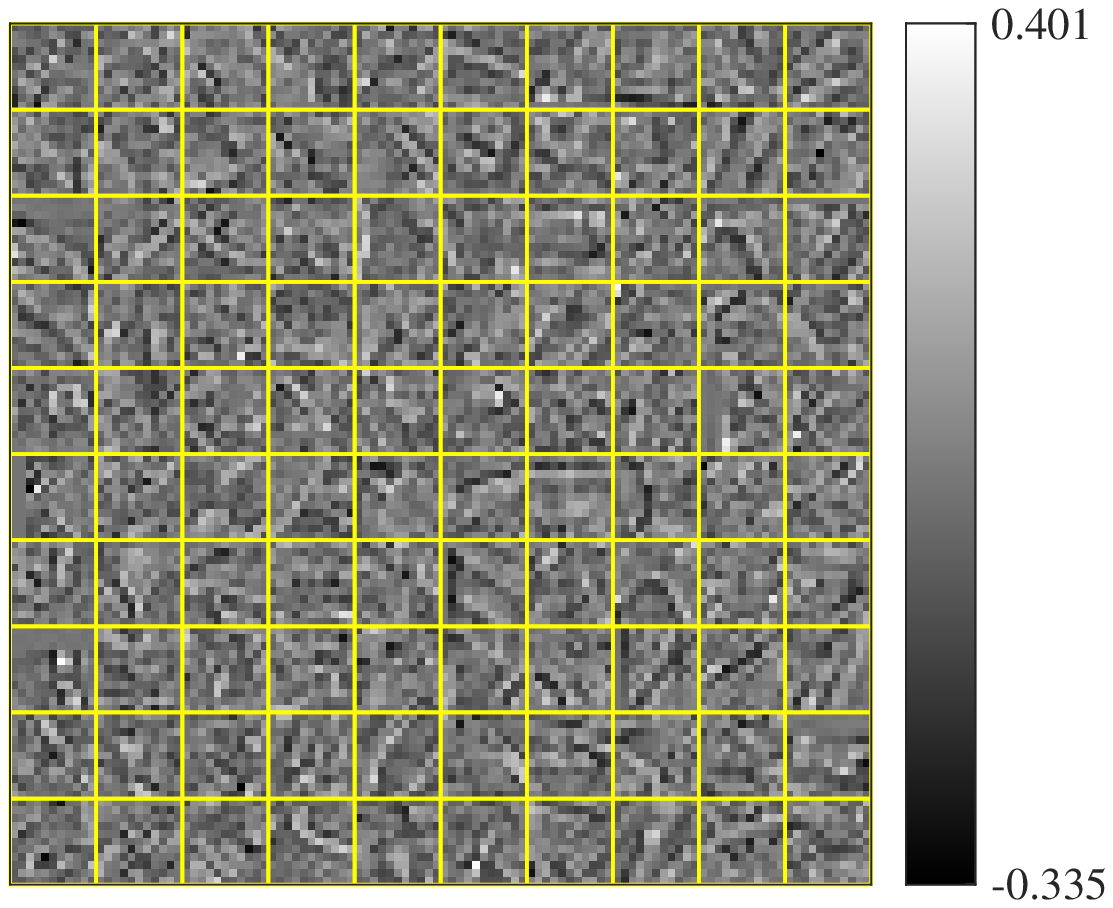} 
}
& \includegraphics[width=2.1cm,height=2.1cm,keepaspectratio]{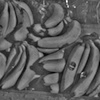}
& \includegraphics[width=2.1cm,height=2.1cm,keepaspectratio]{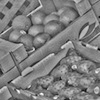}
& \multirow{2}{*}[5.75em]{
\includegraphics[scale=0.54, trim=0.2em 0.2em 1.8em 1.5em, clip]{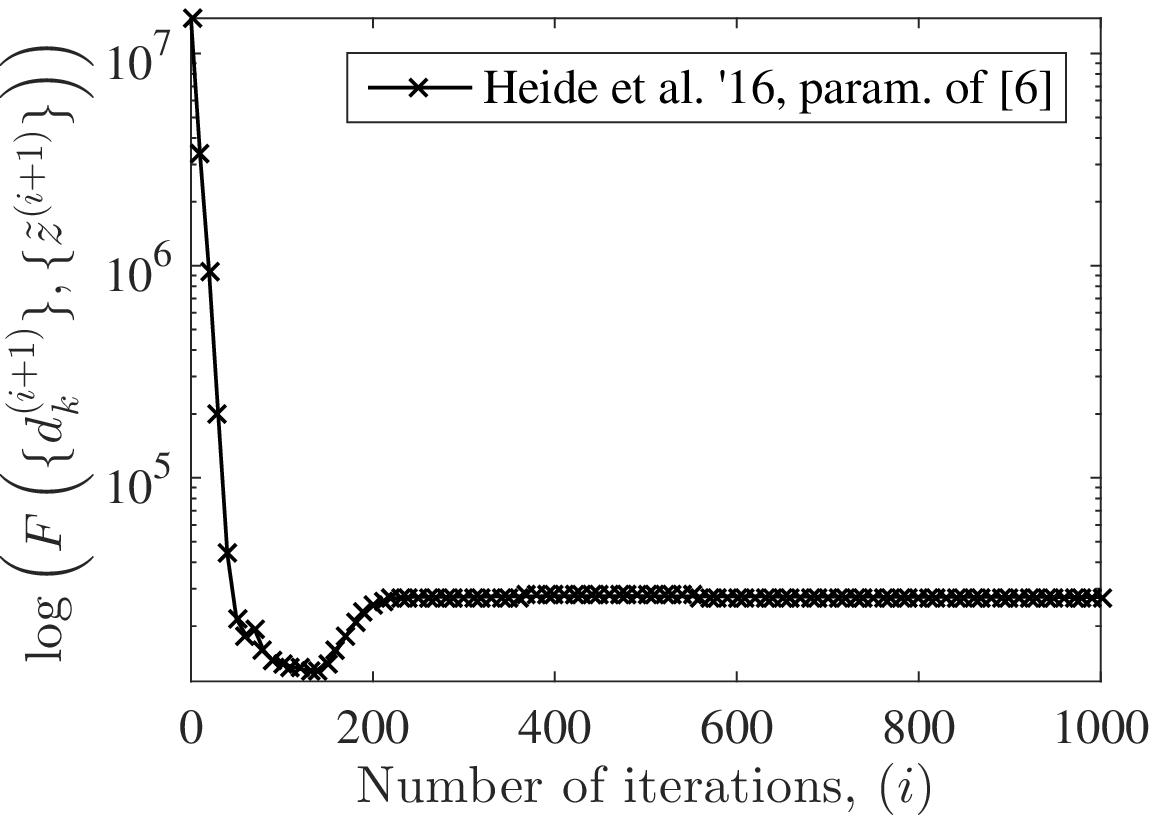} 
} 
\\
& \includegraphics[width=2.1cm,height=2.1cm,keepaspectratio]{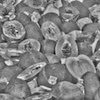} 
& \includegraphics[width=2.1cm,height=2.1cm,keepaspectratio]{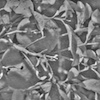}
\\
& {\small $\cdots$} & {\small $\cdots$} & \vspace{-0.5em} \\
\multicolumn{4}{c}{\small (a) The fruit dataset  ($L = 10$, $N = 100 \!\times\! 100$)} 
\vspace{0.5em} \\
\multirow{2}{*}[5.9em]{
\includegraphics[scale=0.52, trim=4em 6em 1.4em 5.2em, clip]{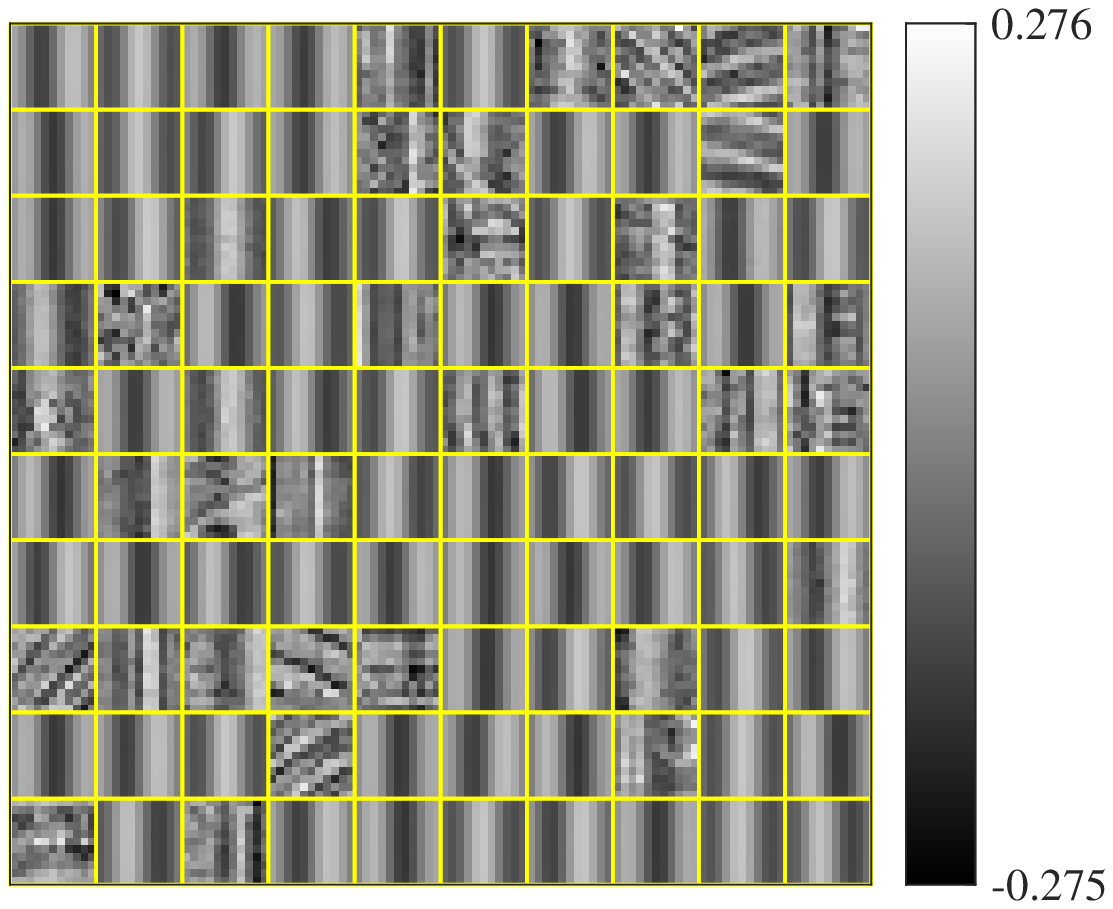} 
}
& \includegraphics[width=2.1cm,height=2.1cm,keepaspectratio]{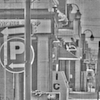}
& \includegraphics[width=2.1cm,height=2.1cm,keepaspectratio]{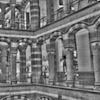}
& \multirow{2}{*}[5.75em]{
\includegraphics[scale=0.54, trim=0.2em 0.2em 1.8em 1.5em, clip]{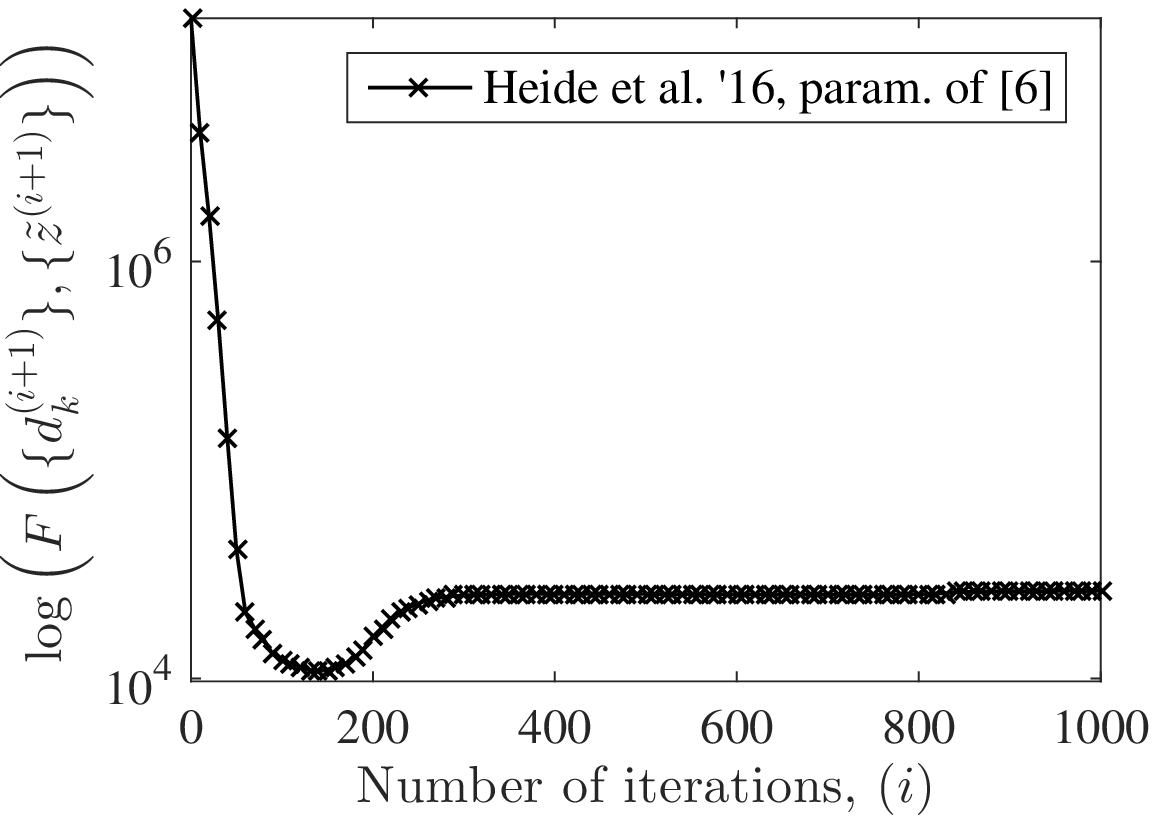} 
} 
\\
& \includegraphics[width=2.1cm,height=2.1cm,keepaspectratio]{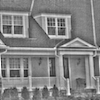} 
& \includegraphics[width=2.1cm,height=2.1cm,keepaspectratio]{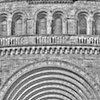}
\\
& {\small $\cdots$} & {\small $\cdots$} & \vspace{-0.5em} \\
\multicolumn{4}{c}{\small (b) The city dataset  ($L = 10$, $N = 100 \!\times\! 100$)} 
\vspace{0.5em} \\
\multirow{2}{*}[5.9em]{
\includegraphics[scale=0.52, trim=4em 6em 1.4em 5.2em, clip]{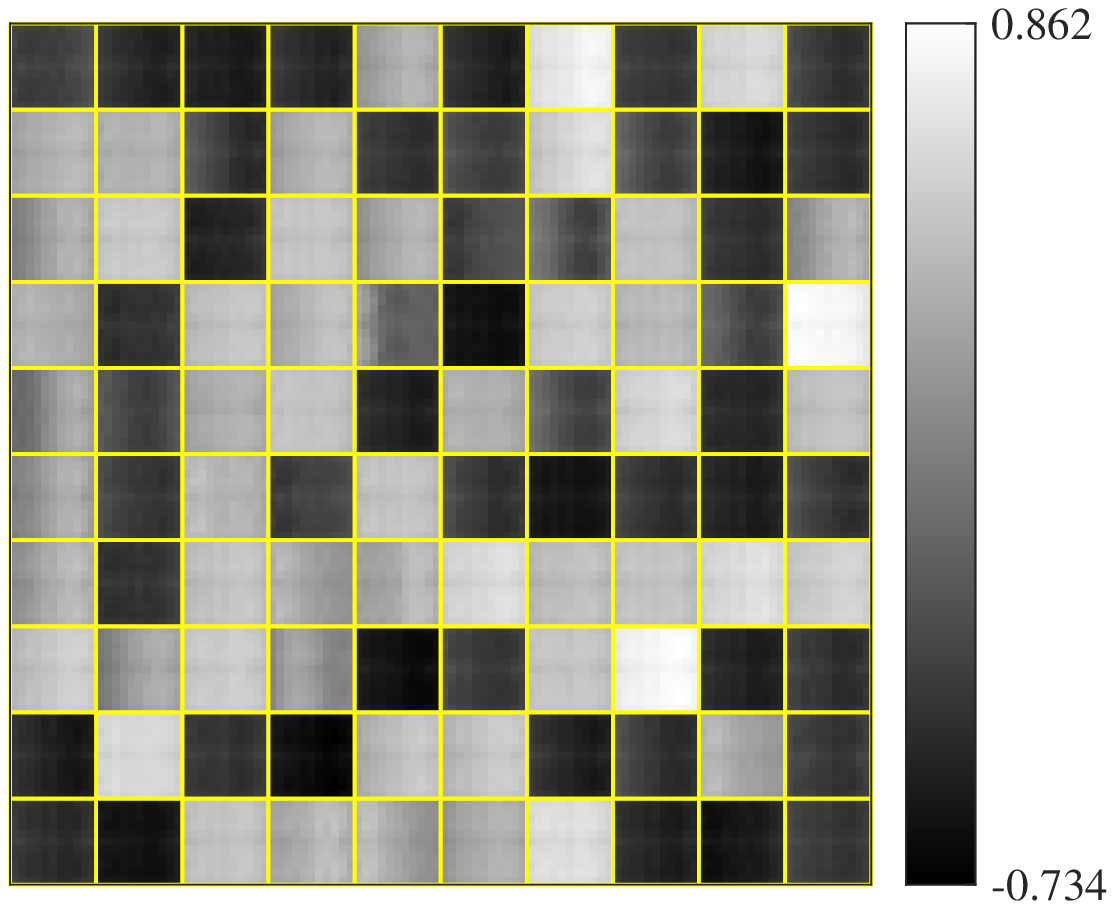} 
}
& \includegraphics[width=2.1cm,height=2.1cm,keepaspectratio]{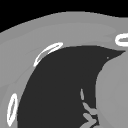}
& \includegraphics[width=2.1cm,height=2.1cm,keepaspectratio]{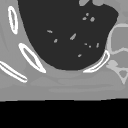}
& \multirow{2}{*}[5.75em]{
\includegraphics[scale=0.54, trim=0.2em 0.2em 1.8em 1.5em, clip]{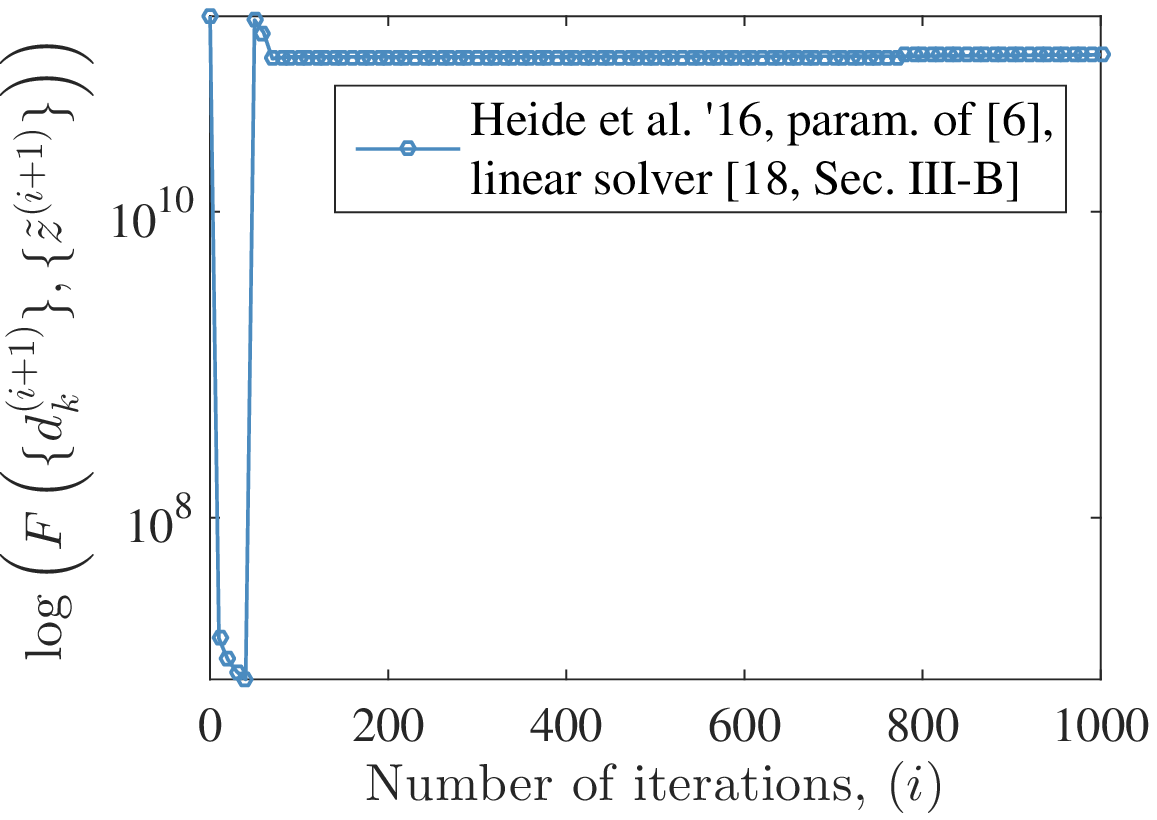} 
} 
\\
& \includegraphics[width=2.1cm,height=2.1cm,keepaspectratio]{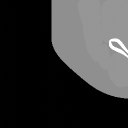} 
& \includegraphics[width=2.1cm,height=2.1cm,keepaspectratio]{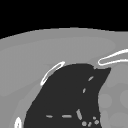}
\\
& {\small $\cdots$} & {\small $\cdots$} & \vspace{-0.5em} \\
\multicolumn{4}{c}{\small (c) The CT-(\romnum{2}) dataset  ($L = 80$, $N = 128 \!\times\! 128$)} 
\end{tabular}

\caption{Examples of CDL results by the ADMM algorithm \protect\citeSupp{Heide&eta:supp} from different datasets (the number of whole iterations is the product of the number of inner iterations and that of outer iterations). Although using the suggested ADMM parameters in \protect\citeSupp{Heide&eta:supp}, ADMM shows unstable convergence and the (empirically) convergent filters fail to capture structures of training images and do not have Gabor-like shapes. In addition, the sparse codes at the termination point have the sparsity of $100$\%.
Because the system matrices in each filter and sparse code update keep changing with the dependency of updated filters of sparse codes, one may want to develop adaptive ADMM parameter control schemes, e.g., selection schemes based on primal and dual residual norms \protect\citeSupp[\S 3.4.1]{Boyd&Parikh&Chu&Peleato&Eckstein:supp}, \protect\citeSupp[\S \Romnum{3}-D]{Wohlberg:supp} or condition number \protect\citeSupp{Goldstein&Osher:supp}, \protect\citeSupp[\S \Romnum{4}-C]{Ramani&Fessler:supp}.}
\label{fig:filters_ADMM}
\end{figure*}

\bibliographystyleSupp{IEEEtran}
\bibliographySupp{referencesSupp_Bobby}

\end{document}